\documentclass[10pt,journal,compsoc]{IEEEtran}

\usepackage{mathrsfs}
\usepackage{amsfonts}
\usepackage{amsmath}
\usepackage{bm}
\usepackage{bbm}
\usepackage{multirow}
\usepackage{makecell} 
\usepackage{tabto}
\usepackage{graphicx}
\usepackage{float} 
\usepackage{subfigure}
\usepackage{enumitem}
\usepackage{ragged2e}
\usepackage{color}
\usepackage{booktabs}

\ifCLASSOPTIONcompsoc
  \usepackage[nocompress]{cite}
\else
  \usepackage{cite}
\fi
\ifCLASSINFOpdf
\else
\fi

\begin{document}

\title{Hierarchical Contrastive Learning Enhanced Heterogeneous Graph Neural Network}

\author{Nian~Liu,
        Xiao~Wang,~\IEEEmembership{Member,~IEEE,}
        Hui~Han,
        Chuan~Shi*,~\IEEEmembership{Member,~IEEE,}
\IEEEcompsocitemizethanks{\IEEEcompsocthanksitem Nian Liu, Xiao Wang, Hui Han, Chuan Shi (corresponding author) are with the Beijing Key Lab of Intelligent Telecommunications Software and Multimedia, Beijing University of Posts and Telecommunications, China.
E-mail: {nianliu, xiaowang, hanhui, shichuan}@bupt.edu.cn
}}

\markboth{Journal of \LaTeX\ Class Files,~Vol.~14, No.~8, August~2015}%
{Shell \MakeLowercase{\textit{et al.}}: Bare Demo of IEEEtran.cls for Computer Society Journals}

\IEEEtitleabstractindextext{%
\begin{abstract}
\justifying
Heterogeneous graph neural networks (HGNNs) as an emerging technique have shown superior capacity of dealing with heterogeneous information network (HIN). However, most HGNNs follow a semi-supervised learning manner, which notably limits their wide use in reality since labels are usually scarce in real applications. Recently, contrastive learning, a self-supervised method, becomes one of the most exciting learning paradigms and shows great potential when there are no labels. In this paper, we study the problem of self-supervised HGNNs and propose a novel co-contrastive learning mechanism for HGNNs, named HeCo. Different from traditional contrastive learning which only focuses on contrasting positive and negative samples, HeCo employs cross-view contrastive mechanism. Specifically, two views of a HIN (network schema and meta-path views) are proposed to learn node embeddings, so as to capture both of local and high-order structures simultaneously. Then the cross-view contrastive learning, as well as a view mask mechanism, is proposed, which is able to extract the positive and negative embeddings from two views. This enables the two views to collaboratively supervise each other and finally learn high-level node embeddings. Moreover, to further boost the performance of HeCo, two additional methods are designed to generate harder negative samples with high quality. The essence of HeCo is to make positive samples from different views close to each other by cross-view contrast, and learn the factors invariant to two proposed views. However, besides the invariant factors, view-specific factors complementally provide the diverse structure information between different nodes, which also should be contained into the final embeddings. Therefore, we need to further explore each view independently and propose a modified model, called HeCo++. Specifically, HeCo++ conducts hierarchical contrastive learning, including cross-view and intra-view contrasts, which aims to enhance the mining of respective structures. Extensive experiments conducted on a variety of real-world networks show the superior performance of the proposed methods over the state-of-the-arts.
\end{abstract}

}

\maketitle

\IEEEdisplaynontitleabstractindextext

\IEEEpeerreviewmaketitle
\IEEEraisesectionheading{\section{Introduction}\label{sec:introduction}}

\IEEEPARstart{I}{n} the real world, heterogeneous information network (HIN) or heterogeneous graph (HG) \cite{zhang2023marml} is ubiquitous, due to the capacity of modeling various types of nodes and diverse interactions between them, such as bibliographic network \cite{hu2020strategies}, biomedical network \cite{DBLP:journals/nar/DavisGJSKMWWM17} and so on~\cite{wang2017community}. Recently, heterogeneous graph neural networks (HGNNs) have achieved great success in dealing with HIN data, because they are able to effectively combine the mechanism of message passing with complex heterogeneity, so that the complex structures and rich semantics can be well captured. So far, HGNNs have significantly promoted the development of HIN analysis towards real-world applications, e.g., recommender \cite{DBLP:conf/kdd/FanZHSHML19} and security system \cite{DBLP:conf/kdd/FanHZYA18}.

Basically, most HGNN studies belong to the semi-supervised learning paradigm, i.e., they usually design different heterogeneous message passing mechanisms to learn node embeddings, and then the learning procedure is supervised by a part of node labels. However, the requirement that some node labels have to be known beforehand is actually frequently violated, because it is very challenging or expensive to obtain labels in some real-world environments. For example, labeling an unknown gene accurately usually needs the enormous knowledge of molecular biology, which is not easy even for veteran researchers \cite{hu2020strategies}. 

Recently, self-supervised learning, aiming to spontaneously find supervised signals from the data itself, becomes a promising solution for the setting without explicit labels \cite{liu2020self}. Contrastive learning, as one typical technique of self-supervised learning, has attracted considerable attentions \cite{moco,simclr,dgi,mvgrl}. By extracting positive and negative samples in data, contrastive learning aims at maximizing the similarity between positive samples while minimizing the similarity between negative samples. In this way, contrastive learning is able to learn the discriminative embeddings even without labels. Despite the wide use of contrastive learning in computer vision \cite{simclr, moco} and natural language processing \cite{nlp1,nlp2}, little effort has been made towards investigating the great potential on HIN.

In practice, designing heterogeneous graph neural networks with contrastive learning is non-trivial, we need to carefully consider the characteristics of HIN and contrastive learning. This requires us to address the following three fundamental problems:

\textit{(1) How to design a heterogeneous contrastive mechanism.} A HIN consists of multiple types of nodes and relations, which naturally implies it possesses very complex structures. For example, meta-path, the composition of multiple relations, is usually used to capture the long-range structure in a HIN \cite{pathsim}. Different meta-paths represent different semantics, each of which reflects one aspect of HIN. To learn an effective node embedding which can fully encode these semantics, performing contrastive learning only on single meta-path view \cite{dmgi} is actually distant from sufficient. Therefore, investigating the heterogeneous cross-view contrastive mechanism is especially important for HGNNs. 

\textit{(2) How to select proper views in a HIN.} As mentioned before, cross-view contrastive learning is desired for HGNNs. Despite that one can extract many different views from a HIN because of the heterogeneity, one fundamental requirement is that the selected views should cover both of the local and high-order structures. Network schema, a meta template of HIN \cite{DBLP:journals/sigkdd/SunH12}, reflects the direct connections between nodes, which naturally captures the local structure. By contrast, meta-path is widely used to extract the high-order structure. As a consequence, both of the network schema and meta-path structure views should be carefully considered.

\textit{(3) How to set a difficult contrastive task.} It is well known that a proper contrastive task will further promote to learn a more discriminative embedding \cite{simclr,moco,cmc}. If two views are too similar, the supervised signal will be too weak to learn informative embedding. So we need to make the contrastive learning on these two views more complicated. For example, one strategy is to enhance the information diversity in two views, and the other is to generate harder negative samples of high quality. In short, designing a proper contrastive task is very crucial for HGNNs.

In this paper, we study the problem of self-supervised learning on HIN and propose a novel heterogeneous graph neural network with co-contrastive learning (HeCo). Specifically, different from previous contrastive learning which contrasts original network and the corrupted network, we choose network schema and meta-path structure as two views to collaboratively supervise each other. In network schema view, the node embedding is learned by aggregating information from its direct neighbors, which is able to capture the local structure. In meta-path view, the node embedding is learned by passing messages along multiple meta-paths, which aims at capturing high-order structure. In this way, we design a novel cross-view contrastive mechanism to capture complex structures in HIN. It means that for a node in one view, we push it towards its positives, while push it away from its negatives in the other view. To make contrast harder, we propose a view mask mechanism that hides different parts of network schema and meta-path, respectively, which will further enhance the diversity of two views and help extract higher-level factors from these two views. Moreover, we propose two additional methods to generate more negative samples with high quality. Finally, we modestly adapt traditional contrastive loss to the graph data, where a node has many positive samples rather than only one, different from methods \cite{simclr,moco} for CV. With the training going on, these two views are guided by each other and collaboratively optimize. 

HeCo maintains the invariant relationships between positives and negatives across views, which obeys a frequently-used inductive bias in traditional contrastive learning. This bias hints that the invariant part between different views often describes the essence of target nodes, which is beneficial to behave in downstream tasks. However, different views often contain mutually complementary information, and this information also should be taken into account to ensure the completeness of information. Therefore, it is necessary to exploit the invariant and view-specific messages simultaneously in a comprehensive way. To maintain invariant factors, we project target node embeddings from network schema view and meta-path view into a specific space, and still deploy cross-view contrastive learning, which is analogous to HeCo. Moreover, to depict view-specific factors, we additionally project target node embeddings into another two different spaces, each of which is for one view respectively. In each space, we introduce intra-view contrast to mine corresponding view, which means for a node in this view, we only shorten its distances with positives and enlarge that with negatives only in the same view. Thus the extended new model, HeCo++, can better explore view-specific information rather than only focus on the invariant part. Two new involved intra-view contrasts serve as a supplement to original HeCo.

It is worthwhile to highlight our contributions as follows:

\begin{itemize}
    \item To our best knowledge, this is the first attempt to study the self-supervised heterogeneous graph neural networks based on the cross-view contrastive learning. By contrastive learning based on cross-view manner, the high-level factors can be captured, enabling HGNNs to be better applied to real world applications without label supervision.
    \item We propose a novel heterogeneous graph neural network with co-contrastive learning, HeCo. HeCo innovatively employs network schema and meta-path views to collaboratively supervise each other, moreover, a view mask mechanism is designed to further enhance contrastive performance. Additionally, two additional methods of HeCo, named as HeCo\_GAN and HeCo\_MU, are proposed to generate high-quality negative samples. 
    \item We futher introduce the intra-view contrast into original HeCo, as the extended model HeCo++. The modified model can not only capture the beneficial commonality of two views, but also fully explore view-specific information.
    \item We conduct diverse experiments on four public datasets and the proposed HeCo outperforms the state-of-the-arts and even semi-supervised method, which demonstrates the effectiveness of HeCo from various aspects.
\end{itemize}

Please note that the preliminary work has been accepted as a full paper at the research track of the 27th ACM SIGKDD Conference on Knowledge Discovery and Data Mining \cite{me}. Based on the conference paper, we substantially extend the original work from following aspects: (1) We extend original HeCo by introducing intra-view contrastive learning, which aims to enhance the mining of view-specific structures. With the combination of cross-view contrastive learning and intra-view contrastive learning, the learnt embeddings can capture both the invariant factors and view-specific factors of two proposed views. And we provide detailed justifications of the extended model and algorithm. (2) To further verify the effectiveness of the proposed models, we significantly enrich experiments. On one hand, we complement existing experiments about original HeCo on more datasets. On the other hand, we also evaluate the extended model, HeCo++, on original experiments, including node classification, clustering, visualization and variant analysis. Besides, We also give more analyses on hyper-parameters of HeCo++ to further explore its properties. (3) A comprehensive survey of related work is presented in our paper, including graph neural network, heterogeneous graph neural network and contrastive learning. We also carefully polish our paper to improve the language quality.

\section{Related Work}\label{sec:reference}

In this section,we review some closely related studies, including graph neural network, heterogeneous graph neural network and contrastive learning.

\textbf{Graph Neural Network.} Graph neural networks (GNNs) have attracted considerable attentions, because of their capacity to deal with non-euclidean graph data. The main focus of GNNs is how to effectively aggregate and transform information from neighbors. Generally speaking, GNNs can be broadly divided into two categories, spectral-based and spatial-based.

Spectral-based GNNs are inheritance of graph signal processing, and define graph convolution operation in spectral domain. For example, \cite{spe1} utilizes Fourier bias to decompose graph signals; \cite{spe2} employs the Chebyshev expansion of the graph Laplacian to improve the efficiency. For another line, spatial-based GNNs greatly simplify above convolution by only focusing on neighbors. For example, GCN \cite{gcn} simply averages information of one-hop neighbors. GraphSAGE \cite{GraphSAGE} only randomly fuses a part of neighbors with various poolings, and it is suitable to inductive learning. GAE \cite{gae} follows the idea of generative models, and tries to reconstruct the adjacency matrix with hidden embeddings. More detailed surveys can be found in \cite{wu2021comprehensive}.

Although GNNs show a great success, they are proposed to homogeneous graphs, while cannot directly deal with heterogeneous graphs containing different types of nodes and edges.

\textbf{Heterogeneous Graph Neural Network}. Facing with the above problem, some researchers focus on adjusting GNNs to heterogeneous graphs. One major line of work leverages meta-paths to pass message so as to capture the multiple semantics. For example, HAN \cite{han} firstly fuses meta-path based neighbors with node-level attention and then aggregates different meta-paths with semantic-level attention. On this basis, MAGNN \cite{magnn} takes intermediate nodes of meta-paths into account, and proposes several mechanisms to combine all nodes over meta-paths. Discarding pre-defined meta-paths, GTN \cite{gtn} is proposed to automatically identify useful connections. On another line, some methods either directly use the one-hop relation between nodes. For example, HGT \cite{hgt} assigns different attention weight matrices to different meta-relations, empowering the model to take type information into consideration. HetGNN \cite{hetegnn} samples a fixed size of neighbors, and fuses their features using LSTMs. NSHE \cite{nshe} focuses on network schema, and preserves pairwise and network schema proximity simultaneously. However, the above methods can not exploit supervised signals from data itself to learn general node embeddings.

\textbf{Contrastive Learning}. As one typical technique in self-supervised learning, contrastive learning is devoted to pushing forward positive samples while pushing away negative samples, and achieves great success in CV \cite{simclr,moco} and NLP \cite{nlp1,nlp2}. Here we mainly focus on reviewing the graph related contrastive learning methods. Specifically, DGI \cite{dgi} builds local patches and global summary as positive pairs and considers nodes in a randomly corrupted graph as negatives, and utilizes Infomax \cite{infomax} theory to contrast. Along this line, GMI \cite{gmi} is proposed to contrast between center node and its local patch from node features and topological structure. MVGRL \cite{mvgrl} employs cross-view contrast in homogeneous graphs and experiments composition between different views. In heterogeneous domain, DMGI \cite{dmgi} conducts contrastive learning between original network and corrupted network on each single view, meta-path, and designs a consensus regularization to guide the fusion of different meta-paths. In application, social recommendation~\cite{DBLP:journals/snam/TangHL13} can be viewed as a classical scenario of HIN, where users and items are interacted via multiple ways and social relations between users are also provided. In this field, contrastive learning also contributes a lot. MHCH~\cite{mhcn} utilizes hierarchical mutual information maximization between users, sub-hypergraph and whole graph. SPET~\cite{sept} firstly augments users data views with social relation, and then proposes a tri-training framework to encode multi-view information. SMIN~\cite{smin} models global context-aware and topology-aware mutual information to inject high-order interactive patterns into social recommenders. KCGN~\cite{kcgn} firstly learns the user- and item-specific embeddings to preserve local connection information and then conducts noise-contrastive learning for users and items. In summary, these methods generally choose intra-view contrast between local patches and high-order vectors~\cite{mhcn, smin, kcgn}, or between calculated positives and negatives~\cite{sept}. Nevertheless, there is still a lack of methods contrasting across views in HIN so that the high-level and invariant factors can be captured.
\begin{figure}[t]
    \centering
    \includegraphics[scale=0.2]{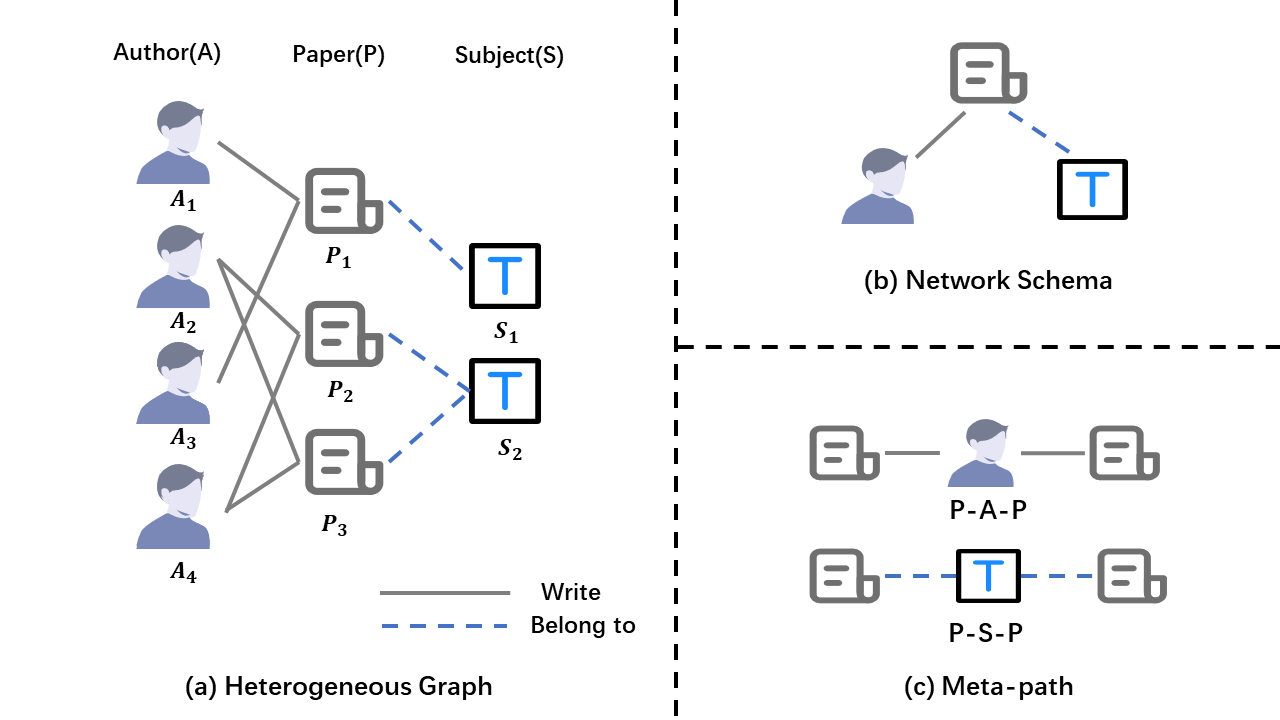}
    \caption{A toy example of HIN (ACM) and relative illustrations of meta-path and network schema.}
    \label{hg}
\end{figure}

\begin{figure*}[t]
  \centering
  \includegraphics[width=0.85\linewidth, height=70mm]{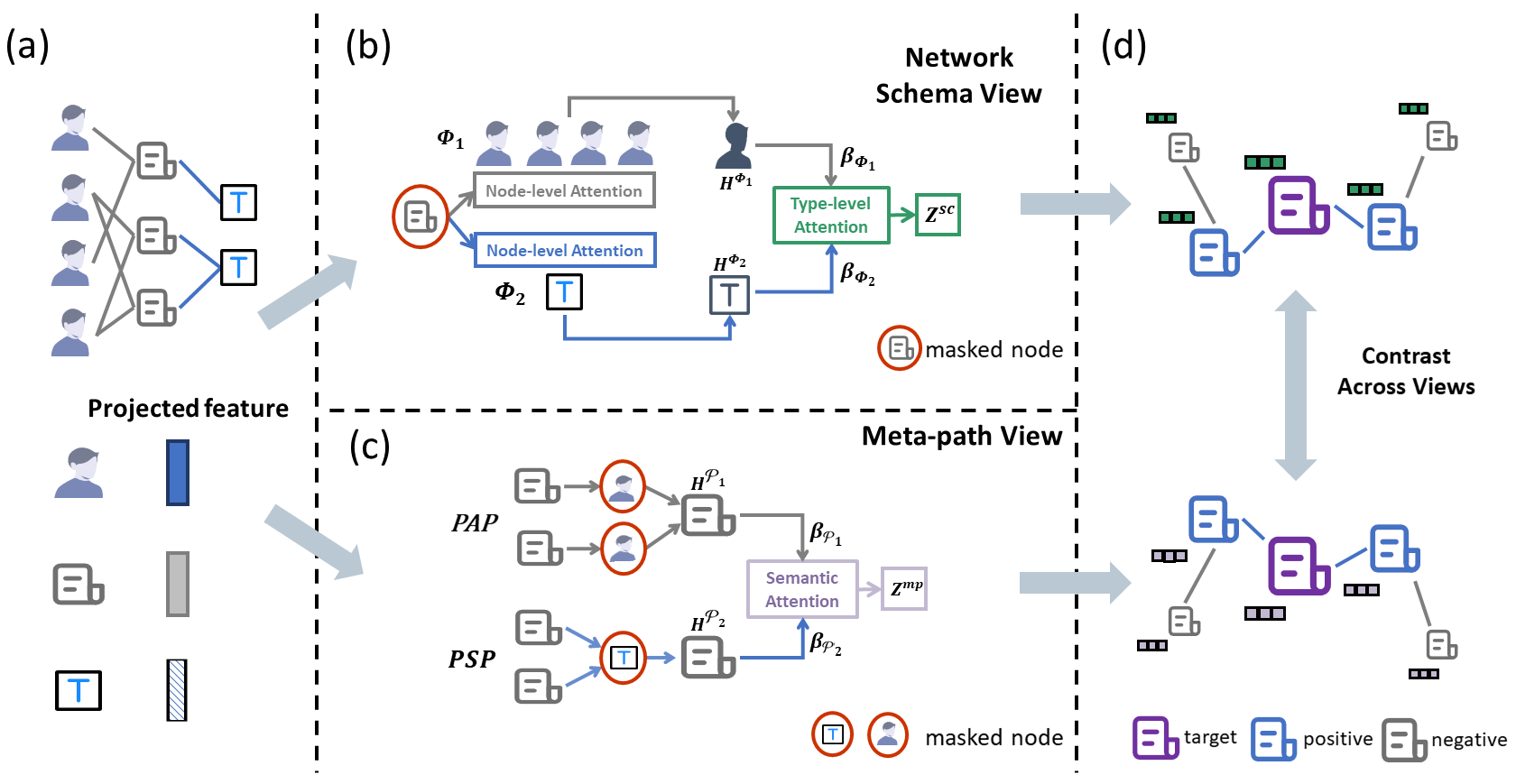}
  \caption{The overall architecture of our proposed HeCo.}
  \label{model}
\end{figure*}

\section{Preliminary}\label{sec:preliminary}

In this section, we formally define some significant concepts related to HIN as follows:

\textbf{Definition 3.1. Heterogeneous Information Network.} Heterogeneous Information Network (HIN) is defined as a network $\mathcal{G}=(\mathcal{V},\mathcal{E},\mathcal{A},\mathcal{R},\mathcal{\phi},\mathcal{\varphi})$, where $\mathcal{V}$ and $\mathcal{E}$ denote sets of nodes and edges, and it is associated with a node type mapping function $\mathcal{\phi}:\mathcal{V}\rightarrow\mathcal{A}$ and a edge type mapping function $\mathcal{\varphi}:\mathcal{E}\rightarrow\mathcal{R}$, where $\mathcal{A}$ and $\mathcal{R}$ denote sets of object and link types, and $|\mathcal{A}+\mathcal{R}|>2$.

Figure \ref{hg} (a) illustrates an example of HIN. There are three types of nodes, including author (A), paper (P) and subject (S). Meanwhile, there are two types of relations ("write" and "belong to"), i.e., author writes paper, and paper belongs to subject.

\textbf{Definition 3.2. Network Schema.} The network schema, noted as $T_G=(\mathcal{A},\mathcal{R})$, is a meta template for a HIN $\mathcal{G}$. $T_G$ is a directed graph defined over object types $\mathcal{A}$, with edges as relations from $\mathcal{R}$.

For example, Figure \ref{hg} (b) is the network schema of (a), in which we can know that paper is written by author and belongs to subject. Network schema is used to describe the direct connections between different nodes, which represents local structure.

\textbf{Definition 3.3. Meta-path.} A meta-path $\mathcal{P}$ is defined as a path, which is in the form of $A_1\stackrel{R_1}{\longrightarrow}A_2\stackrel{R_2}{\longrightarrow}\dots\stackrel{R_l}{\longrightarrow}A_{l+1}$ (abbreviated as $A_1A_2\dots A_{l+1}$), which describes a composite relation $R=R_1\circ R_2\circ \dots\circ R_l$ between node types $A_1$ and $A_{l+1}$, where $\circ$ denotes the composition operator on relations.

For example, Figure \ref{hg} (c) shows two meta-paths extracted from HIN in Figure \ref{hg} (a). PAP describes that two papers are written by the same author, and PSP describes that two papers belong to the same subject. Because meta-path is the combination of multiple relations, it contains high-order structures.
\section{The Proposed Model: HeC\lowercase{o}}\label{sec:model}

In this section, we propose HeCo, a novel heterogeneous graph neural network with co-contrastive learning, and the overall architecture is shown in Figure \ref{model}. Our model encodes nodes from network schema view and meta-path view, which fully captures the structures of an HIN. During the encoding, we creatively involve a view mask mechanism, which makes these two views complement and supervise mutually. With the two view-specific embeddings, we employ a contrastive learning across these two views. Given the high correlation between nodes, we redefine the positive samples of a node in HIN and design a optimization strategy specially. 

\subsection{Node Feature Transformation}

Because there are different types of nodes in an HIN, their features usually lie in different spaces. So first, we need to project features of all types of nodes into a common latent vector space, as shown in Figure \ref{model} (a). Specifically, for a node $i$ with type $\phi_i$, we design a type-specific mapping matrix $\bm{W_{\phi_i}}$ to transform its feature $\bm{x_i}$ into common space as follows:
\begin{equation}
  \bm{h_i}=\sigma\left(\bm{W_{\phi_i}} \cdot \bm{x_i}+\bm{b_{\phi_i}}\right),
\end{equation}
where $\bm{h_i}\in\mathbb{R}^{d\times1}$ is the projected feature of node $i$, $\sigma(\cdot)$ is an activation function, and $\bm{b_{\phi_i}}$ denotes as vector bias, respectively.

\subsection{Network Schema View Guided Encoder}

Now we aim to learn the embedding of node $i$ under network schema view, illustrated as Figure \ref{model} (b). According to network schema, we assume that the target node $i$ connects with $S$ other types of nodes$\{\Phi_1, \Phi_2,\dots,\Phi_S\}$, so the neighbors with type $\Phi_m$ of node $i$ can be defined as $N_i^{\Phi_m}$. For node $i$, different types of neighbors contribute differently to its embedding, and so do the different nodes with the same type. So, we employ attention mechanism here in node-level and type-level to hierarchically aggregate messages from other types of neighbors to target node $i$. Specifically, we first apply node-level attention to fuse neighbors with type $\Phi_m$:
\begin{equation}
    \bm{h_i^{\Phi_m}}=\sigma\left(\sum\limits_{j \in N_i^{\Phi_m}}\alpha_{i,j}^{\Phi_m} \cdot \bm{h_j}\right),
    \label{sc_1}
\end{equation}
where $\sigma$ is a nonlinear activation, $\bm{h_j}$ is the projected feature of node $j$, and $\alpha_{i,j}^{\Phi_m}$ denotes the attention value of node $j$ with type $\Phi_m$ to node $i$. It can be calculated as follows:
\begin{equation}
    \alpha_{i,j}^{\Phi_m}=\frac{\exp\left(LeakyReLU\left(\textbf{a}_{\Phi_m}^\top\cdot[\bm{h_i||h_j}]\right)\right)}{\sum\limits_{l\in N_i^{\Phi_m}} \exp\left(LeakyReLU\left(\textbf{a}_{\Phi_m}^\top\cdot[\bm{h_i||h_l}]\right)\right)},
    \label{sc_2}
\end{equation}
where $\textbf{a}_{\Phi_m}\in \mathbb{R}^{2d\times1}$ is the node-level attention vector for $\Phi_m$ and $||$ denotes concatenate operation. Please notice that in practice, we do not aggregate the information from all the neighbors in $N_i^{\Phi_m}$, but we randomly sample a part of neighbors every epoch. Specifically, if the number of neighbors with type $\Phi_m$ exceeds a predefined threshold $T_{\Phi_m}$, we unrepeatably select $T_{\Phi_m}$ neighbors as $N_i^{\Phi_m}$, otherwise the $T_{\Phi_m}$ neighbors are selected repeatably. In this way, we ensure that every node aggregates the same amount of information from neighbors, and promote diversity of embeddings in each epoch under this view, which will make following contrast task more challenging. 

Once we get all type embeddings $\{\bm{h_i^{\Phi_1}, ..., h_i^{\Phi_S}}\}$, we utilize type-level attention to fuse them together to get the final embedding $\bm{z_i^{sc}}$ for node i under network schema view. First, we measure the weight of each node type as follows:
\begin{equation}
\begin{aligned}
    w_{\Phi_m}&=\frac{1}{|V|}\sum\limits_{i\in V} \textbf{a}_{sc}^\top \cdot \tanh\left(\textbf{W}_{sc}\bm{h_i^{\Phi_m}}+\textbf{b}_{sc}\right),\\
    \beta_{\Phi_m}&=\frac{\exp\left(w_{\Phi_m}\right)}{\sum_{i=1}^S\exp\left(w_{\Phi_i}\right)},
    \label{sc_4}
\end{aligned}
\end{equation}
where $V$ is the set of target nodes, \textbf{W}$_{sc}$ $\in\mathbb{R}^{d\times d}$ and \textbf{b}$_{sc}$ $\in\mathbb{R}^{d\times1}$ are learnable parameters, and $\textbf{a}_{sc}$ denotes type-level attention vector. $\beta_{\Phi_m}$ is interpreted as the importance of type $\Phi_m$ to target node $i$. So, we weighted sum the type embeddings to get $\bm{z_i^{sc}}$:
\begin{equation}
    \bm{z_i^{sc}}=\sum_{m=1}^S \beta_{\Phi_m}\cdot \bm{h_i^{\Phi_m}}.
    \label{sc_5}
\end{equation}

\subsection{Meta-path View Guided Encoder}

Here we aim to learn the node embedding in the view of high-order meta-path structure, described in Figure \ref{model} (c). Specifically, given a meta-path $\mathcal{P}_n$ from $M$ meta-paths $\{\mathcal{P}_1,\mathcal{P}_2,\dots,\mathcal{P}_M\}$ that start from node $i$, we can get the meta-path based neighbors $N_i^{\mathcal{P}_n}$. For example, as shown in Figure \ref{hg} (a), $P_2$ is a neighbor of $P_3$ based on meta-path $PAP$. Each meta-path represents one semantic similarity, and we apply meta-path specific GCN \cite{gcn} to encode this characteristic:
\begin{equation}
    \bm{h_i^{\mathcal{P}_n}}=\frac{1}{d_i+1}\bm{h_i}+\sum\limits_{j\in{N_i^{\mathcal{P}_n}}}\frac{1}{\sqrt{(d_i+1)(d_j+1)}}\bm{h_j},
\end{equation}
where $d_i$ and $d_j$ are degrees of node $i$ and $j$, and $\bm{h_i}$ and $\bm{h_j}$ are their projected features, respectively. With $M$ meta-paths, we can get $M$ embeddings $\{\bm{h_i^{\mathcal{P}_1}, ..., h_i^{\mathcal{P}_M}}\}$ for node $i$. Then we utilize semantic-level attentions to fuse them into the final embedding $\bm{z_i^{mp}}$ under the meta-path view: 
\begin{equation}
\label{zimp}
    \bm{z_i^{mp}}=\sum_{n=1}^M \beta_{\mathcal{P}_n}\cdot \bm{h_i^{\mathcal{P}_n}},
\end{equation}
where $\beta_{\mathcal{P}_n}$ weighs the importance of meta-path $\mathcal{P}_n$, which is calculated as follows:
\begin{equation}
\begin{aligned}
    w_{\mathcal{P}_n}&=\frac{1}{|V|}\sum\limits_{i\in V} \textbf{a}_{mp}^\top \cdot \tanh\left(\textbf{W}_{mp}\bm{h_i}^{\mathcal{P}_n}+\textbf{b}_{mp}\right),\\
    \beta_{\mathcal{P}_n}&=\frac{\exp\left(w_{\mathcal{P}_n}\right)}{\sum_{i=1}^M\exp\left(w_{\mathcal{P}_i}\right)},
\end{aligned}
\end{equation}
where \textbf{W}$_{mp}$ $\in\mathbb{R}^{d\times d}$ and \textbf{b}$_{mp}$ $\in\mathbb{R}^{d\times1}$ are the learnable parameters, and $\textbf{a}_{mp}$ denotes the semantic-level attention vector.

\begin{figure}[t]
    \centering
    \includegraphics[scale=0.2]{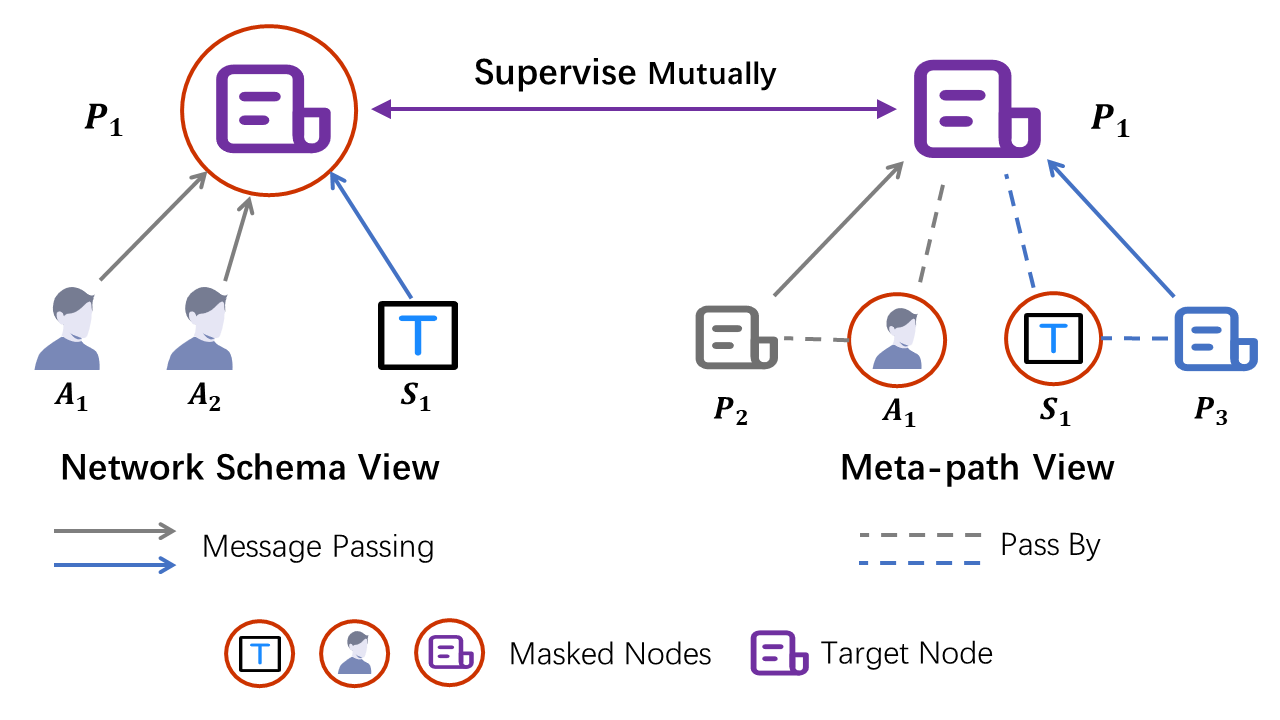}
    \caption{A schematic diagram of view mask mechanism.}
    \label{view}
\end{figure}

\subsection{View Mask Mechanism}
During the generation of $\bm{z_i^{sc}}$ and $\bm{z_i^{mp}}$, we design a view mask mechanism that hides different parts of network schema and meta-path views, respectively. In particular, we give a schematic diagram on ACM~\cite{nshe} in Figure \ref{view}, where the target node is $P_1$. In the process of network schema encoding, $P_1$ only aggregates its neighbors including authors $A_1, A_2$ and subject $S_1$ into $\bm{z_1^{sc}}$, but the message from itself is masked. While in the process of meta-path encoding, message only passes along meta-paths (e.g. PAP, PSP) from $P_2$ and $P_3$ to target $P_1$ to generate $\bm{z_1^{mp}}$, while the information of intermediate nodes $A_1$ and $S_1$ are discarded. Therefore, the embeddings of node $P_1$ learned from these two parts are correlated but also complementary. They can supervise the training of each other, which presents a collaborative trend.

\subsection{Collaboratively Contrastive Optimization}
After getting the $\bm{z_i^{sc}}$ and $\bm{z_i^{mp}}$ for node $i$ from the above two views, we feed them into an MLP with one hidden layer to map them into the space where contrastive loss is calculated:
\begin{equation}
\begin{aligned}
    \bm{z_i^{sc}\_cross} &= \bm{W^{(2)}}\sigma\left(\bm{W^{(1)}z_i^{sc}}+\bm{b^{(1)}}\right)+\bm{b^{(2)}},\\
    \bm{z_i^{mp}\_cross} &= \bm{W^{(2)}}\sigma\left(\bm{W^{(1)}z_i^{mp}}+\bm{b^{(1)}}\right)+\bm{b^{(2)}},
    \label{proj}
\end{aligned}
\end{equation}
where $\sigma$ is ELU non-linear function. It should be pointed out that $\{\bm{W^{(2)},W^{(1)},b^{(2)},b^{(1)}}\}$ are shared by two views embeddings. Next, when calculate contrastive loss, we need to define positive and negative samples in HIN. In computer vision, generally, one image only considers its augmentations as positive samples, and treats other images as negative samples \cite{moco,simclr}. In an HIN, given a node under network schema view, we can simply define its embedding learned by meta-path view as the positive sample. However, consider that nodes are highly-correlated because of edges, we propose a new positive selection strategy, i.e., if two nodes are connected by many meta-paths, they are positive samples, as shown in Figure \ref{model} (d) where links between papers represent they are positive samples of each other. One advantage of such strategy is that the selected positive samples can well reflect local structure of the target node.

For node $i$ and $j$, we first define a function $\mathbb{C}_i(\cdot)$ to count the number of meta-paths connecting these two nodes:
\begin{equation}
    \mathbb{C}_i(j) = \sum\limits_{n=1}^M \mathbbm{1}\left(j\in N_i^{\mathcal{P}_n}\right),
    \label{s_matrix}
\end{equation}
where $\mathbbm{1}(\cdot)$ is the indicator function. Then we construct a set $S_i=\{j|j\in V\ and\ \mathbb{C}_i(j)\neq 0\}$ and sort it in the descending order based on the value of $\mathbb{C}_i(\cdot)$. Next we set a threshold $T_{pos}$, and if $|S_i|\textgreater T_{pos}$, we select first $T_{pos}$ nodes from $S_i$ as positive samples of $i$, denotes as $\mathbb{P}_i$, otherwise all nodes in $S_i$ are retained. And we naturally treat all left nodes as negative samples of $i$, denotes as $\mathbb{N}_i$.

With the positive sample set $\mathbb{P}_i$ and negative sample set $\mathbb{N}_i$, we have the following contrastive loss under network schema view:
\begin{equation}
\begin{aligned}
\label{cl}
    &\mathcal{L}_i^{sc}\_cross\\
    &=-\log\frac{\sum_{j\in\mathbb{P}_i}exp\left(sim\left(\bm{z_i^{sc}}\_cross,\bm{z_j^{mp}}\_cross\right)/\tau\right)}{\sum_{k\in\{\mathbb{P}_i\bigcup\mathbb{N}_i\}}exp\left(sim\left(\bm{z_i^{sc}}\_cross,\bm{z_k^{mp}}\_cross\right)/\tau\right)},
\end{aligned}
\end{equation}
where $sim(\bm{u,v})$ denotes the cosine similarity between two vectors $\bm{u}$ and $\bm{v}$, and $\tau$ denotes a temperature parameter. We can see that different from traditional contrastive loss \cite{simclr,moco}, which usually only focuses on one positive pair in the numerator of eq.\eqref{cl}, here we consider multiple positive pairs. Also please note that for two nodes in a pair, the target embedding is from the network schema view ($\bm{z_i^{sc}}\_cross$) and the embeddings of positive and negative samples are from the meta-path view ($\bm{z_k^{mp}}\_cross$). In this way, we realize the cross-view self-supervision.

The contrastive loss $\mathcal{L}_i^{mp}$ is similar as $\mathcal{L}_i^{sc}$, but differently, the target embedding is from the meta-path view while the embeddings of positive and negative samples are from the network schema view:
\begin{equation}
\begin{aligned}
\label{cll}
    &\mathcal{L}_i^{mp}\_cross\\
    &=-\log\frac{\sum_{j\in\mathbb{P}_i}exp\left(sim\left(\bm{z_i^{mp}}\_cross,\bm{z_j^{sc}}\_cross\right)/\tau\right)}{\sum_{k\in\{\mathbb{P}_i\bigcup\mathbb{N}_i\}}exp\left(sim\left(\bm{z_i^{mp}}\_cross,\bm{z_k^{sc}}\_cross\right)/\tau\right)},
\end{aligned}
\end{equation}

The overall objective is given as follows:
\begin{equation}
    \label{loss}
    \mathcal{L} = \frac{1}{|V|}\sum_{i\in V}\left[\lambda\cdot\mathcal{L}_i^{sc}\_cross+\left(1-\lambda\right)\cdot\mathcal{L}_i^{mp}\_cross\right],
\end{equation}
where $\lambda$ is a coefficient to balance the effect of two views. We can optimize the proposed model via back propagation and learn the embeddings of nodes. In the end, we use $\bm{z^{mp}}$ to perform downstream tasks because nodes of target type explicitly participates into the generation of $\bm{z^{mp}}$.

\subsection{Generate Harder Negative Samples}
\label{extension}
It is well established that a harder negative sample is very important for contrastive learning \cite{mochi}. Therefore, to further improve the performance of HeCo, here we propose two additional models with new negative sample generation strategies. 

\textbf{HeCo\_GAN} GAN-based models \cite{gan,hegan} consist of a generator and a discriminator, and aim at forcing generator to generate fake samples, which can finally fool a well-trained discriminator. Besides negatives selected from original HIN, we additionally generate harder negatives with GAN. Specifically, HeCo\_GAN is composed of three components: the proposed HeCo, a discriminator D and a generator G. We alternatively perform the following two steps and more details are provided in the Appendix \ref{hecogan}:

(1) We utilize two view-specific embeddings to train D and G alternatively. First, we train D to identify embeddings from two views as positives and that generated from G as negatives. Then, we train G to generate samples to fool D. The two steps are alternated for some interactions to make D and G trained.

(2) We utilize a well-trained G to generate samples, which can be viewed as the new negative samples with high quality. Then, we continue to train HeCo with the newly generated and original negative samples for some epochs.

\textbf{HeCo\_MU} MixUp \cite{mixup} is proposed to efficiently improve results in supervised learning by adding arbitrary two samples to create a new one. MoCHi \cite{mochi} introduces this strategy into contrastive learning , who mixes the hard negatives to make harder negatives. Inspired by them, we bring this strategy into HIN field for the first time. We can get cosine similarities between node $i$ and nodes from $\mathbb{N}_i$ during calculating eq.\eqref{cl}, and sort them in the descending order. Then, we select first top k negative samples as the hardest negatives, and randomly add them to create new k negatives, which are involved in training. It is worth mentioning that there are no learnable parameters in this version, which is very efficient.

\begin{figure}[t]
    \centering
    \includegraphics[scale=0.25]{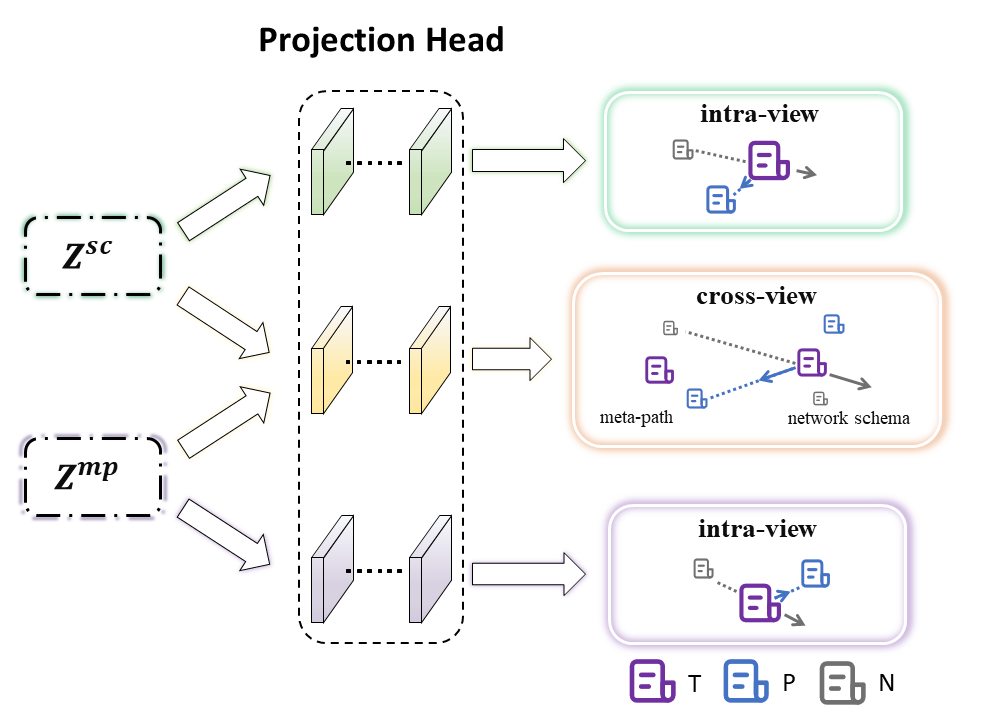}
    \caption{The partial architecture of our extended HeCo++. (T: target, P: positive, N: negative)}
    \label{extension}
\end{figure}

\section{The Extended Model: HeC\lowercase{o}++}\label{sec:ext}

In this section, we elaborate the extended model HeCo++, and the overall diagram is shown in Figure \ref{extension}. Our new extension is based on the target node embeddings $\bm{z^{sc}}$ and $\bm{z^{mp}}$ from network schema view and meta-path view. Then, we map them into three spaces by different projection heads, each of which is MLP. In one space, we remain to conduct cross-view contrast analogous to original HeCo. In the other two spaces, we conduct intra-view contrast for each view, respectively. With the joint training of intra-view and cross-view contrast, we hierarchically maintain the invariant and view-specific information of two views, and the learnt embeddings contain richer and more comprehensive information to boost final results on downstream tasks.

\subsection{Details of HeCo++}
Here, we aim to enhance the mining of view-specific structures by intra-view contrast, and meanwhile promote the process of cross-view contrast. In above section, detailed mechanism of cross-view contrastive learning has been expounded, and we continue to adopt similar technical route to define intra-view contrastive learning.




For target node i from network schema view, we project $\bm{z^{sc}_i}$ into the space for intra-view contrast under this view. This projection consists of an MLP with one hidden layer as follows:
\begin{equation}
\begin{aligned}
    \bm{z_i^{sc}}\_intra &= \bm{W^{(2)}_{sc}}\sigma\left(\bm{W^{(1)}_{sc}}\bm{z_i^{sc}}+\bm{b^{(1)}_{sc}}\right)+\bm{b^{(2)}_{sc}},
    \label{proj1}
\end{aligned}
\end{equation}
where $\sigma$ is also ELU activation function, \{$\bm{W^{(1)}_{sc}}$, $\bm{W^{(2)}_{sc}}$, $\bm{b^{(1)}_{sc}}$, $\bm{b^{(2)}_{sc}}$\} are the learnable parameters, and $\bm{z_i^{sc}}\_intra$ is the embedding of node $i$ after mapping. Similarly, embedding space for intra-view contrast under meta-path view is also designed and the similar projection process is as follows:
\begin{equation}
\begin{aligned}
    \bm{z_i^{mp}}\_intra &= \bm{W^{(2)}_{mp}}\sigma\left(\bm{W^{(1)}_{mp}}\bm{z_i^{mp}}+\bm{b^{(1)}_{mp}}\right)+\bm{b^{(2)}_{mp}},
    \label{proj2}
\end{aligned}
\end{equation}
Notice that in eq.\eqref{proj1} and eq.\eqref{proj2}, the parameters of MLP are not shared. This is because we need to utilize different embedding spaces to exploit each view respectively. Meanwhile, we maintain the dimensions of \{$\bm{z_i^{sc}}\_cross$, $\bm{z_i^{sc}}\_intra$, $\bm{z_i^{mp}}\_cross$, $\bm{z_i^{mp}}\_intra$\} are the same, to ensure each embedding space has the same complexity.

Then, we formally detail how to conduct intra-view contrastive learning, which also needs to define positive and negative samples. In cross-view contrast, for node i, $\mathbb{P}_i$ and $\mathbb{N}_i$ are defined according to the number of meta-paths between target node i and other nodes, shown in eq.\eqref{s_matrix}. This definition chooses the nodes that have highly semantic correlations with target node, so reflects its local structures. Due to the advantage of this definition, the same $\mathbb{P}_i$ and $\mathbb{N}_i$ are also adopted in intra-view contrast.

With the projected embeddings $\bm{z_i^{sc}}\_intra$ and positive and negative sets, intra-view contrast under network schema view is defined as the following form:
\begin{equation}
\begin{aligned}
    \label{loss1}
    &\mathcal{L}_i^{sc}\_intra\\
    &=-\log\frac{\sum_{j\in\mathbb{P}_i}exp\left(sim\left(\bm{z_i^{sc}}\_intra,\bm{z_j^{sc}}\_intra\right)/\tau_{sc}\right)}{\sum_{k\in\{\mathbb{P}_i\bigcup\mathbb{N}_i\}}exp\left(sim\left(\bm{z_i^{sc}}\_intra,\bm{z_k^{sc}}\_intra\right)/\tau_{sc}\right)},
\end{aligned}
\end{equation}
where $sim(\bm{u}, \bm{v})$ is also the cosine similarity of two vectors, and $\tau_{sc}$ denotes the temperature parameter special for network schema view. In the same way, we analogously define the intra-view contrast under meta-path view, given $\bm{z_i^{mp}}\_intra$:
\begin{equation}
\begin{aligned}
    \label{loss2}
    &\mathcal{L}_i^{mp}\_intra\\
    &=-\log\frac{\sum_{j\in\mathbb{P}_i}exp\left(sim\left(\bm{z_i^{mp}}\_intra,\bm{z_j^{mp}}\_intra\right)/\tau_{mp}\right)}{\sum_{k\in\{\mathbb{P}_i\bigcup\mathbb{N}_i\}}exp\left(sim\left(\bm{z_i^{mp}}\_intra,\bm{z_k^{mp}}\_intra\right)/\tau_{mp}\right)},
\end{aligned}
\end{equation}

Actually, three designed embedding spaces are separate. However, instead of training and adjusting each space separately, it is more efficient to combine training processes of these three spaces to get a joint optimization. Combined with eq.\eqref{loss}, \eqref{loss1} and \eqref{loss2}, the complete optimization object for HeCo++ is:
\begin{equation}
\begin{aligned}
    \mathcal{J} = \mathcal{L} + \frac{1}{|V|}\sum_{i\in V}\left[\lambda_1\cdot\mathcal{L}_i^{sc}\_intra + \lambda_2\cdot\mathcal{L}_i^{mp}\_intra\right],
\end{aligned}
\end{equation}
where $\lambda_1$ and $\lambda_2$ are hyper-parameters for balance. We can optimize the extended model via back propagation and learn the embeddings of nodes. The analyses of model complexity of HeCo and HeCo++ are given in Appendix~\ref{time}.

\textbf{Combination with semi-supervised signals.} As discussed above, HeCo and HeCo++ attempt to discover high-quality self-supervised signals only from data itself. However, if additional knowledge from tasks can be obtained, the training will be further well-guided~\cite{gcn}. For example, in semi-supervised node classification, a set of nodes are assigned with labels before training, whose label set is denoted as $\mathcal{Y}_L$. We design a classifier $f(\cdot)$ to predict the class of embedding of node $i$ under meta-path view obtained from eq.~\eqref{zimp}: $\bm{\hat{Y}_{i}}=softmax(f(\bm{z_i^{mp}}))$. Then, we minimize the cross-entropy loss over $\mathcal{Y}_L$:
\begin{equation}
\label{semi}
    \mathcal{L}_{semi}=-\frac{1}{|\mathcal{Y}_L|}\sum\limits_{i\in\mathcal{Y}_L}\sum\limits_{c}\bm{Y_{ic}}\ln\bm{\hat{Y}_{ic}},
\end{equation}
where $\bm{Y_{i}}$ is the ground truth one-hot label of $i$. Integrating eq.~\eqref{semi} into contrastive loss, the objective of semi-supervised version of HeCo++ is given as:
\begin{equation}
\label{semi_heco}
\mathcal{J}_{semi} = \mathcal{J} + \aleph\cdot\mathcal{L}_{semi},
\end{equation}
where $\aleph$ is the combination coefficient.

\subsection{Discussion of HeCo and HeCo++}
The proposed HeCo aims to maximize the similarity of positive samples from network schema view and meta-path view, while minimizing the similarity of negative samples. As a consequence, two view-guided encoders are trained to encode the invariant relationships between positives and negatives across two views, and so some view-specific information will be discarded. Nevertheless, the performances of different views are very dissimilar in some cases, and invariant part will be seriously affected by the worse one, while the benefit of better one is neglected.

For example, given an academic network, we only get access to the information about authors and papers, and the interactions between papers by meta-path PAP. For a revolutionary paper, it is usually composed by many famous scientists together, who usually study in various fields. So by PAP, this paper will absorb much noise from papers belonging to other fields. In this case, we need to know the scientists' major interests. For a newly published paper, we may lack the knowledge about this paper, so the interaction between it and other papers should be involved to support the inference on its label. 

Note that, in above two examples, we only consider a simple scenario that only consists of papers and authors. If we take other relations into account, the situation will be more complicated. Both of two examples show the cases where only one view is effective and the other one is invalid, while view-specific information both provides a necessary complement. Hence, besides beneficial invariant factors, it is intuitive that useful view-specific structures should be further explored. Therefore, HeCo++ is proposed to conduct hierarchical contrastive learning, including cross-view and intra-view contrasts. As the newly introduced module, intra-view contrast not only further enhances the mining of each view to learn more view-specific similarity, but also contributes to the better cross-view contrast with higher quality embeddings.

\begin{table*}[t]
  \caption{Quantitative results on node classification. (bold: best; underline: runner-up)}
  \label{fenlei}
  \resizebox{\textwidth}{!}{
  \begin{tabular}{c|c|c|cccccccc|cc}
    \hline
    Datasets & Metric & Split & GraphSAGE & GAE & Mp2vec & HERec & HetGNN & HAN & DGI & DMGI & HeCo & HeCo++\\
    \hline
    \multirow{9}{*}{ACM}&
    \multirow{3}{*}{Ma-F1}
    &20&47.13$\pm$4.7&62.72$\pm$3.1&51.91$\pm$0.9&55.13$\pm$1.5&72.11$\pm$0.9&85.66$\pm$2.1&79.27$\pm$3.8&87.86$\pm$0.2&\underline{88.56$\pm$0.8}&\textbf{89.33$\pm$0.5}\\
    &&40&55.96$\pm$6.8&61.61$\pm$3.2&62.41$\pm$0.6&61.21$\pm$0.8&72.02$\pm$0.4&87.47$\pm$1.1&80.23$\pm$3.3&86.23$\pm$0.8&\underline{87.61$\pm$0.5}&\textbf{88.70$\pm$0.7}\\
    &&60&56.59$\pm$5.7&61.67$\pm$2.9&61.13$\pm$0.4&64.35$\pm$0.8&74.33$\pm$0.6&88.41$\pm$1.1&80.03$\pm$3.3&87.97$\pm$0.4&\underline{89.04$\pm$0.5}&\textbf{89.51$\pm$0.7}\\
    \cline{2-13}
    &\multirow{3}{*}{Mi-F1}
    &20&49.72$\pm$5.5&68.02$\pm$1.9&53.13$\pm$0.9&57.47$\pm$1.5&71.89$\pm$1.1&85.11$\pm$2.2&79.63$\pm$3.5&87.60$\pm$0.8&\underline{88.13$\pm$0.8}&\textbf{88.96$\pm$0.5}\\
    &&40&60.98$\pm$3.5&66.38$\pm$1.9&64.43$\pm$0.6&62.62$\pm$0.9&74.46$\pm$0.8&87.21$\pm$1.2&80.41$\pm$3.0&86.02$\pm$0.9&\underline{87.45$\pm$0.5}&\textbf{88.40$\pm$0.8}\\
    &&60&60.72$\pm$4.3&65.71$\pm$2.2&62.72$\pm$0.3&65.15$\pm$0.9&76.08$\pm$0.7&88.10$\pm$1.2&80.15$\pm$3.2&87.82$\pm$0.5&\underline{88.71$\pm$0.5}&\textbf{89.30$\pm$0.7}\\
    \cline{2-13}
    &\multirow{3}{*}{AUC}
    &20&65.88$\pm$3.7&79.50$\pm$2.4&71.66$\pm$0.7&75.44$\pm$1.3&84.36$\pm$1.0&93.47$\pm$1.5&91.47$\pm$2.3&\underline{96.72$\pm$0.3}&96.49$\pm$0.3&\textbf{97.25$\pm$0.2}\\
    &&40&71.06$\pm$5.2&79.14$\pm$2.5&80.48$\pm$0.4&79.84$\pm$0.5&85.01$\pm$0.6&94.84$\pm$0.9&91.52$\pm$2.3&96.35$\pm$0.3&\underline{96.40$\pm$0.4}&\textbf{97.08$\pm$0.2}\\
    &&60&70.45$\pm$6.2&77.90$\pm$2.8&79.33$\pm$0.4&81.64$\pm$0.7&87.64$\pm$0.7&94.68$\pm$1.4&91.41$\pm$1.9&\underline{96.79$\pm$0.2}&96.55$\pm$0.3&\textbf{97.50$\pm$0.2}\\
    \hline
    \multirow{9}{*}{DBLP}&
    \multirow{3}{*}{Ma-F1}
    &20&71.97$\pm$8.4&90.90$\pm$0.1&88.98$\pm$0.2&89.57$\pm$0.4&89.51$\pm$1.1&89.31$\pm$0.9&87.93$\pm$2.4&89.94$\pm$0.4&\underline{91.28$\pm$0.2}&\textbf{91.40$\pm$0.2}\\
    &&40&73.69$\pm$8.4&89.60$\pm$0.3&88.68$\pm$0.2&89.73$\pm$0.4&88.61$\pm$0.8&88.87$\pm$1.0&88.62$\pm$0.6&89.25$\pm$0.4&\underline{90.34$\pm$0.3}&\textbf{90.56$\pm$0.2}\\
    &&60&73.86$\pm$8.1&90.08$\pm$0.2&90.25$\pm$0.1&90.18$\pm$0.3&89.56$\pm$0.5&89.20$\pm$0.8&89.19$\pm$0.9&89.46$\pm$0.6&\underline{90.64$\pm$0.3}&\textbf{91.01$\pm$0.3}\\
    \cline{2-13}
    &\multirow{3}{*}{Mi-F1}
    &20&71.44$\pm$8.7&91.55$\pm$0.1&89.67$\pm$0.1&90.24$\pm$0.4&90.11$\pm$1.0&90.16$\pm$0.9&88.72$\pm$2.6&90.78$\pm$0.3&\underline{91.97$\pm$0.2}&\textbf{92.03$\pm$0.1}\\
    &&40&73.61$\pm$8.6&90.00$\pm$0.3&89.14$\pm$0.2&90.15$\pm$0.4&89.03$\pm$0.7&89.47$\pm$0.9&89.22$\pm$0.5&89.92$\pm$0.4&\underline{90.76$\pm$0.3}&\textbf{90.87$\pm$0.2}\\
    &&60&74.05$\pm$8.3&90.95$\pm$0.2&91.17$\pm$0.1&91.01$\pm$0.3&90.43$\pm$0.6&90.34$\pm$0.8&90.35$\pm$0.8&90.66$\pm$0.5&\underline{91.59$\pm$0.2}&\textbf{91.86$\pm$0.2}\\
    \cline{2-13}
    &\multirow{3}{*}{AUC}
    &20&90.59$\pm$4.3&98.15$\pm$0.1&97.69$\pm$0.0&98.21$\pm$0.2&97.96$\pm$0.4&98.07$\pm$0.6&96.99$\pm$1.4&97.75$\pm$0.3&\underline{98.32$\pm$0.1}&\textbf{98.39$\pm$0.1}\\
    &&40&91.42$\pm$4.0&97.85$\pm$0.1&97.08$\pm$0.0&97.93$\pm$0.1&97.70$\pm$0.3&97.48$\pm$0.6&97.12$\pm$0.4&97.23$\pm$0.2&\underline{98.06$\pm$0.1}&\textbf{98.17$\pm$0.1}\\
    &&60&91.73$\pm$3.8&98.37$\pm$0.1&98.00$\pm$0.0&98.49$\pm$0.1&97.97$\pm$0.2&97.96$\pm$0.5&97.76$\pm$0.5&97.72$\pm$0.4&\underline{98.59$\pm$0.1}&\textbf{98.62$\pm$0.1}\\
    \hline
    \multirow{9}{*}{Freebase}&
    \multirow{3}{*}{Ma-F1}
    &20&45.14$\pm$4.5&53.81$\pm$0.6&53.96$\pm$0.7&55.78$\pm$0.5&52.72$\pm$1.0&53.16$\pm$2.8&54.90$\pm$0.7&55.79$\pm$0.9&\underline{59.23$\pm$0.7}&\textbf{59.87$\pm$1.0}\\
    &&40&44.88$\pm$4.1&52.44$\pm$2.3&57.80$\pm$1.1&59.28$\pm$0.6&48.57$\pm$0.5&59.63$\pm$2.3&53.40$\pm$1.4&49.88$\pm$1.9&\underline{61.19$\pm$0.6}&\textbf{61.33$\pm$0.5}\\
    &&60&45.16$\pm$3.1&50.65$\pm$0.4&55.94$\pm$0.7&56.50$\pm$0.4&52.37$\pm$0.8&56.77$\pm$1.7&53.81$\pm$1.1&52.10$\pm$0.7&\underline{60.13$\pm$1.3}&\textbf{60.86$\pm$1.0}\\
    \cline{2-13}
    &\multirow{3}{*}{Mi-F1}
    &20&54.83$\pm$3.0&55.20$\pm$0.7&56.23$\pm$0.8&57.92$\pm$0.5&56.85$\pm$0.9&57.24$\pm$3.2&58.16$\pm$0.9&58.26$\pm$0.9&\underline{61.72$\pm$0.6}&\textbf{62.29$\pm$1.9}\\
    &&40&57.08$\pm$3.2&56.05$\pm$2.0&61.01$\pm$1.3&62.71$\pm$0.7&53.96$\pm$1.1&63.74$\pm$2.7&57.82$\pm$0.8&54.28$\pm$1.6&\underline{64.03$\pm$0.7}&\textbf{64.27$\pm$0.5}\\
    &&60&55.92$\pm$3.2&53.85$\pm$0.4&58.74$\pm$0.8&58.57$\pm$0.5&56.84$\pm$0.7&61.06$\pm$2.0&57.96$\pm$0.7&56.69$\pm$1.2&\underline{63.61$\pm$1.6}&\textbf{64.15$\pm$0.9}\\
    \cline{2-13}
    &\multirow{3}{*}{AUC}
    &20&67.63$\pm$5.0&73.03$\pm$0.7&71.78$\pm$0.7&73.89$\pm$0.4&70.84$\pm$0.7&73.26$\pm$2.1&72.80$\pm$0.6&73.19$\pm$1.2&\underline{76.22$\pm$0.8}&\textbf{76.68$\pm$0.7}\\
    &&40&66.42$\pm$4.7&74.05$\pm$0.9&75.51$\pm$0.8&76.08$\pm$0.4&69.48$\pm$0.2&77.74$\pm$1.2&72.97$\pm$1.1&70.77$\pm$1.6&\underline{78.44$\pm$0.5}&\textbf{79.51$\pm$0.3}\\
    &&60&66.78$\pm$3.5&71.75$\pm$0.4&74.78$\pm$0.4&74.89$\pm$0.4&71.01$\pm$0.5&75.69$\pm$1.5&73.32$\pm$0.9&73.17$\pm$1.4&\underline{78.04$\pm$0.4}&\textbf{78.27$\pm$0.7}\\
    \hline
    \multirow{9}{*}{AMiner}&
    \multirow{3}{*}{Ma-F1}
    &20&42.46$\pm$2.5&60.22$\pm$2.0&54.78$\pm$0.5&58.32$\pm$1.1&50.06$\pm$0.9&56.07$\pm$3.2&51.61$\pm$3.2&59.50$\pm$2.1&\underline{71.38$\pm$1.1}&\textbf{72.28$\pm$1.4}\\
    &&40&45.77$\pm$1.5&65.66$\pm$1.5&64.77$\pm$0.5&64.50$\pm$0.7&58.97$\pm$0.9&63.85$\pm$1.5&54.72$\pm$2.6&61.92$\pm$2.1&\underline{73.75$\pm$0.5}&\textbf{75.35$\pm$0.5}\\
    &&60&44.91$\pm$2.0&63.74$\pm$1.6&60.65$\pm$0.3&65.53$\pm$0.7&57.34$\pm$1.4&62.02$\pm$1.2&55.45$\pm$2.4&61.15$\pm$2.5&\underline{75.80$\pm$1.8}&\textbf{76.28$\pm$0.6}\\
    \cline{2-13}
    &\multirow{3}{*}{Mi-F1}
    &20&49.68$\pm$3.1&65.78$\pm$2.9&60.82$\pm$0.4&63.64$\pm$1.1&61.49$\pm$2.5&68.86$\pm$4.6&62.39$\pm$3.9&63.93$\pm$3.3&\underline{78.81$\pm$1.3}&\textbf{80.00$\pm$1.0}\\
    &&40&52.10$\pm$2.2&71.34$\pm$1.8&69.66$\pm$0.6&71.57$\pm$0.7&68.47$\pm$2.2&76.89$\pm$1.6&63.87$\pm$2.9&63.60$\pm$2.5&\underline{80.53$\pm$0.7}&\textbf{82.01$\pm$0.6}\\
    &&60&51.36$\pm$2.2&67.70$\pm$1.9&63.92$\pm$0.5&69.76$\pm$0.8&65.61$\pm$2.2&74.73$\pm$1.4&63.10$\pm$3.0&62.51$\pm$2.6&\underline{82.46$\pm$1.4}&\textbf{82.80$\pm$0.7}\\
    \cline{2-13}
    &\multirow{3}{*}{AUC}
    &20&70.86$\pm$2.5&85.39$\pm$1.0&81.22$\pm$0.3&83.35$\pm$0.5&77.96$\pm$1.4&78.92$\pm$2.3&75.89$\pm$2.2&85.34$\pm$0.9&\underline{90.82$\pm$0.6}&\textbf{91.59$\pm$0.6}\\
    &&40&74.44$\pm$1.3&88.29$\pm$1.0&88.82$\pm$0.2&88.70$\pm$0.4&83.14$\pm$1.6&80.72$\pm$2.1&77.86$\pm$2.1&88.02$\pm$1.3&\underline{92.11$\pm$0.6}&\textbf{93.46$\pm$0.2}\\
    &&60&74.16$\pm$1.3&86.92$\pm$0.8&85.57$\pm$0.2&87.74$\pm$0.5&84.77$\pm$0.9&80.39$\pm$1.5&77.21$\pm$1.4&86.20$\pm$1.7&\underline{92.40$\pm$0.7}&\textbf{93.68$\pm$0.3}\\
    \hline
  \end{tabular}
  }
\end{table*}

\section{EXPERIMENTS}\label{sec:experiment}
\subsection{Experimental Setup}
\textbf{Datasets}\quad We employ four open heterogeneous datasets, including ACM~\cite{nshe}, DBLP~\cite{magnn}, Freebase~\cite{freebase} and AMiner~\cite{hegan}. The basic information about datasets is summarized in Appendix~\ref{Datasetss}.

\noindent\textbf{Baselines}\quad We compare the proposed HeCo with three categories of baselines: (1) unsupervised homogeneous methods GraphSAGE~\cite{GraphSAGE}, GAE~\cite{gae}, DGI~\cite{dgi}, (2) unsupervised heterogeneous methods Mp2vec~\cite{mp2vec}, HERec~\cite{herec}, HetGNN~\cite{hetegnn}, DMGI~\cite{dmgi}, and (3) a semi-supervised heterogeneous method HAN~\cite{han}. Actually, to make results reliable, we use official codes for all baselines.

\noindent\textbf{Implementation Details}\quad For random walk-based methods (i.e., Mp2vec, HERec, HetGNN), we set the number of walks per node to 40, the walk length to 100 and the window size to 5. For GraphSAGE, GAE, Mp2vec, HERec and DGI, we test all the meta-paths for them and report the best performance. For HAN, the number of heads is set as 8, and the dimension of each head is 8.  The patience is set as 100 for fully training. For DMGI, we set patience for early stop as 20, set the weight of self-loop as 3, and set dropout as 0.5. For HeteGNN, we replace NN-1 with one MLP to map raw feature into node embedding, considering that only one raw feature for each node is provided without other information (e.g., text or image). In terms of other parameters, we follow the settings in their original papers.

For the proposed HeCo, we use Glorot initialization \cite{glorot2010understanding}, and Adam \cite{adam} optimizer. We search on learning rate from 1e-4 to 5e-3, and tune patience for early stopping from 5 to 50. For dropout rate, we test ranging from 0.1 to 0.5 with step 0.05, and $\tau$ is tuned from 0.5 to 0.9 with step 0.05. For meta-path view, we use one-layer GCN for every meta-path. For network schema view, we only consider interactions between nodes of target type and their one-hop neighbors of other types, and utilize one-head attention to fuse them. For the extended HeCo++, we search the best values of $\lambda_1$ and $\lambda_2$ from 1e2 to 1e-9, and further fine tune other hyperparameters based on the settings for HeCo.

For all methods, we set the embedding dimension as 64 and randomly run 10 times and report the average results. For every dataset, we only use original attributes of target nodes, and assign one-hot id vectors to nodes of other types, if they are needed. This is because most of datasets only provide features for target type of nodes, so to stay the same, we assign other type of nodes with one-hot features.

\begin{figure*}[t]
\centering
\hspace{-8mm}
\subfigure[Mp2vec]{
\label{Mp2vec_acm}
\includegraphics[scale=0.25]{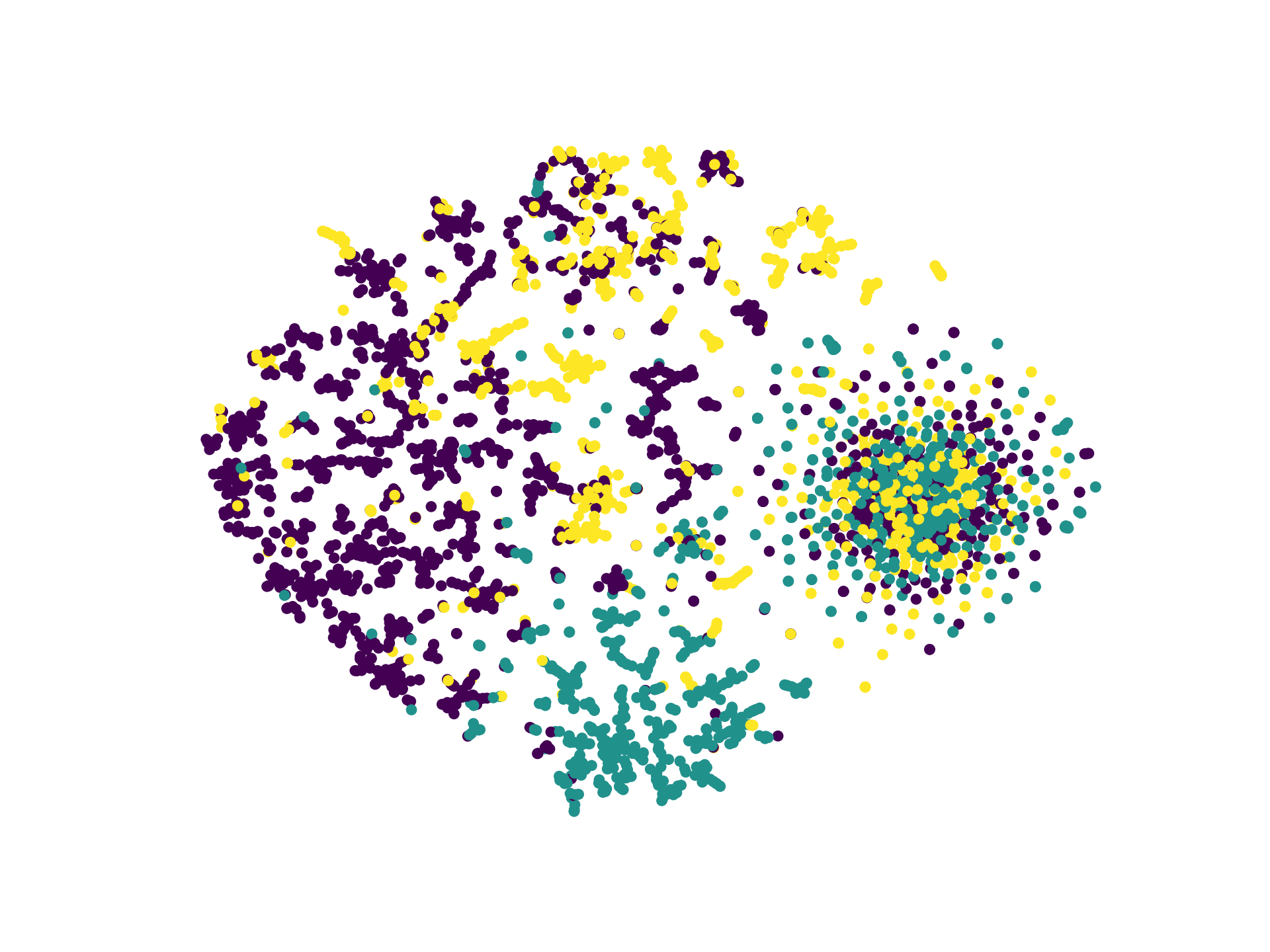}
}\hspace{-8mm}
\subfigure[DGI]{
\label{DGI_acm}
\includegraphics[scale=0.25]{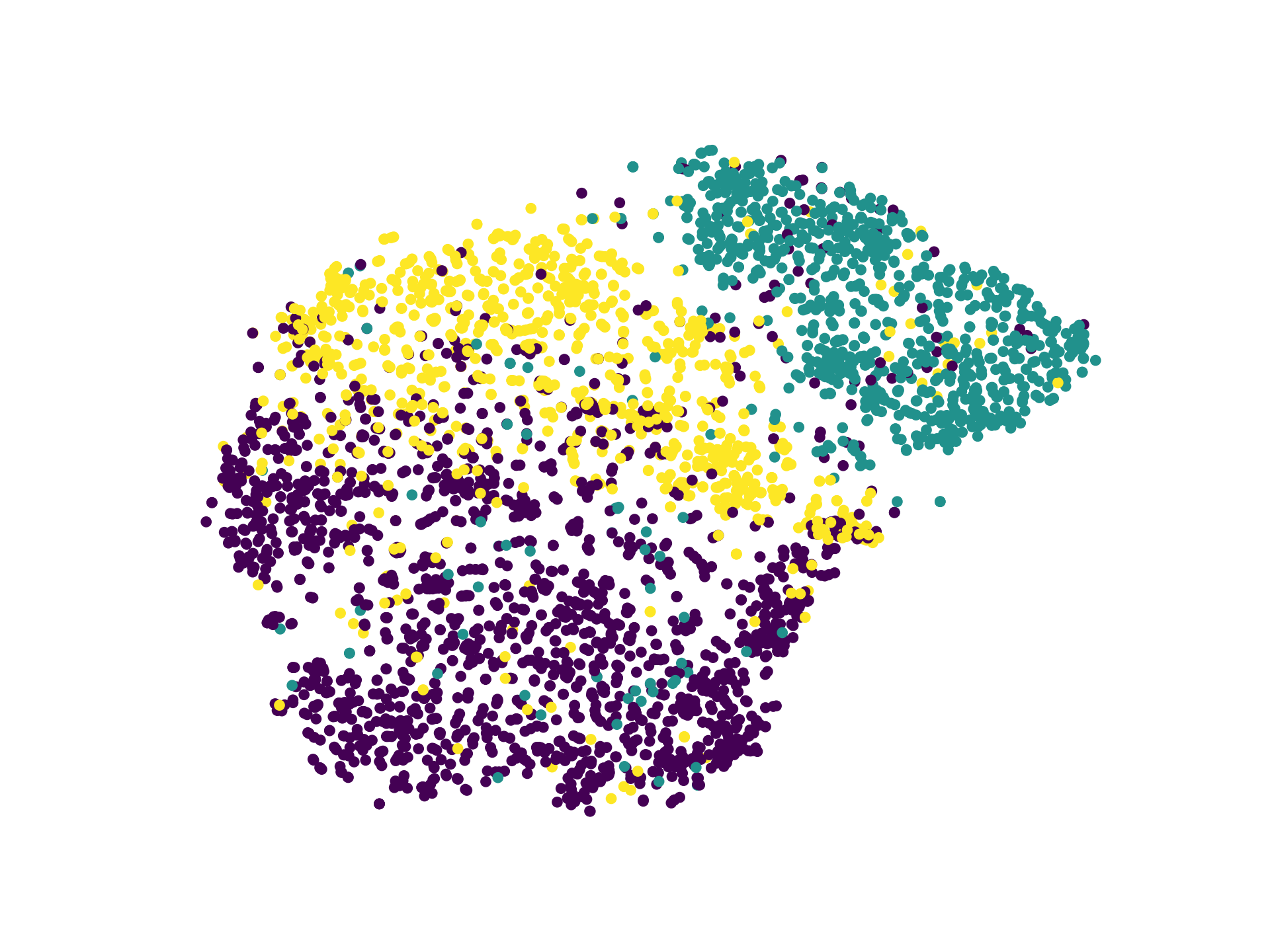}}\hspace{-8mm}
\subfigure[DMGI]{
\label{DMGI_v_acm}
\includegraphics[scale=0.25]{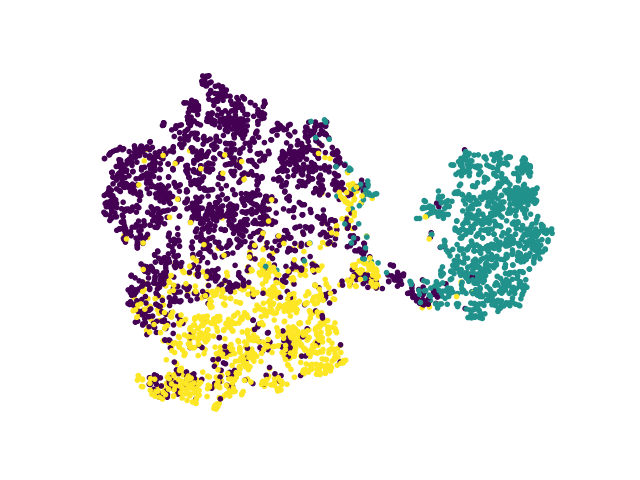}}\hspace{-8mm}
\subfigure[HeCo]{
\label{HeCo_v_acm}
\includegraphics[scale=0.25]{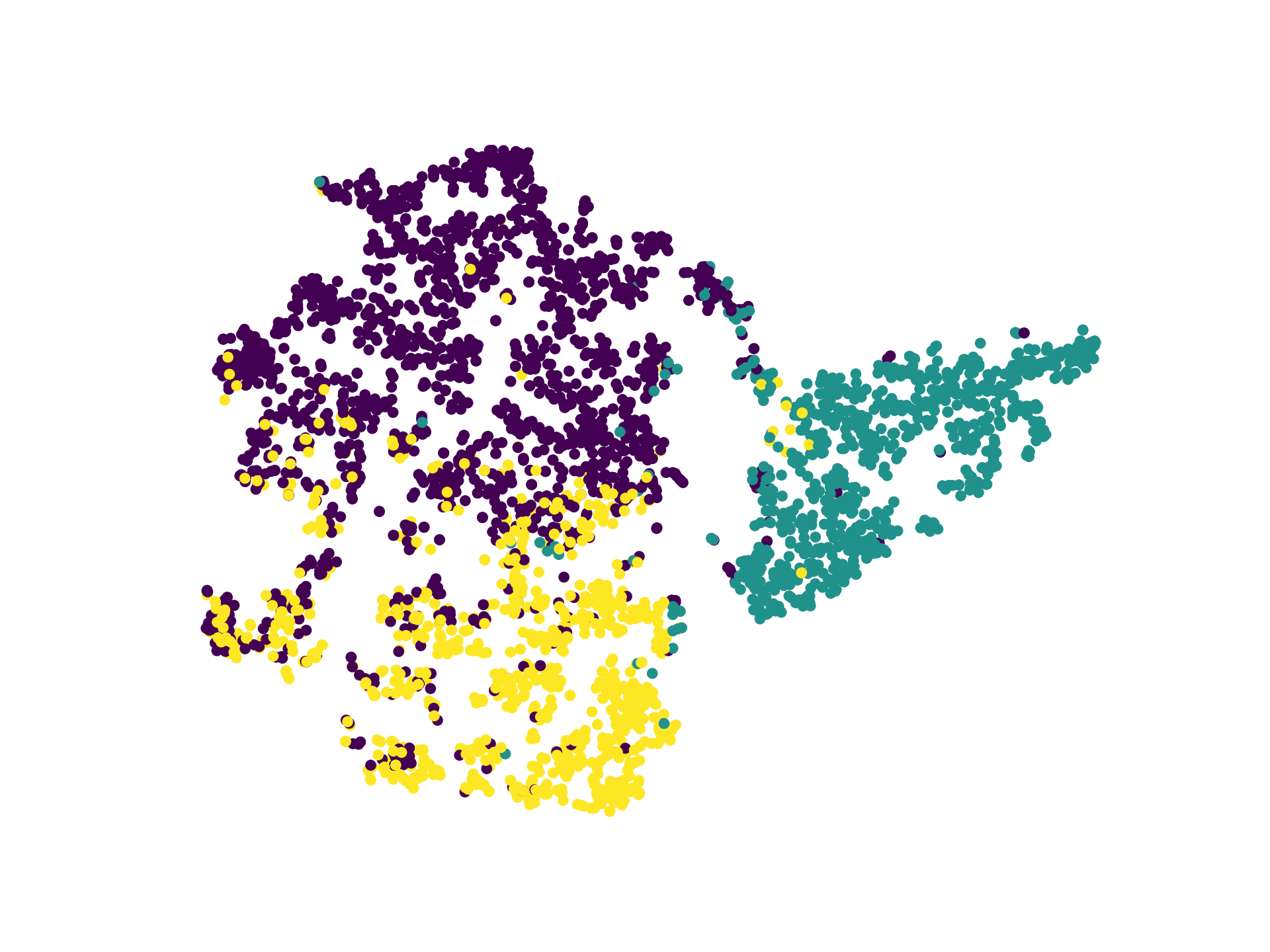}}\hspace{-8mm}
\subfigure[HeCo++]{
\label{HeCo++_v_acm}
\includegraphics[scale=0.25]{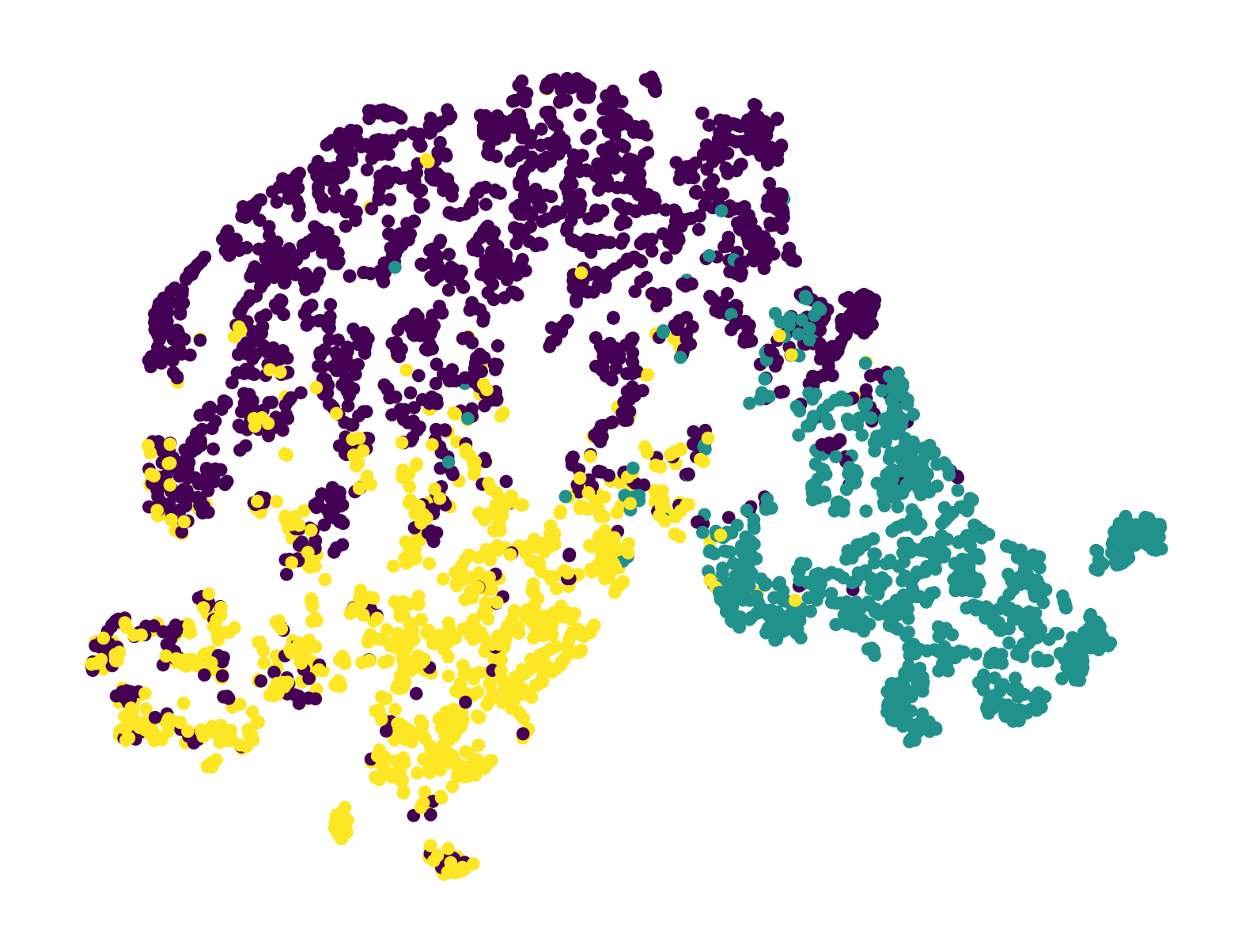}
}
\caption{Visualization of the learned node embedding on ACM. The Silhouette scores for (a) (b) (c) (d) (e) are 0.0292, 0.1862, 0.3015, \underline{0.3642} and \textbf{0.3885}, respectively.}
\label{visiulization_acm}
\end{figure*}

\begin{figure*}[t]
\centering
    \subfigure[ACM]{
        \label{xiao_acm40}
        \centering
        \includegraphics[scale=0.25]{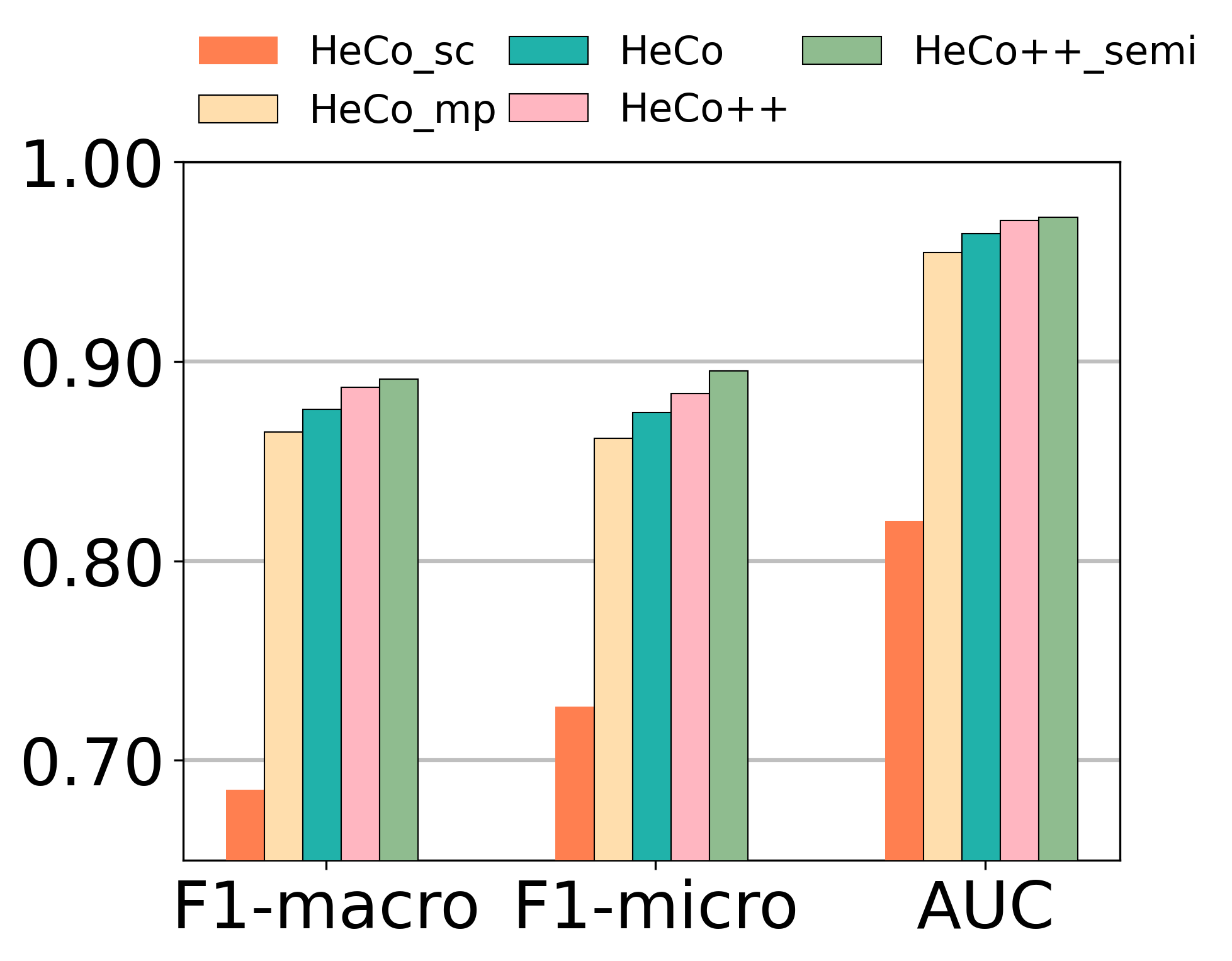}
        }
    \subfigure[DBLP]{
        \label{xiao_dblp40}
        \centering
        \includegraphics[scale=0.25]{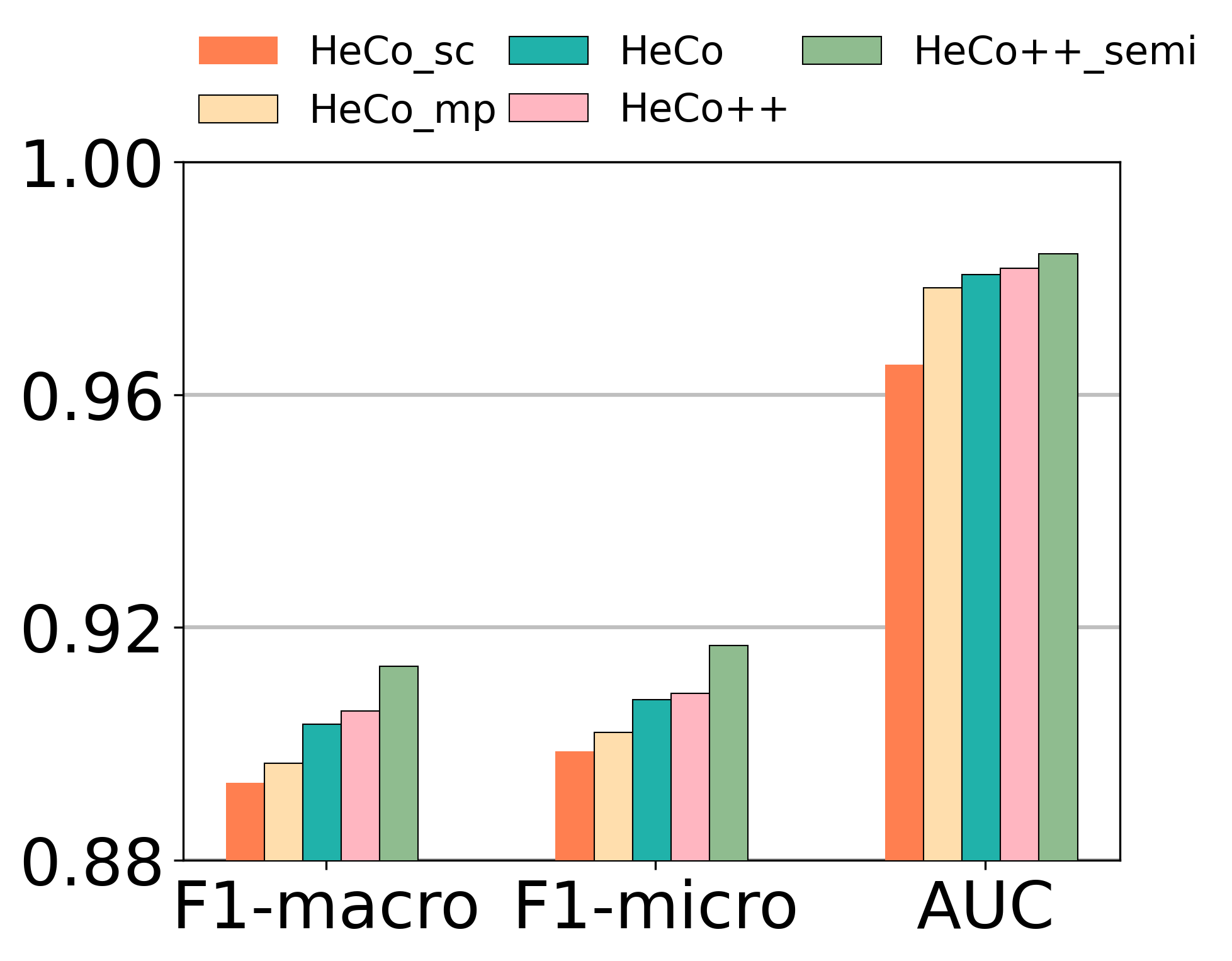}
        }
    \subfigure[Freebase]{
        \label{xiao_fb40}
        \centering
        \includegraphics[scale=0.25]{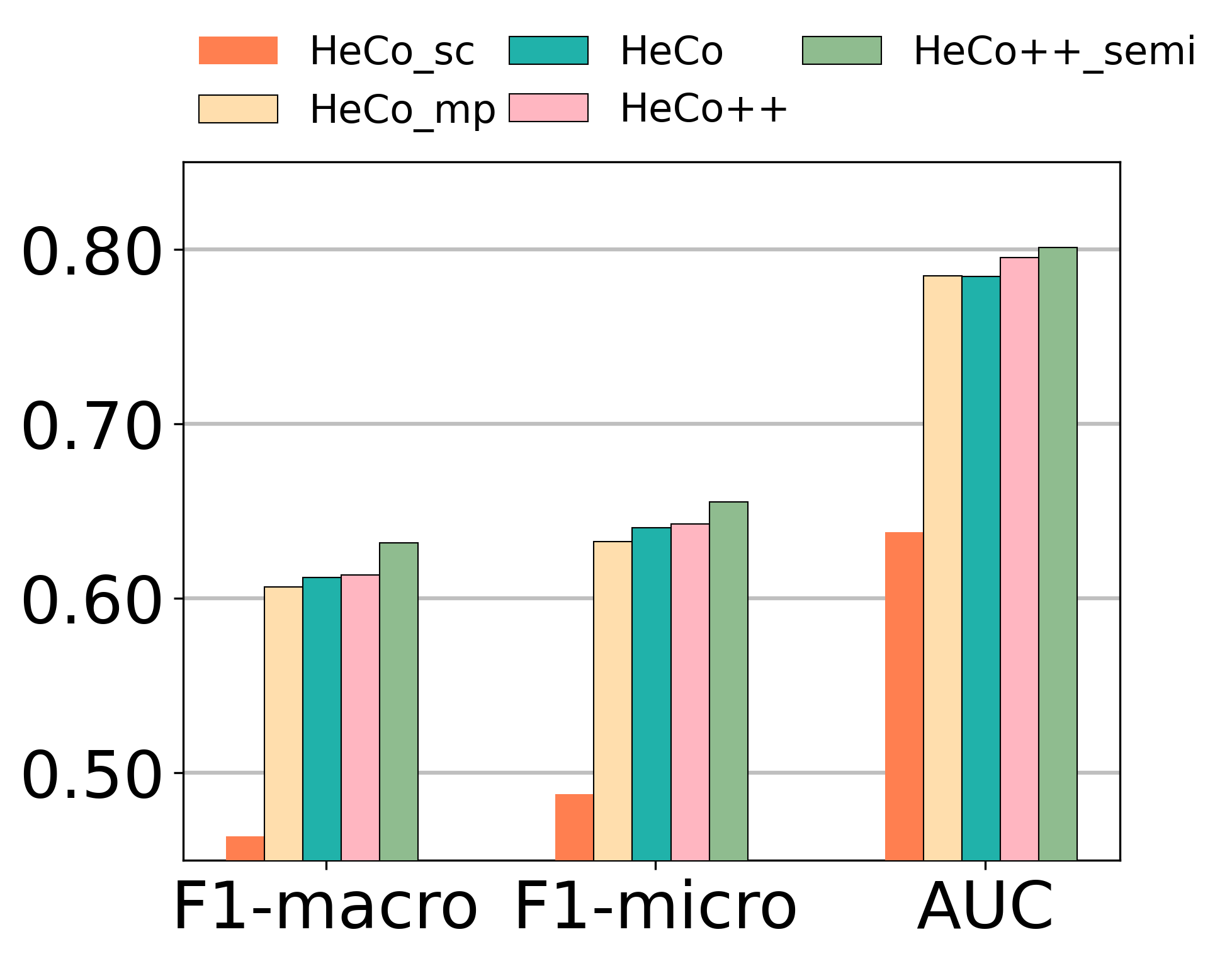}
        }
    \subfigure[AMiner]{
        \label{xiao_aminer40}
        \centering
        \includegraphics[scale=0.25]{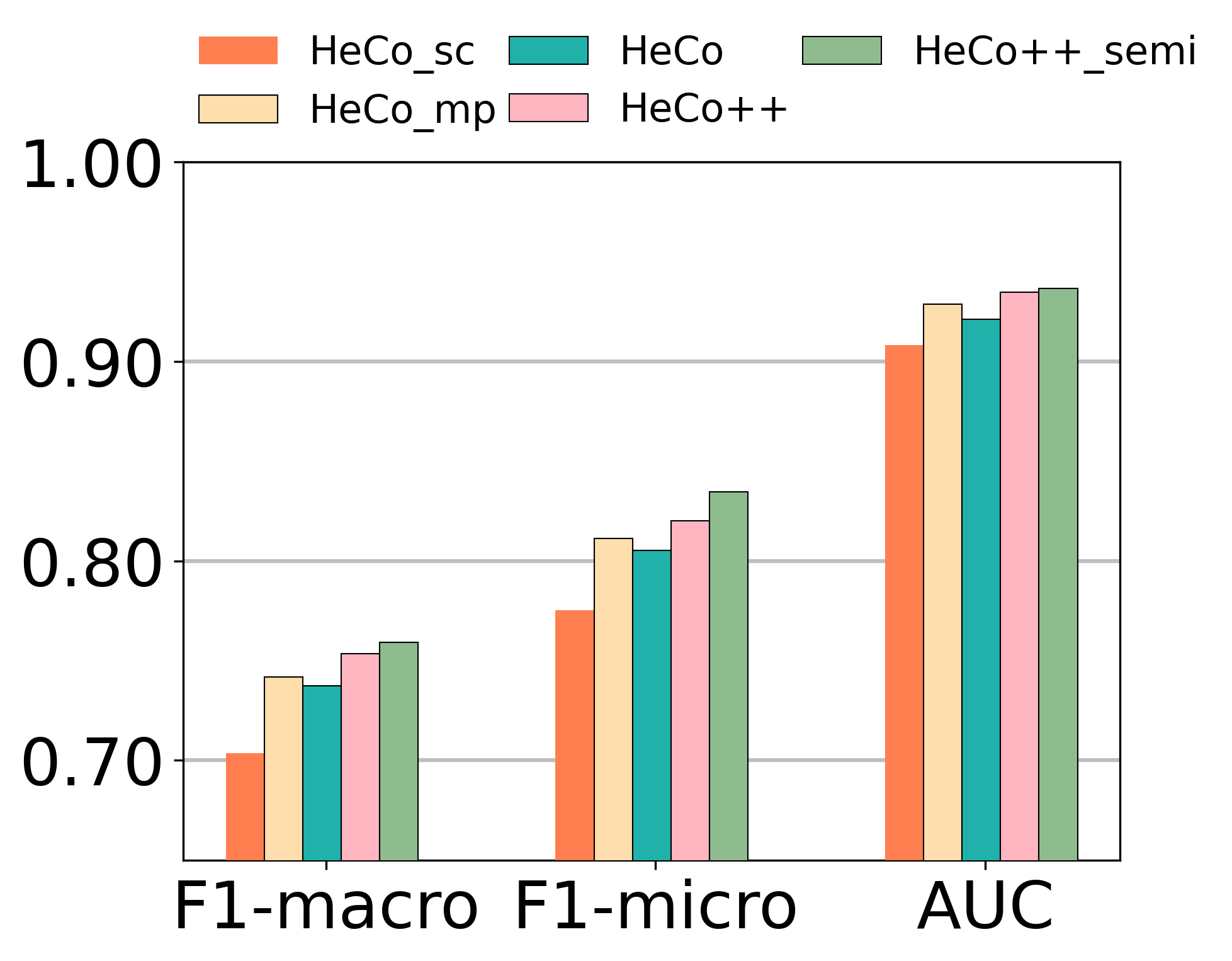}
        }
\caption{The comparison of HeCo and its variants.}
\label{variants}
\end{figure*}

\subsection{Node Classification}
The learned embeddings of nodes are used to train a linear classifier. To more comprehensively evaluate our model, we choose 20, 40, 60 labeled nodes per class as training set, and select 1000 nodes as validation and 1000 as test set respectively, for each dataset. We follow DMGI that reports the test performance when performance on validation gives the best result. We use common evaluation metrics, including Macro-F1, Micro-F1 and AUC. The results are reported in Table \ref{fenlei}. As can be seen, the proposed HeCo++ and HeCo generally achieve the best and the runner-up performance on all datasets and all splits, even compared with HAN, a semi-supervised method. Besides, the extended model HeCo++ always shows improvements against HeCo on all datasets, which means intra-view contrast is effective. We can also see that HeCo outperforms DMGI in most cases, while DMGI is even worse than other baselines with some settings, indicating that single-view is noisy and incomplete. So, performing contrastive learning across views is effective. Moreover, even HAN utilizes the label information, HeCo performs better than it in all cases. This well indicates the great potential of cross-view contrastive learning.

\subsection{Node Clustering}
In this task, we utilize K-means algorithm to the learned embeddings of all nodes and adopt normalized mutual information (NMI) and adjusted rand index (ARI) to assess the quality of the clustering results. To alleviate the instability due to different initial values, we repeat the process for 10 times, and report average results, shown in Table \ref{julei}. Notice that, we do not compare with HAN, because it has known the labels of training set and been guided by validation.
As we can see, HeCo++ and HeCo consistently achieve the best and the runner-up results on all datasets, which proves the effectiveness of HeCo++ and HeCo on this task. HeCo++ also outperforms HeCo, because of the further mining of view-specific information.
Moreover, HeCo outperforms DMGI in all cases, further suggesting the importance of contrasting across views. 
\begin{table}[h]
  \caption{Quantitative results on node clustering. (bold: best; underline: runner-up)}
  \label{julei}
  \resizebox{0.47\textwidth}{!}{
  \begin{tabular}{c|cc|cc|cc|cc}
    \hline
    Datasets & \multicolumn{2}{c|}{ACM} & \multicolumn{2}{c|}{DBLP} & \multicolumn{2}{c|}{Freebase} & \multicolumn{2}{c}{AMiner} \\
    \hline
    Metrics & NMI & ARI & NMI & ARI & NMI & ARI & NMI & ARI\\
    \hline
    GraphSage&29.20&27.72&51.50&36.40&9.05&10.49&15.74&10.10\\
    GAE&27.42&24.49&72.59&77.31&19.03&14.10&28.58&20.90\\
    Mp2vec&48.43&34.65&73.55&77.70&16.47&17.32&30.80&25.26\\
    HERec&47.54&35.67&70.21&73.99&19.76&19.36&27.82&20.16\\
    HetGNN&41.53&34.81&69.79&75.34&12.25&15.01&21.46&26.60\\
    DGI&51.73&41.16&59.23&61.85&18.34&11.29&22.06&15.93\\
    DMGI&51.66&46.64&70.06&75.46&16.98&16.91&19.24&20.09\\
    \hline
    HeCo&\underline{56.87}&\underline{56.94}&\underline{74.51}&\underline{80.17}&\underline{20.38}&\underline{20.98}&\underline{32.26}&\underline{28.64}\\
    HeCo++&\textbf{60.82}&\textbf{60.09}&\textbf{75.39}&\textbf{81.20}&\textbf{20.62}&\textbf{21.88}&\textbf{38.07}&\textbf{36.44}\\
    \hline
  \end{tabular}}
\end{table}

\subsection{Visualization}
To provide a more intuitive evaluation, we conduct embedding visualization on ACM dataset. We plot learnt embeddings of Mp2vec, DGI, DMGI, HeCo and HeCo++ using t-SNE, and the results are shown in Figure \ref{visiulization_acm}, where different colors mean different labels.

We can see that Mp2vec and DGI present blurred boundaries between different types of nodes, because they cannot fuse all kinds of semantics. For DMGI, nodes are still mixed to some degree. Apparently, the proposed HeCo++ and HeCo correctly separate different nodes with relatively clear boundaries. Then, we find that the embeddings of HeCo++ are more separable than that of HeCo, implying that intra-view contrast will make embeddings more discriminative. To give a quantitative analysis, we calculate the silhouette scores of different clusters, and HeCo++ and HeCo also outperform other three methods, demonstrating the effectiveness of the proposed models again. We further visualize node embeddings in Appendix~\ref{Visualization on AMiner}.

\subsection{Variant Analysis}
In this section, we design two variants of proposed HeCo: HeCo\_sc and HeCo\_mp. For variant HeCo\_sc, nodes are only encoded in network schema view, and the embeddings of corresponding positive and negatives samples also come from network schema view, rather than meta-path view. For variant HeCo\_mp, the practice is similar, where we only focus on meta-path view and neglect network schema view. We conduct comparison between them and two proposed models, HeCo++ and HeCo, on ACM, DBLP, Freebase and AMiner, and report the results of 40 labeled nodes per class. Besides, we also test the semi-supervised version of HeCo++, where the labeled nodes are those utilized to train linear classifier. All results are given in Figure \ref{variants}.

From Figure \ref{variants}, some conclusions are got as follows: (1) The results of HeCo are better than two variants in most cases, indicating the effectiveness and necessity of the cross-view contrastive learning. (2) HeCo++ constantly beats HeCo, indicating the importance of combining the intra-view contrast and cross-view contrast. (3) The performance of HeCo\_mp is also very competitive, especially in AMiner dataset, which demonstrates that meta-path is a powerful tool to handle the heterogeneity. (4) HeCo\_sc is the worst one, which makes us realize the necessity of involving the features of target nodes into embeddings if contrast is done only in a single view. (5) HeCo++\_semi outperforms HeCo++ and the other variants, where the significance of labels is adequately displayed.

\subsection{Collaborative Trend Analysis}
\begin{figure}[h]
\centering
\subfigure[ACM]{
\label{xuan_acm}
\centering
\includegraphics[scale=0.15]{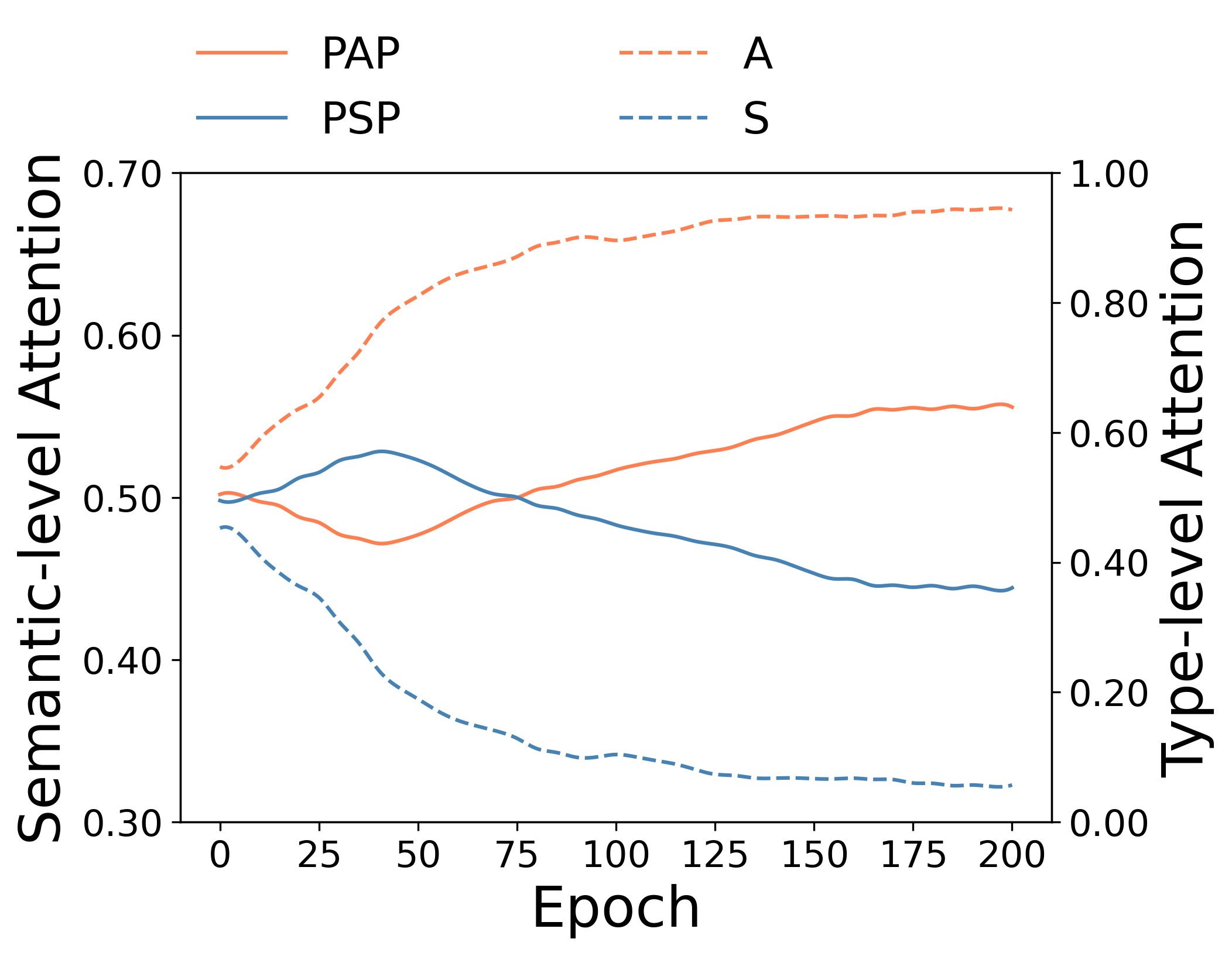}}
\subfigure[Freebase]{
\label{xuan_aminer}
\centering
\includegraphics[scale=0.15]{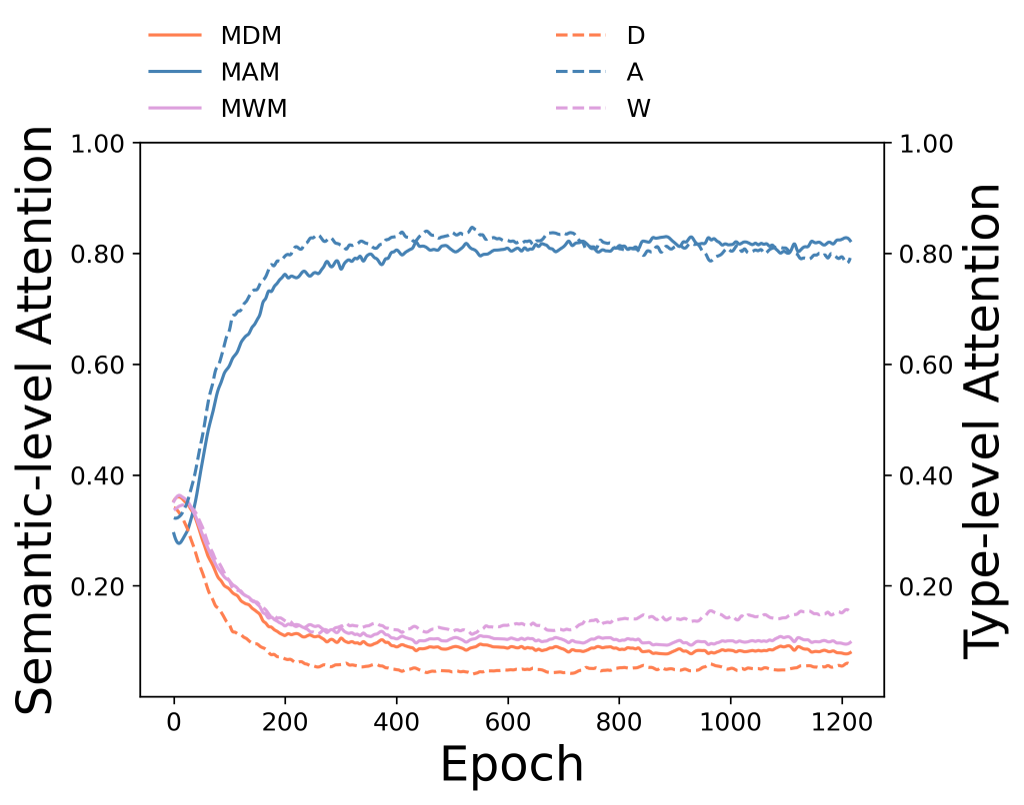}}
\subfigure[AMiner]{
\label{xuan_aminer}
\centering
\includegraphics[scale=0.15]{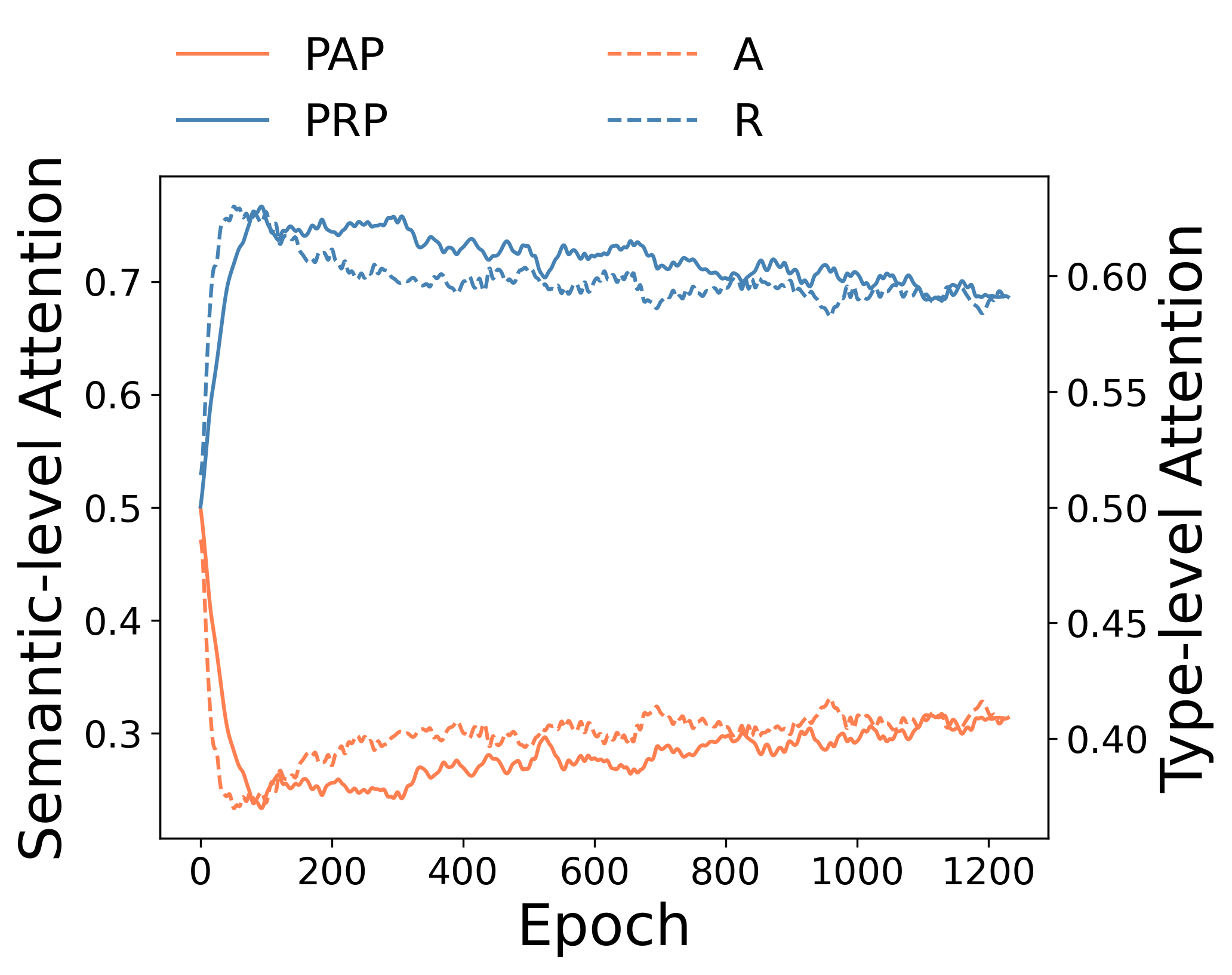}}
\caption{The collaborative changing trends of attentions in two views w.r.t epochs.}
\label{att}
\end{figure}

One salient property of HeCo is the cross-view collaborative mechanism, i.e., HeCo employs the network schema and meta-path views to collaboratively supervise each other to learn the embeddings. In this section, we examine the changing trends of type-level attention $\beta_\Phi$ in network schema view and semantic-level attention $\beta_\mathcal{P}$ in meta-path view w.r.t epochs, and the results are plotted in Figure \ref{att}. 

For ACM, Freebase and AMiner, the changing trends of two views are collaborative and consistent. Specifically, for ACM, $\beta_\Phi$ of type A is higher than type S, and $\beta_\mathcal{P}$ of meta-path PAP also exceeds that of PSP. For Freebase, type A and meta-path MAM are more preferred than other two node types and meta-paths. For AMiner, type R and meta-path PRP are more important in two views respectively. This indicates that network schema view and meta-path view adapt for each other during training and collaboratively optimize by contrasting each other.

\subsection{Model Extension Analysis}
In this section, we examine results of our extensions. As is shown above, DMGI is a rather competitive method on ACM. So, we compare our two extensions with base model and DMGI on classification and clustering tasks using ACM. The results is shown in Table \ref{kuozhan}.

From the table, we can see that the proposed two versions generally outperform base model and DMGI, especially the version of HeCo\_GAN, which improves the results with a clear margin. As expected, GAN based method can generate harder negatives that are closer to positive distributions. HeCo\_MU is the second best in most cases. The better performance of HeCo\_GAN and HeCo\_MU indicates that more and high-quality negative samples are useful for contrastive learning in general.

\begin{table}[ht]
  \caption{Evaluation of two ways to generate harder negative samples on various tasks using ACM (Task 1: Classification; Task 2: Clustering).}
  \centering
  \label{kuozhan}
  \resizebox{0.4\textwidth}{!}{
  \begin{tabular}{c|c|cccc}
       \hline
       \multicolumn{2}{c|}{Task 1}&DMGI&HeCo&HeCo\_MU&HeCo\_GAN\\
       \hline
       \multirow{3}{*}{Ma}
       &20&87.86$\pm$0.2&88.56$\pm$0.8&88.65$\pm$0.8&\textbf{89.22$\pm$1.1}\\
       &40&86.23$\pm$0.8&87.61$\pm$0.5&87.78$\pm$1.7&\textbf{88.61$\pm$1.6}\\
       &60&87.97$\pm$0.4&89.04$\pm$0.5&\textbf{89.83$\pm$0.5}&89.55$\pm$1.3\\
       \hline
       \multirow{3}{*}{Mi}
       &20&87.60$\pm$0.8&88.13$\pm$0.8&88.39$\pm$0.9&\textbf{88.92$\pm$0.9}\\
       &40&86.02$\pm$0.9&87.45$\pm$0.5&87.66$\pm$1.7&\textbf{88.48$\pm$1.7}\\
       &60&87.82$\pm$0.5&88.71$\pm$0.5&\textbf{89.52$\pm$0.5}&89.29$\pm$1.4\\
       \hline
       \multirow{3}{*}{AUC}
       &20&96.72$\pm$0.3&96.49$\pm$0.3&96.38$\pm$0.5&\textbf{96.91$\pm$0.3}\\
       &40&96.35$\pm$0.3&96.40$\pm$0.4&96.54$\pm$0.5&\textbf{97.13$\pm$0.5}\\
       &60&96.79$\pm$0.2&96.55$\pm$0.3&96.67$\pm$0.7&\textbf{97.12$\pm$0.4}\\
       \hline
       \multicolumn{2}{c|}{Task 2}&DMGI&HeCo&HeCo\_MU&HeCo\_GAN\\
       \hline
       \multicolumn{2}{c|}{NMI}
       &51.66&56.87&58.17&\textbf{59.34}\\
       \multicolumn{2}{c|}{ARI}
       &46.64&56.94&59.41&\textbf{61.48}\\
       \hline
  \end{tabular}}
\end{table}

\subsection{Analysis of Hyper-parameters about HeCo}
In this section, we systematically investigate the sensitivity of two main parameters: the threshold of positives $T_{pos}$ and the threshold of sampled neighbors with $T_{\Phi_m}$. We conduct node classification on ACM and AMiner datasets for analysing $T_{\Phi_m}$, and all four used datasets for analysing $T_{pos}$. We report the Micro-F1 values for comparison.

\begin{figure}[h]
\centering
\subfigure[ACM]{
\label{pos_acm}
        \centering
\includegraphics[scale=0.2]{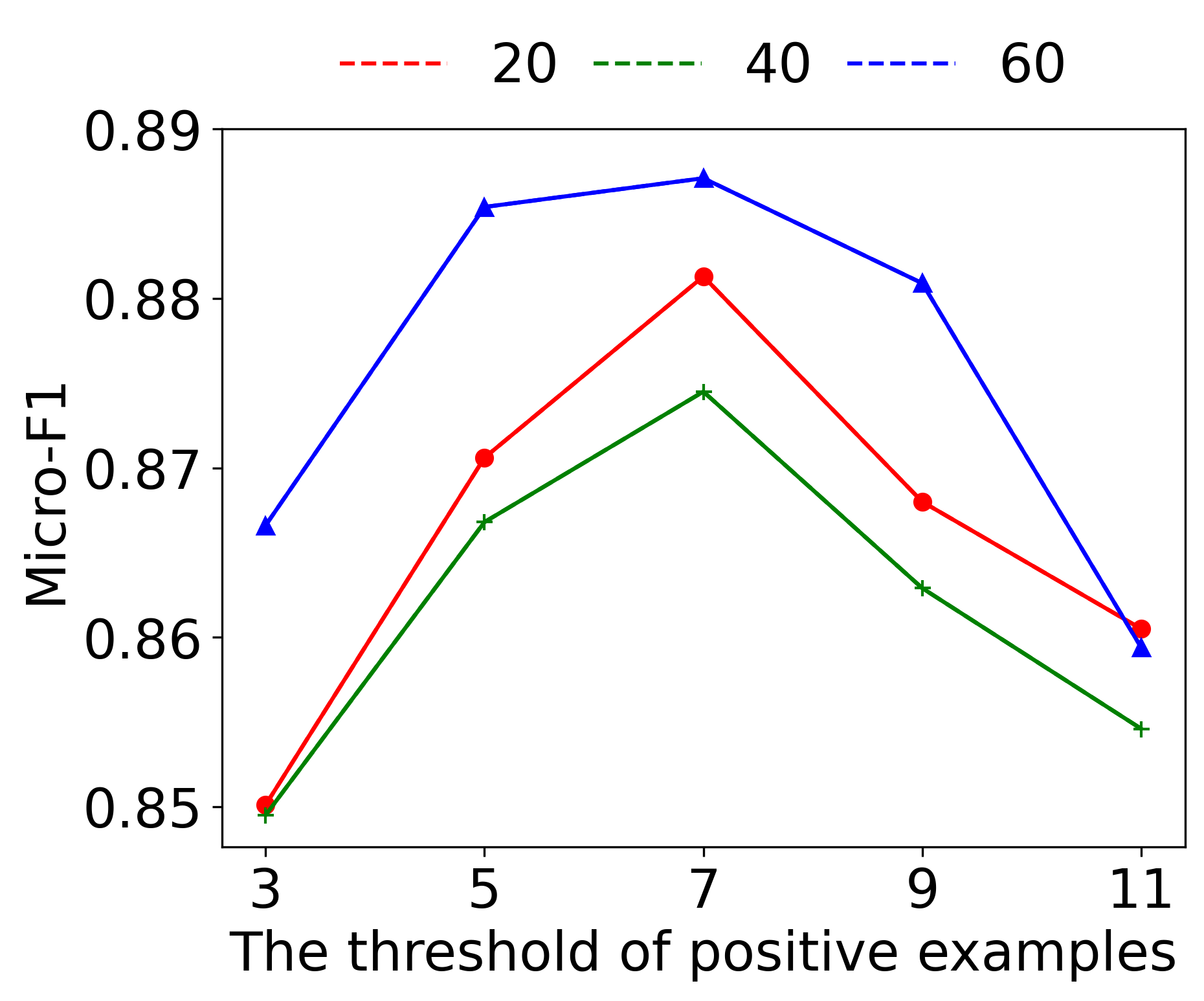}}
\subfigure[AMiner]{
\label{pos_aminer}
        \centering
\includegraphics[scale=0.2]{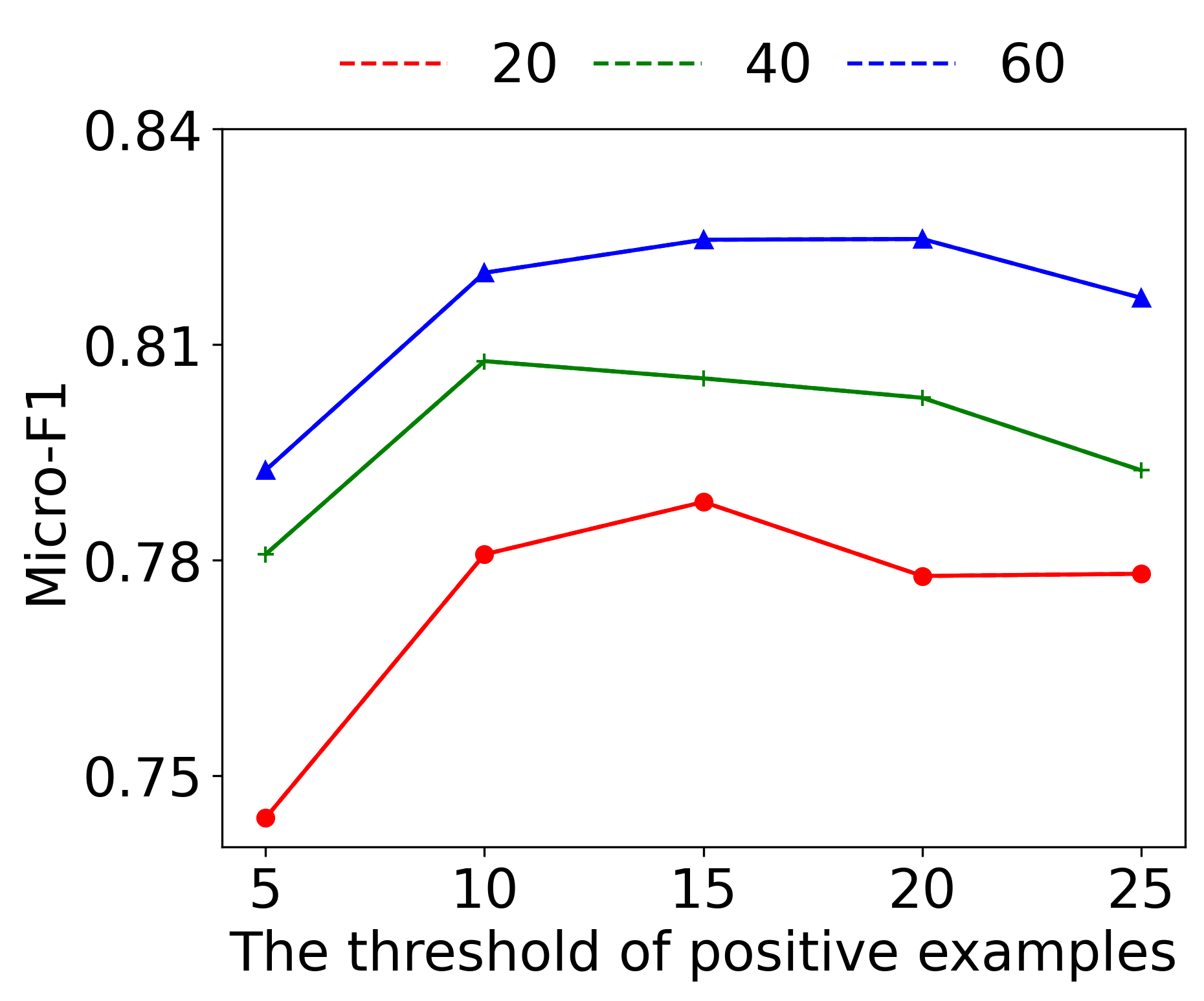}}
\caption{Analysis of the threshold of positive samples.}
\label{pos}
\end{figure}

\textbf{Analysis of $T_{pos}$.} The threshold $T_{pos}$ determines the number of positive samples. We vary the value of it and corresponding results are shown in Figure \ref{pos}. With the increase of $T_{pos}$, the performance goes up first and then declines, and optimum point for ACM is at 7, and at 15 for AMiner. For all datasets, three curves representing different label rates show similar changing trends.

\begin{figure}
\centering
\subfigure[ACM: type A]{
\label{sam_acm}
        \centering
\includegraphics[scale=0.17]{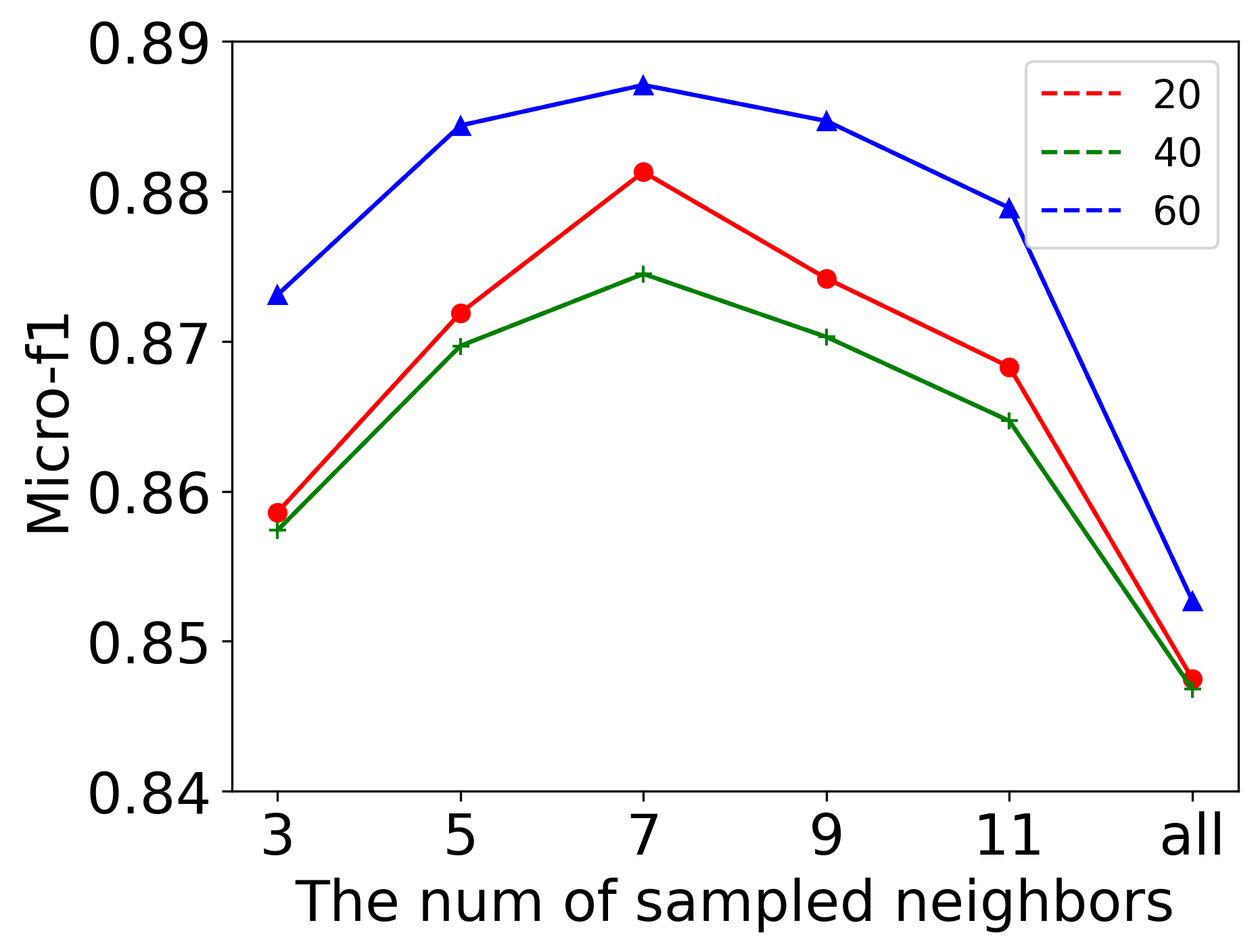}}
\subfigure[AMiner: type A]{
\label{sam_a_aminer}
        \centering
\includegraphics[scale=0.17]{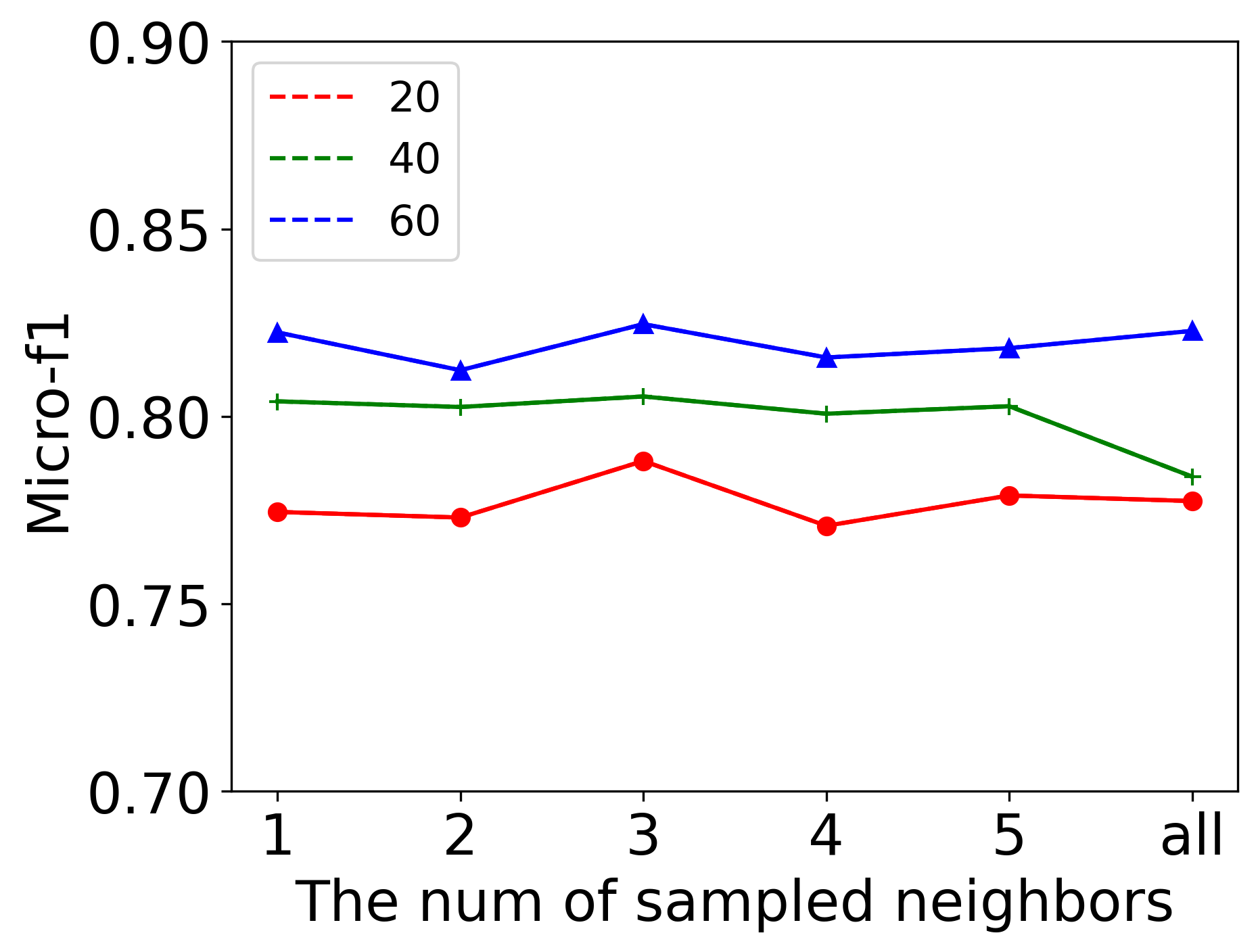}}
\subfigure[AMiner: type R]{
\label{sam_r_aminer}
        \centering
\includegraphics[scale=0.17]{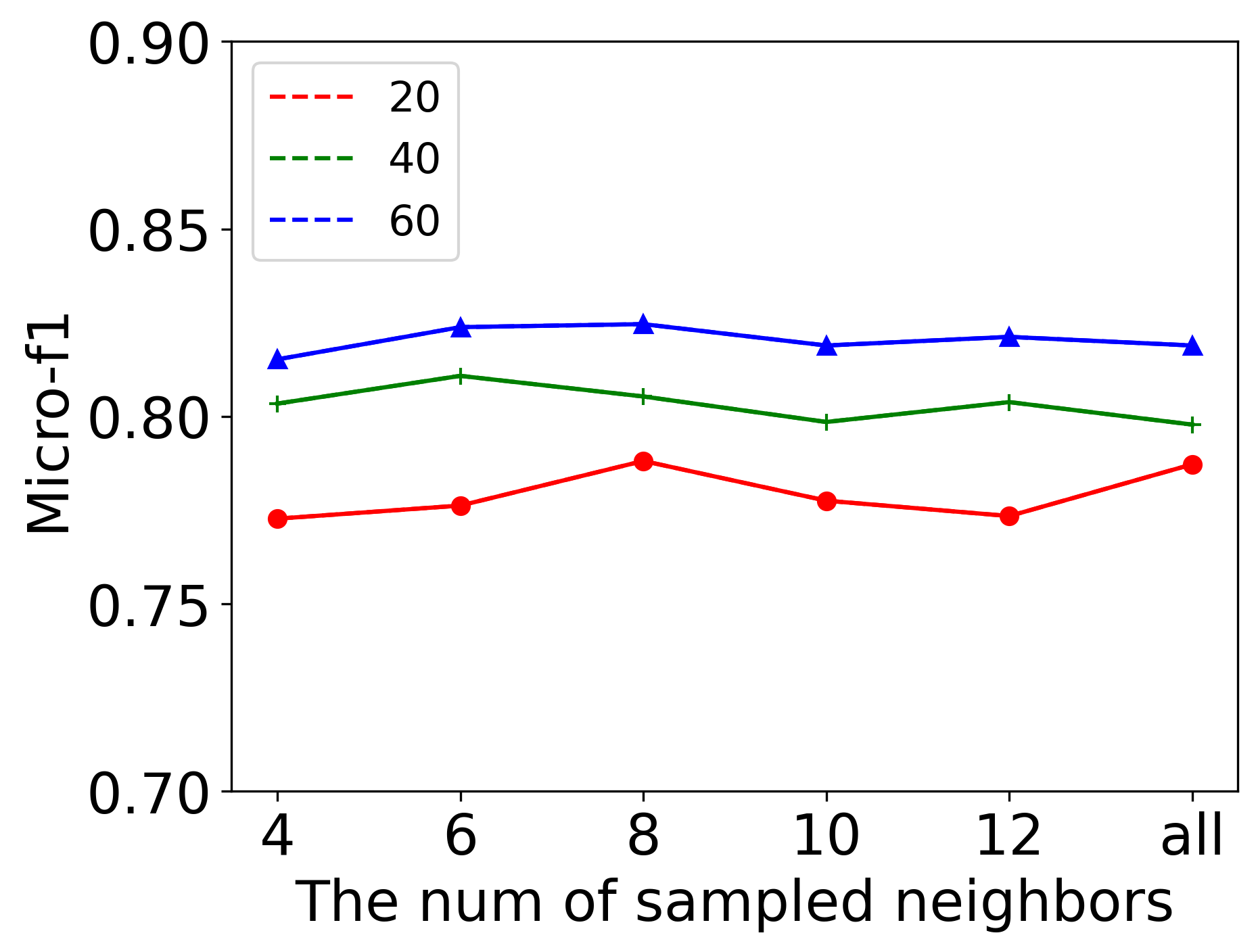}}
\caption{Analysis of the number of sampled neighbors.}
\label{sam}
\end{figure}

\textbf{Analysis of $T_{\Phi_m}$.} To make contrast harder, for target nodes, we randomly sample $T_{\Phi_m}$ neighbors of $\Phi_m$ type, repeatably or not. We again change the value of $T_{\Phi_m}$. It should be pointed out that in ACM, every paper only belongs to one subject (S), so we just change the threshold of type A. The results are shown in Figure \ref{sam}. As can be seen, ACM is sensitive to $T_{\Phi_m}$ of type A, and the best result is achieved when $T_{\Phi_m}=7$. However, AMiner behaves stably with type A or type R. So in our main experiments, we set the $T_{\Phi_m}=3$ for A and $T_{\Phi_m}=8$ for R. Additionally, we also test the case that aggregates all neighbors without sampling, which is marked as "all" in x-axis shown in the figure. In general, "all" cannot perform very well, indicating the usefulness of this sampling strategy.

\subsection{Analysis of Hyper-parameters about HeCo++}
In this section, we systematically analyze the sensitivity of other hyperparameters about the extended model HeCo++, including values of \{$\tau, \tau_{sc}, \tau_{mp}$\}, and number of projection layer. In Appendix~\ref{Additional results on}, analyses on more hyper-parameters are given.

\begin{figure}[h]
\centering
\subfigure[DBLP: $\tau_{mp}$]{
\label{pos_acm}
        \centering
\includegraphics[scale=0.17]{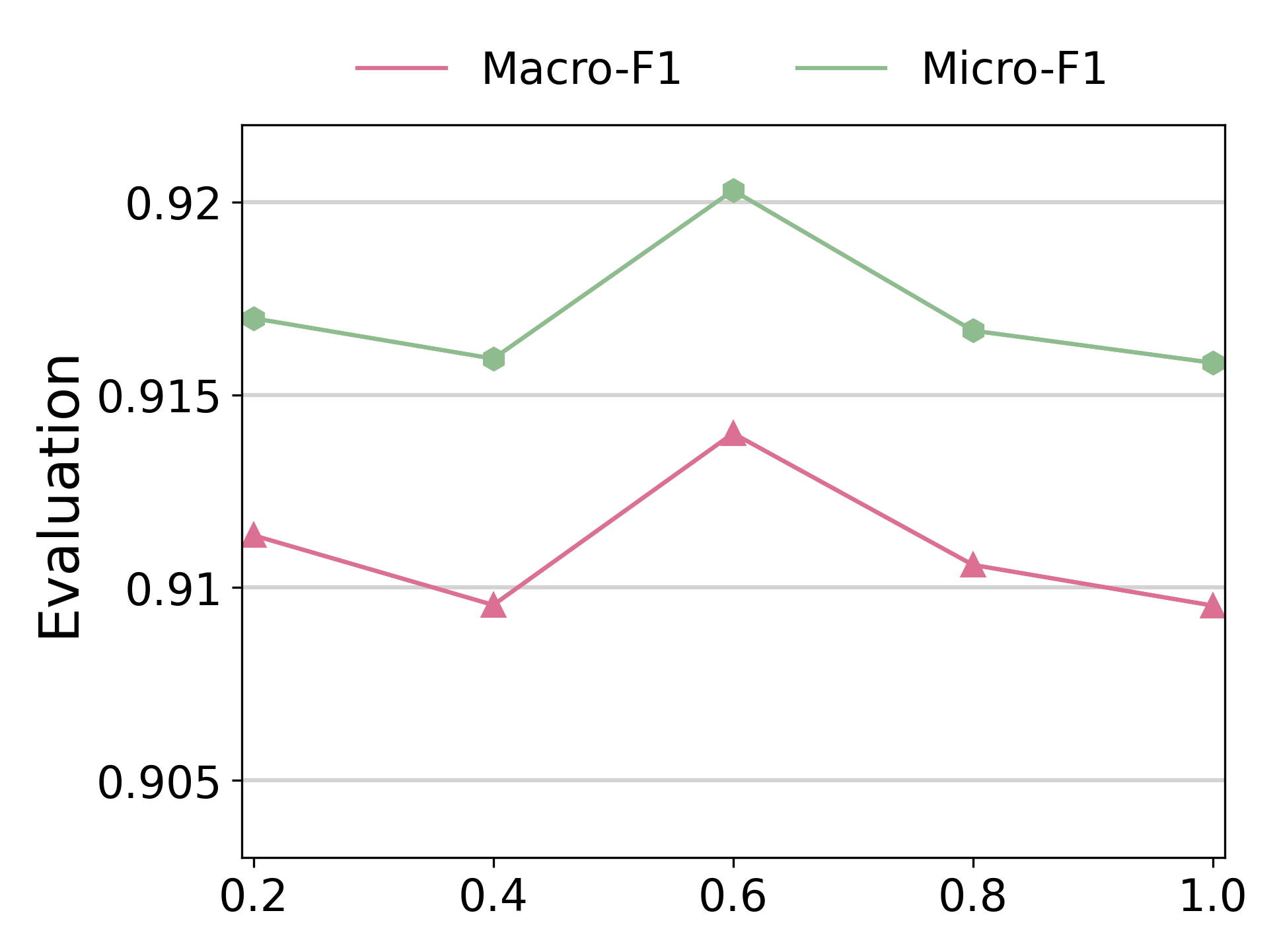}}
\subfigure[DBLP: $\tau_{sc}$]{
\label{pos_acm}
        \centering
\includegraphics[scale=0.17]{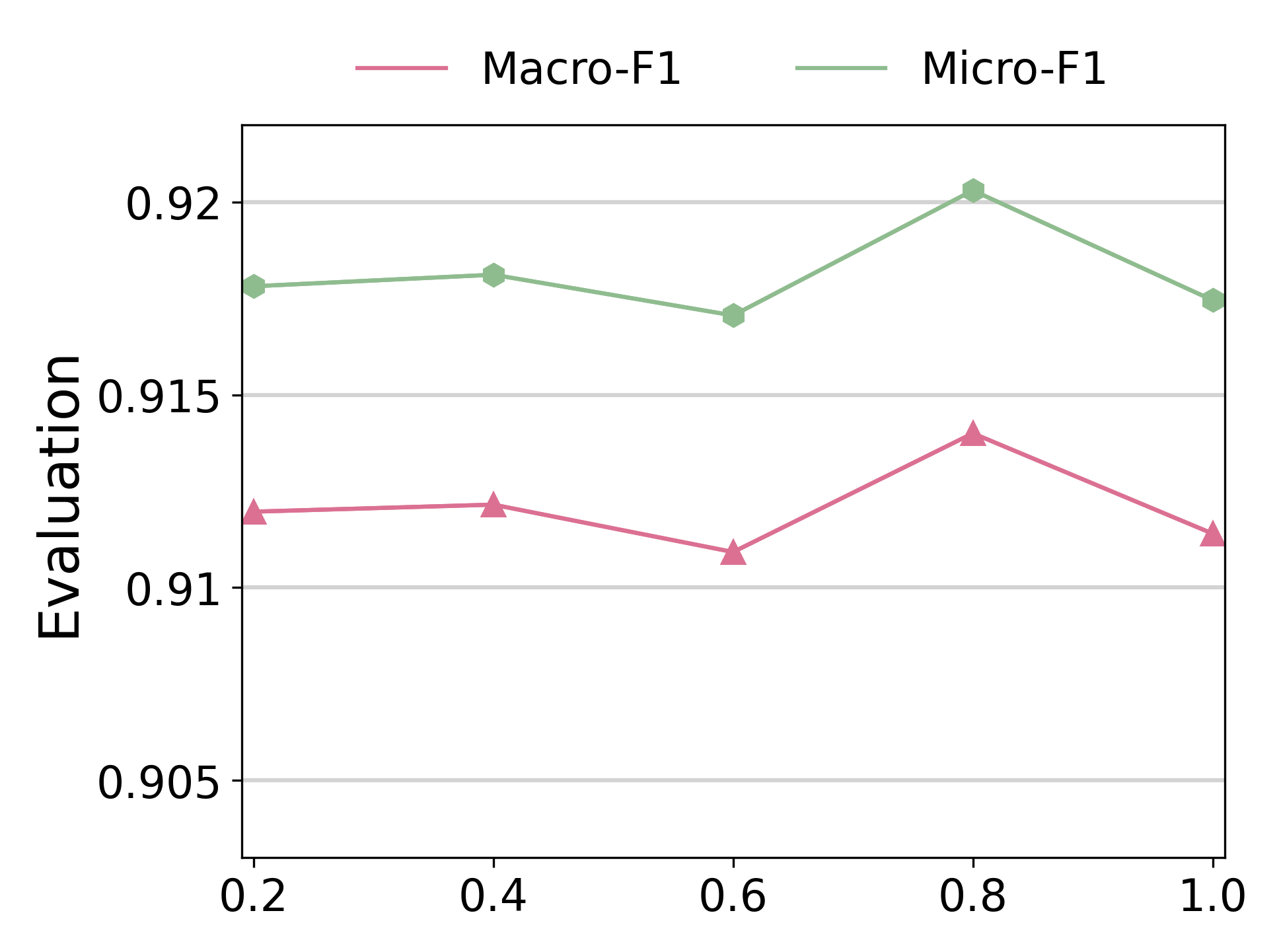}}
\subfigure[DBLP: $\tau$]{
\label{pos_acm}
        \centering
\includegraphics[scale=0.17]{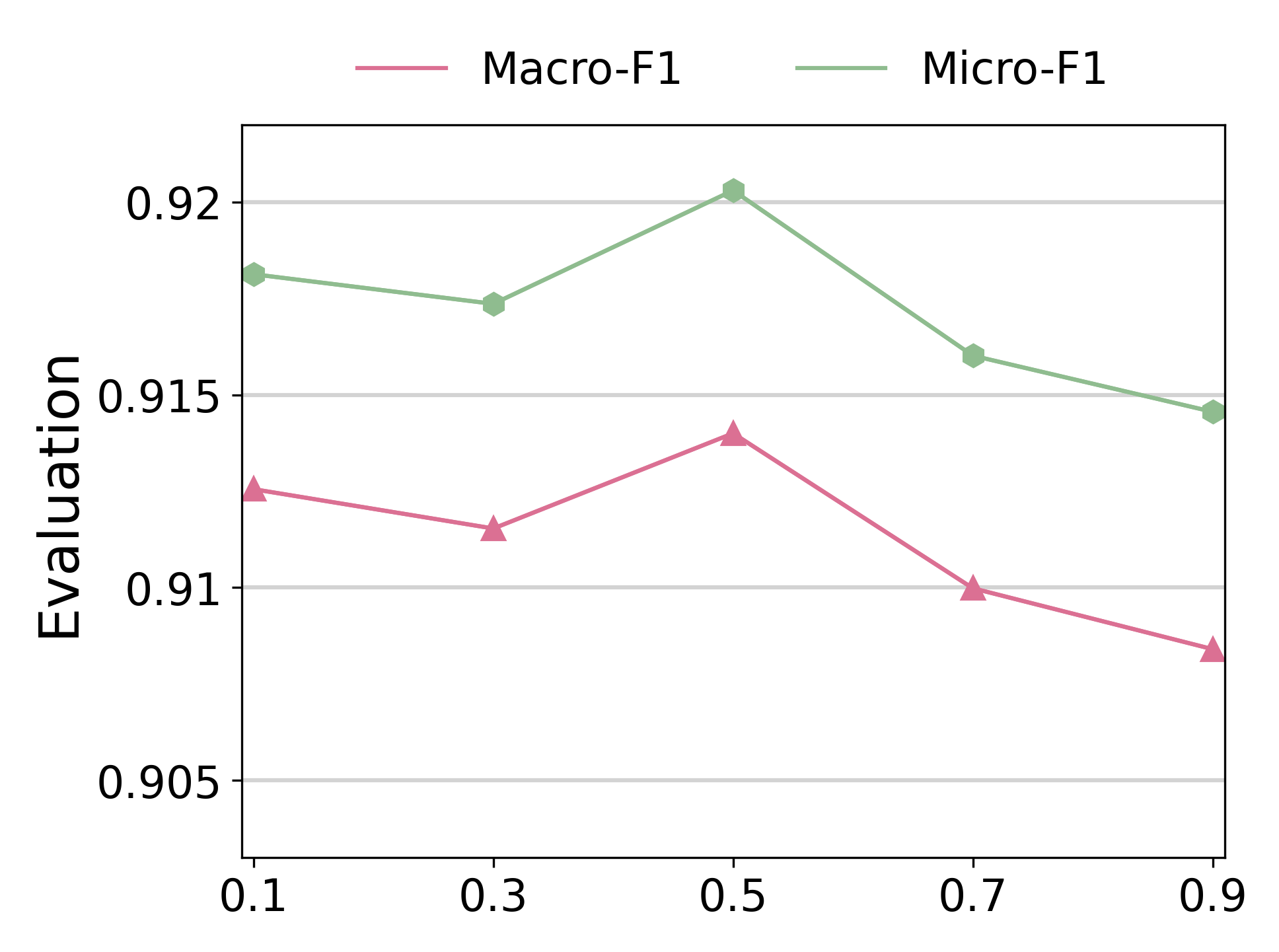}}
\subfigure[Freebase: $\tau_{mp}$]{
\label{pos_acm}
        \centering
\includegraphics[scale=0.17]{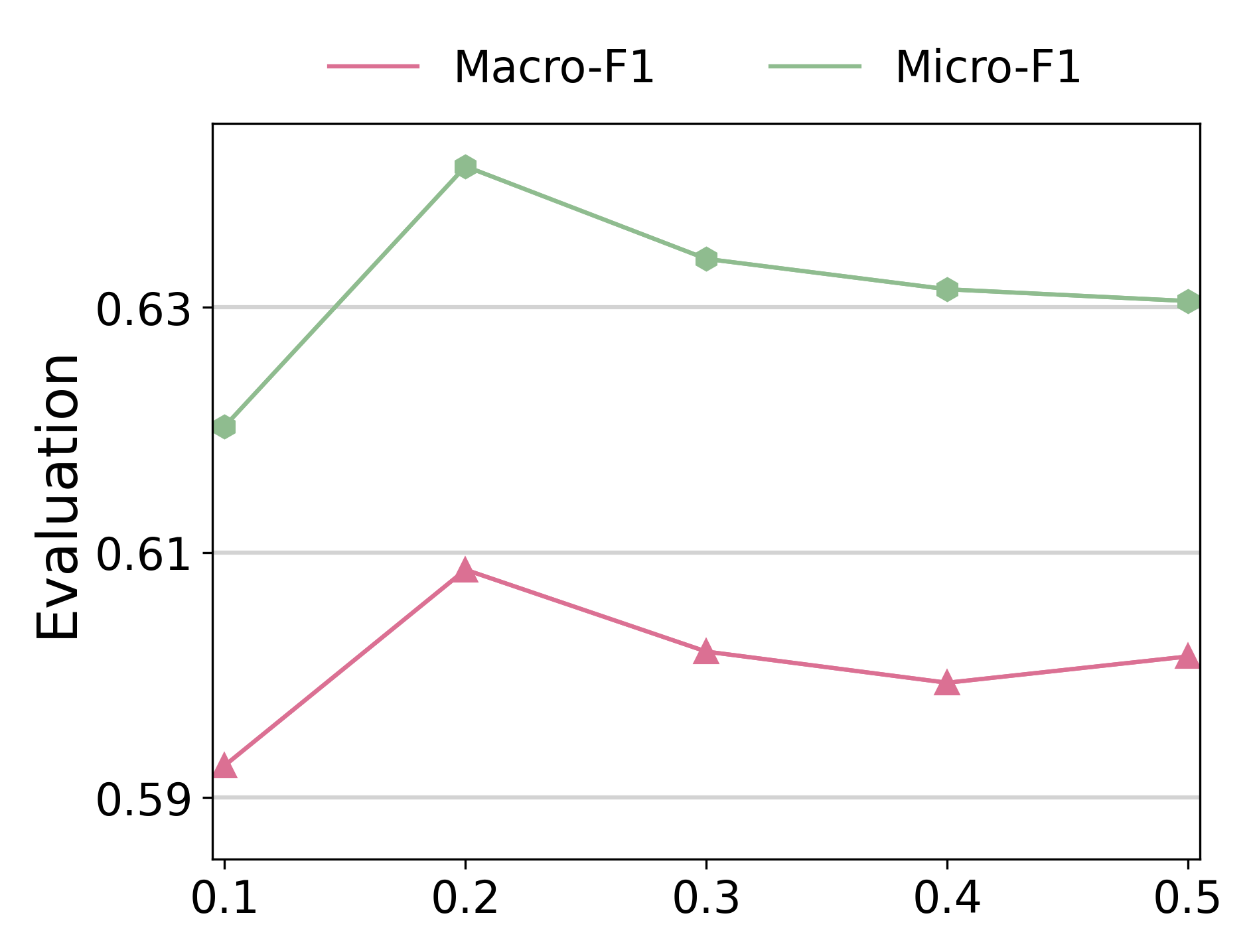}}
\subfigure[Freebase: $\tau_{sc}$]{
\label{pos_acm}
        \centering
\includegraphics[scale=0.17]{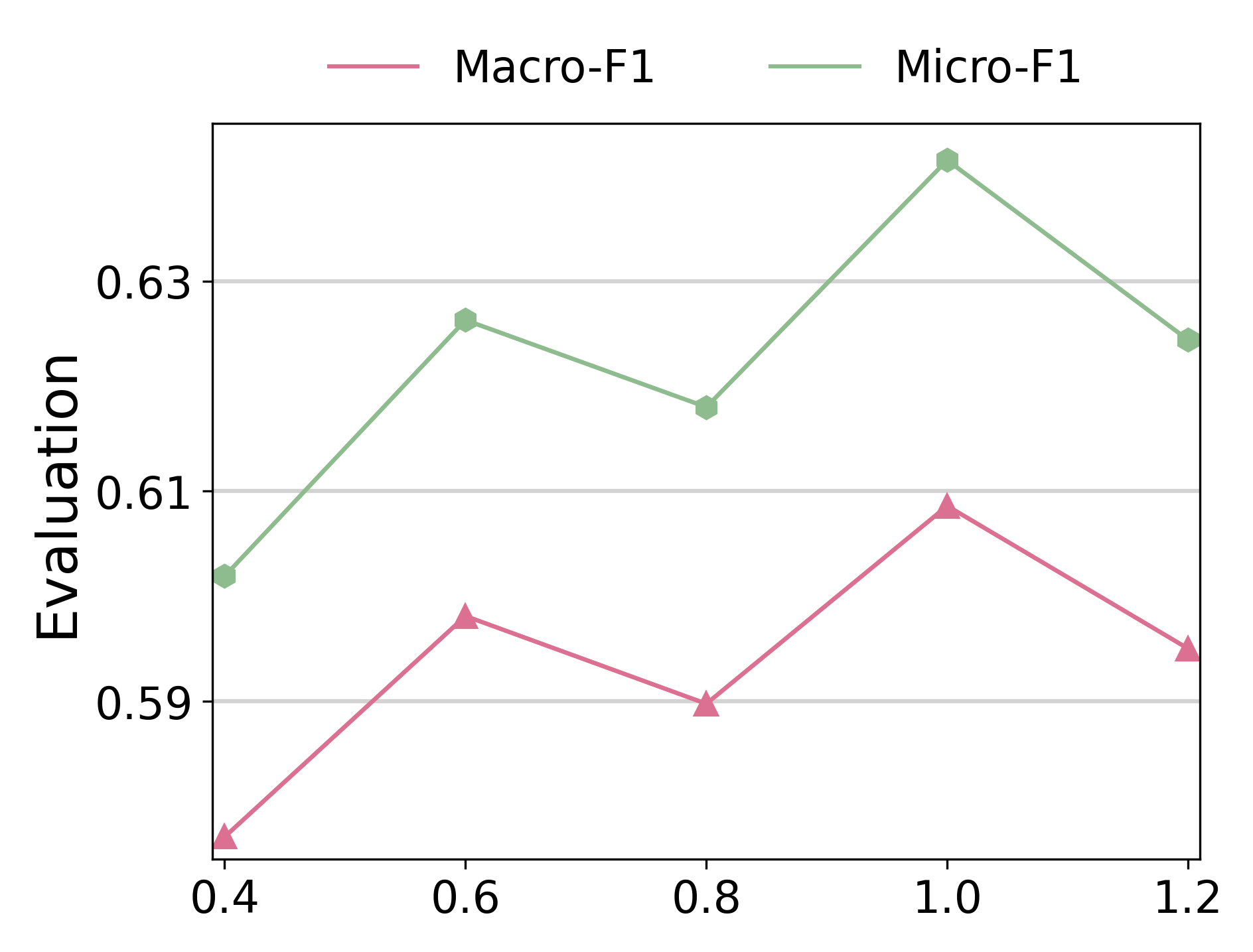}}
\subfigure[Freebase: $\tau$]{
\label{pos_acm}
        \centering
\includegraphics[scale=0.17]{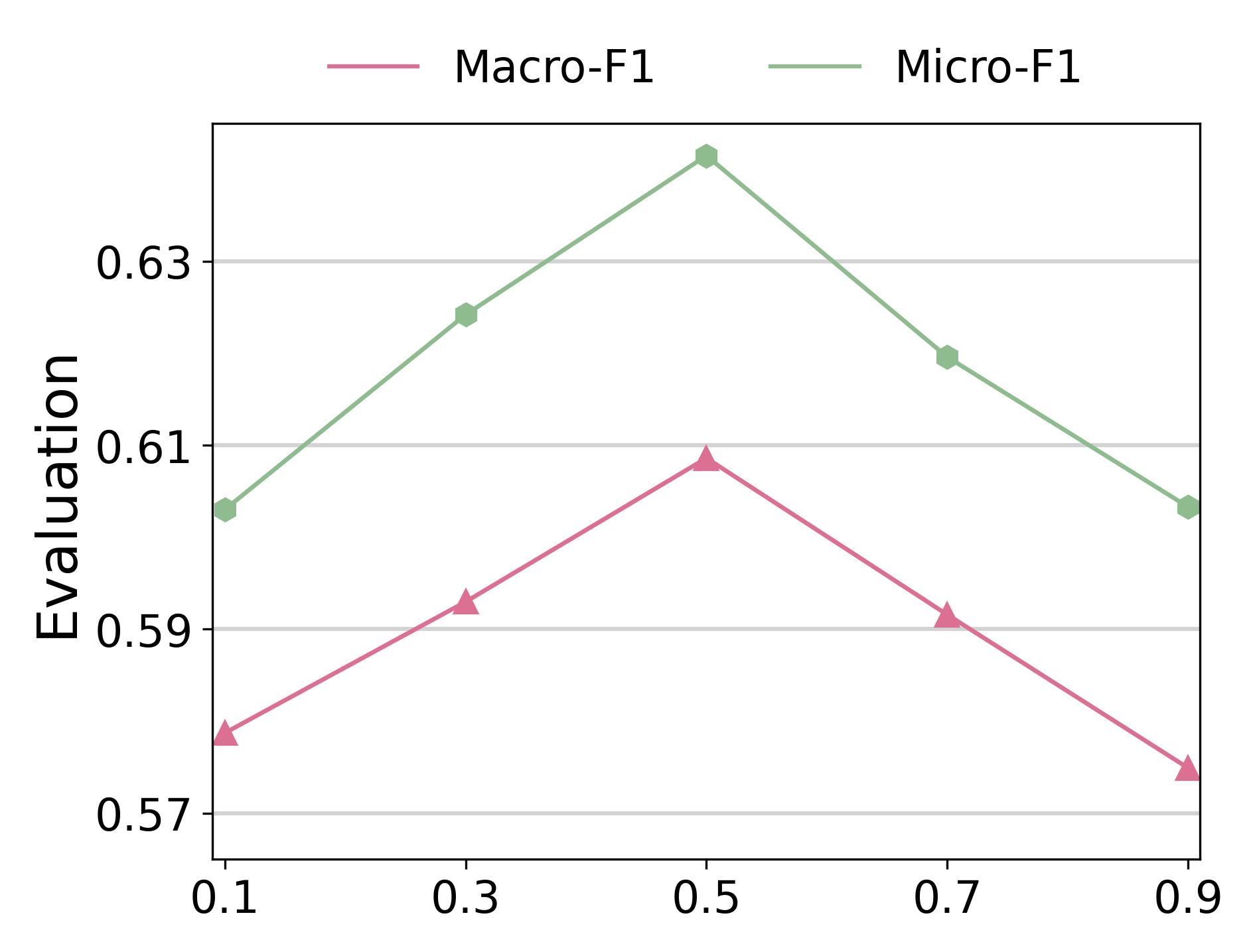}}
\caption{Analysis of $\tau, \tau_{sc}, \tau_{mp}$.}
\label{tau}
\end{figure}

\textbf{Analysis of \{$\tau, \tau_{sc}, \tau_{mp}$\}.} When eq.\eqref{cl}, eq.\eqref{cll}, eq.\eqref{loss1} and eq.\eqref{loss2} are calculated, different $\tau$ control the smoothness of similarity metric in each embedding space. Smaller $\tau$ means a sharped similarity, while larger $\tau$ means a smooth one. We test different $\tau$ of each space on DBLP and Freebase, and the results are shown in Figure \ref{tau}. From the results, we realize that different constructed spaces need specific smooth degrees, due to discrepant structures captured, respectively.

\begin{figure}[h]
\centering
\subfigure[ACM]{
\label{layer_acm}
        \centering
\includegraphics[scale=0.2]{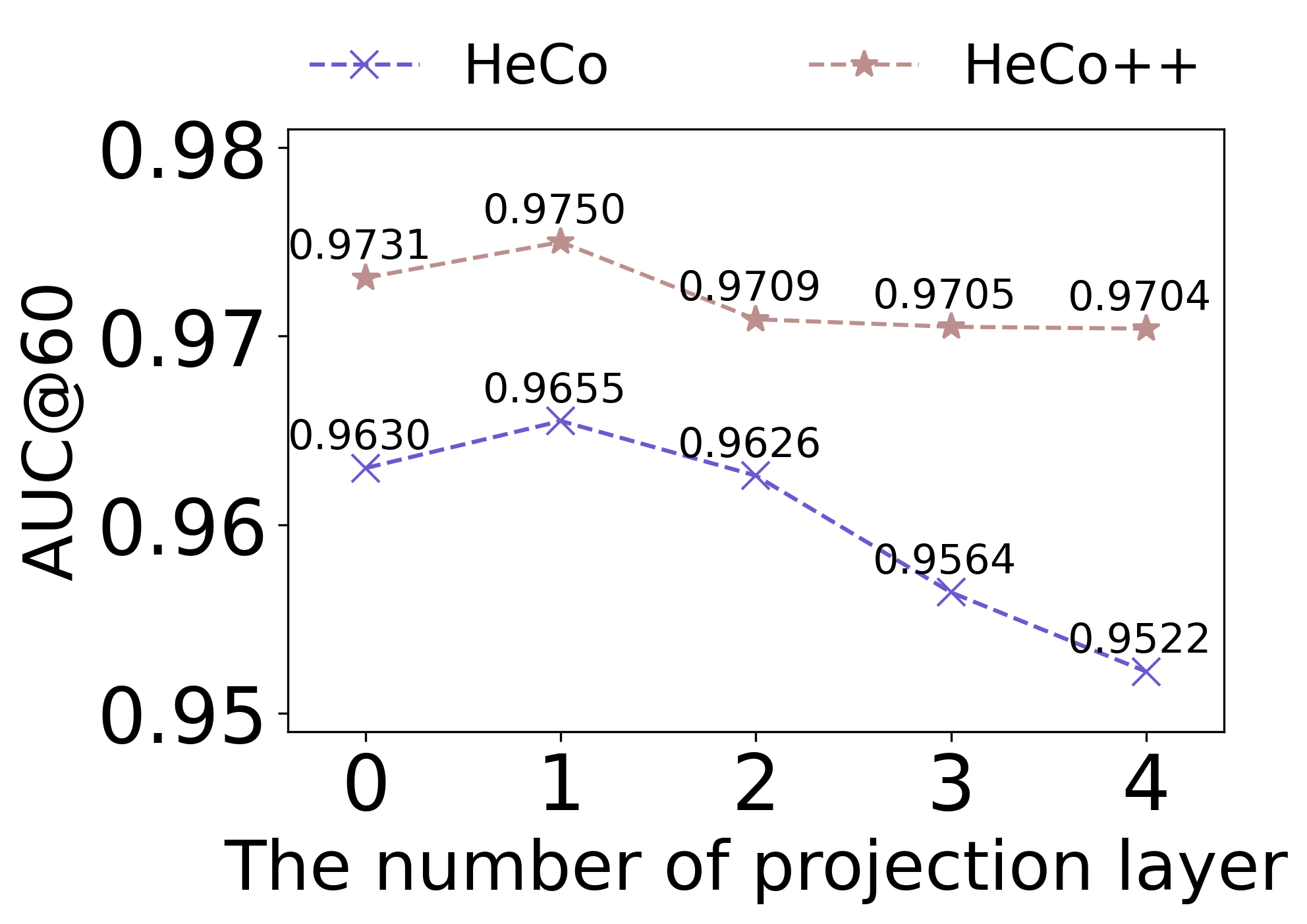}}
\subfigure[DBLP]{
\label{layer_dblp}
        \centering
\includegraphics[scale=0.2]{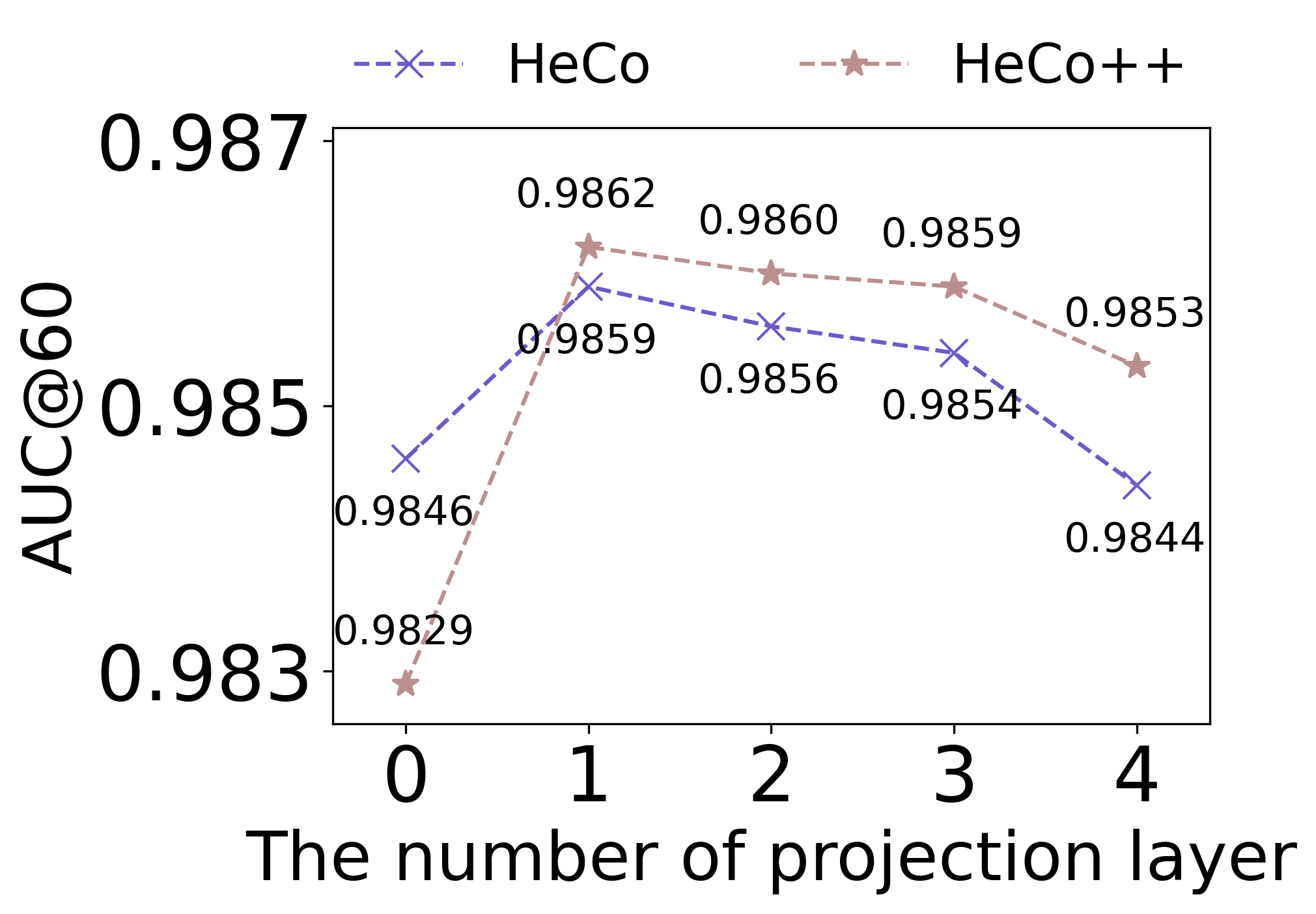}}
\caption{Analysis of number of projection layer.}
\label{layer}
\end{figure}

\textbf{Analysis of number of projection layer.} The importance of projection head has been demonstrated in SimCLRv2~\cite{DBLP:conf/nips/ChenKSNH20}. In eq.~\eqref{proj}~\eqref{proj1} and~\eqref{proj2}, we uniformly rely on one hidden layer to project node embeddings into different spaces. To comprehensively evaluate its effect, we vary the number of layer, and report the AUC of both HeCo and HeCo++ on ACM and DBLP, as shown in Figure~\ref{layer}. From two figures, the optimal numbers on two datasets are both 1. If the number of layer is too small, different spaces can not be fully constructed and separated. If too many layers are used to project, it means many new parameters are involved, which brings the risk of over-fitting.

\section{Conclusion}\label{sec:conclusion}

In this paper, we propose a novel self-supervised heterogeneous graph neural networks with cross-view contrastive learning, named HeCo. HeCo employs network schema and meta-path as two views to capture both of local and high-order structures, and performs the contrastive learning across them. Besides, a view mask mechanism and two ways of generating harder negative samples are designed to make the contrastive learning harder. Moreover, we involve intra-view contrastive learning into original architecture and propose HeCo++ model to further improve the performance of HeCo. Extensive experimental results demonstrate the superior inner mechanism.

In the future, more semantic structures in HIN should be considered for cross-view contrast, for example meta-graph, a high-order and complex structure. Another direction is to avoid the usage of view or augmentation. In HeCo, we attempt to extract network schema and meta-path to depict the local and high-order structure for HIN. However, this process needs domain knowledge and appropriately encoding these two views are non-trivial. SimGRACE~\cite{DBLP:conf/www/XiaWCHL22} is proposed to perturb the encoder parameters, which is a potential reference for heterogeneous contrastive models without data augmentation.

\section*{Acknowledgment}
This work is supported in part by the National Natural Science Foundation of China (No. U20B2045, U1936220, 62192784, 62172052, 62002029, 61772082).


\ifCLASSOPTIONcaptionsoff
  \newpage
\fi



%
\bibliographystyle{IEEEtran}
\bibliography{sample-base.bib}




%

\vspace{-1cm}
\begin{IEEEbiography}[{\includegraphics[width=1in,height=1.25in,clip,keepaspectratio]{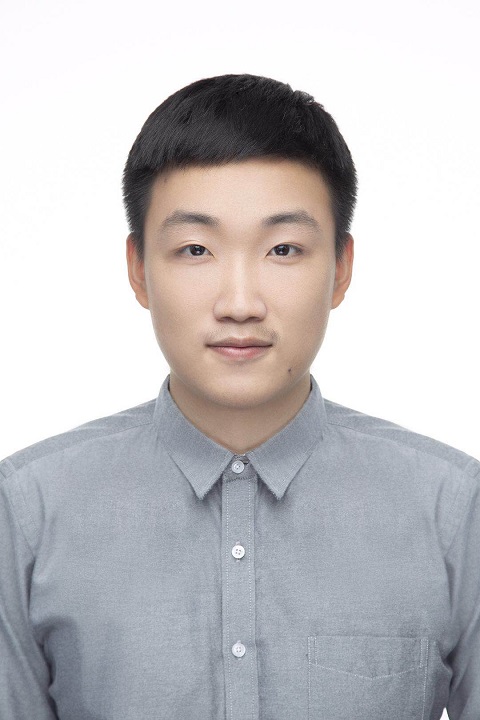}}]{Nian Liu}
received the B.E. degree in 2020 from Beijing University of Posts and Telecommunications. He is a second-year M.S. student in the Department of Computer Science of Beijing University of Posts and Telecommunications. His main research interests including graph mining and contrastive learning.
\end{IEEEbiography}
\vspace{-1cm}
\begin{IEEEbiography}[{\includegraphics[width=1in,height=1.25in,clip,keepaspectratio]{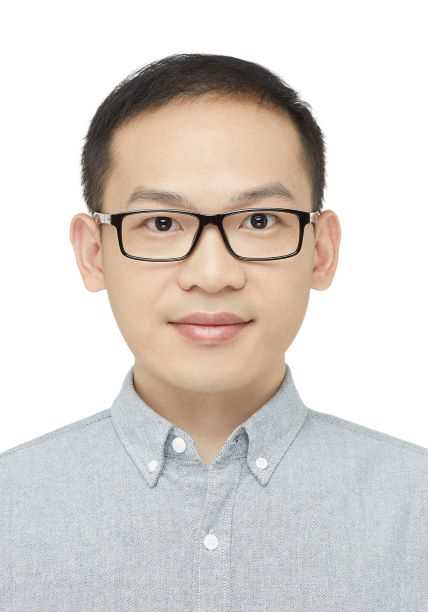}}]{Xiao Wang}
is an Associate Professor in the School of Computer Science, Beijing University of Posts and Telecommunications. He received his Ph.D. degree from the School of Computer Science and Technology, Tianjin University, Tianjin, China, in 2016. He was a postdoctoral researcher in Department of Computer Science and Technology, Tsinghua University, Beijing, China. His current research interests include data mining, social network analysis, and machine learning. Until now, he has published more than 70 papers in refereed journals and conferences.
\end{IEEEbiography}
\vspace{-1cm}
\begin{IEEEbiography}[{\includegraphics[width=1in,height=1.25in,clip,keepaspectratio]{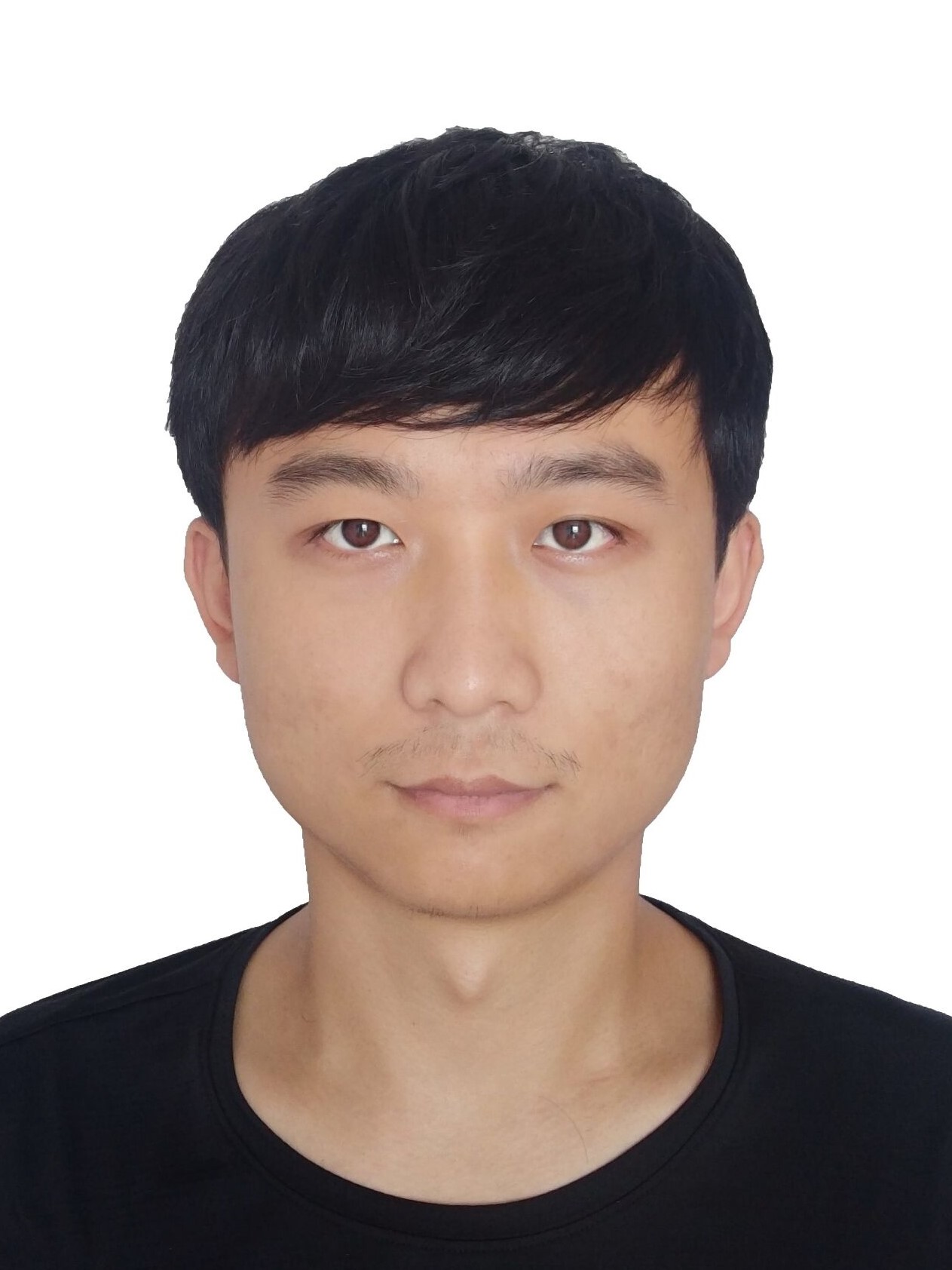}}]{Hui Han}
received the B.S. degree from Beijing Normal University in 2020. He is currently a master student in Beijing University of Posts and Communications. His current research interests are in graph neural networks, data mining and machine learning.
\end{IEEEbiography}
\vspace{-1cm}
\begin{IEEEbiography}[{\includegraphics[width=1in,height=1.25in,clip,keepaspectratio]{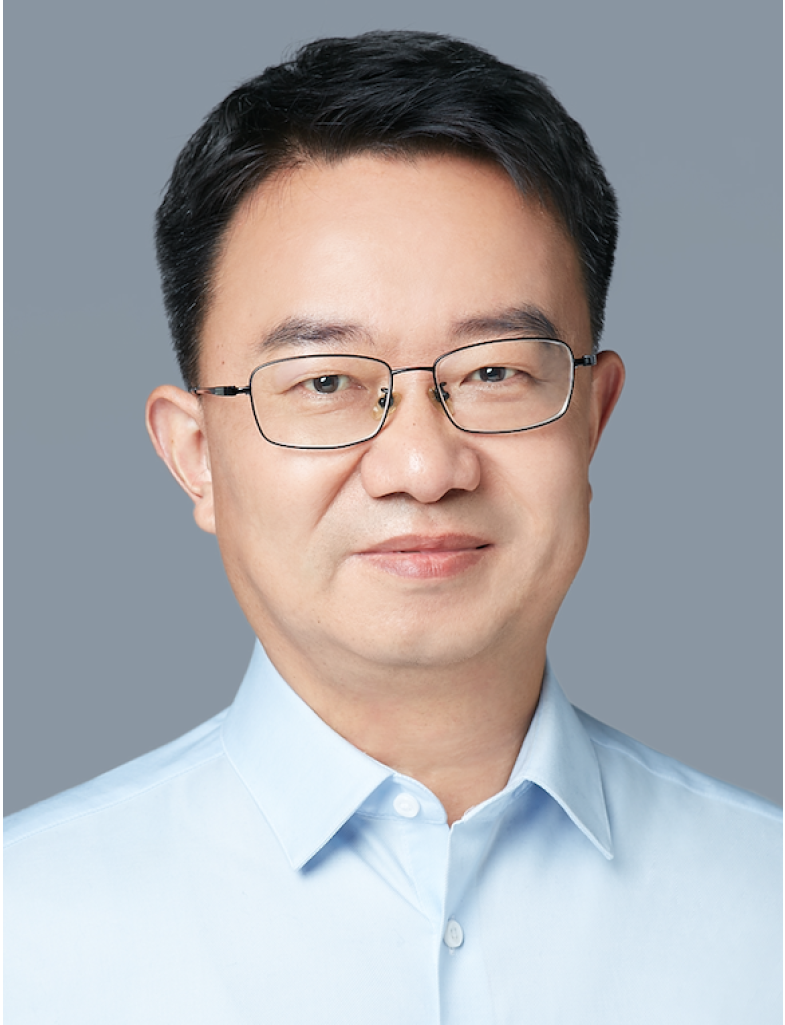}}]{Chuan Shi}
received the B.S. degree from the Jilin University in 2001, the M.S. degree from the Wuhan University in 2004, and Ph.D. degree from the ICT of Chinese Academic of Sciences in 2007. He is a professor and deputy director of Beijing Key Lab of Intelligent Telecommunications Software and Multimedia at present. His research interests are in data mining, machine learning, and evolutionary computing. He has published more than 60 papers in refereed journals and conferences.
\end{IEEEbiography}




\newpage
\appendices
\section{details of H\lowercase{e}C\lowercase{o}\_GAN}
\label{hecogan}
In this section, we further explain the training process of HeCo\_GAN, proposed in section \ref{extension}. 

HeCo\_GAN contains the proposed HeCo, a discriminator D and a generator G. At the beginning of the train, the parameters of HeCo should be warmed up to improve the quality of generated embeddings. So, we first only train HeCo for $K_0$ epochs, which is a hyper-parameter. Then, we get $z^{sc}$ and $z^{mp}$ and utilize them to train D and G alternatively, which is as following two steps:

\begin{itemize}
    \item Freeze G and train D for $K_D$ epochs. For target node i and its embedding $z_i^{sc}$ under network schema view, we can get the embeddings of nodes in $\mathbb{P}_i$ under meta-path view. D outputs a probability that a sample $j$ is from $\mathbb{P}_i$ given $z_i^{sc}$:
    \begin{equation}
        D\left(z_j|z_i^{sc}\right)=\frac{1}{1+\exp\left(-{z_i^{sc}}^\top M^D_{mp}z_j\right)},
    \end{equation}
    where $M^D_{mp}$ is a matrix that projects $z_i^{sc}$ into the space of meta-path view. And the objective function of D under the network schema view is:
    \begin{equation}
    \begin{aligned}
        \mathcal{L}_{i_D}^{sc}=&-\mathop{\mathbb{E}}\limits_{j\sim\mathbbm{p}_i}\log D\left(z_j^{mp}|z_i^{sc}\right)\\
        &-\mathop{\mathbb{E}}\limits_{\widetilde{z_i^{mp}}\sim G\left(z_i^{sc}\right)}\log\left(1-D\left(\widetilde{z_i^{mp}}|z_i^{sc}\right)\right),
    \end{aligned}
    \end{equation}
    where $\mathbbm{p}_i\subset\mathbb{P}_i$, which is chosen randomly, and $\widetilde{z_i^{mp}}$ is generated by generator based on $z_i^{sc}$. This shows that given $z_i^{sc}$, D aims to identify its positive samples from meta-path view as positive and samples generated by G as negative. Notice that the number of fake samples from G is the same as $|\mathbbm{p}_i|$. Similarly, we can also get the objective function of D under the meta-path view $\mathcal{L}_{i_D}^{mp}$. So, we train the discriminator D by minimizing the following loss:
    \begin{equation}
        \mathcal{L}_{D}=\frac{1}{|B|}\sum\limits_{i\in B}\frac{1}{2}\left(\mathcal{L}_{i_D}^{sc}+\mathcal{L}_{i_D}^{mp}\right),
    \end{equation}
    where $B$ denotes the batch of nodes that are trained in current epoch.
    
    \item Freeze D and train G for $K_G$ epochs. G gradually improves the quality of generated samples by fooling D. Specifically, given the target $i$ and its embedding $z_i^{sc}$ under network schema view, G first constructs a Gaussian distribution center on $i$, and draws samples from it, which is related to $z_i^{sc}$:
    \begin{equation}
        e^{mp}_{i}\sim\mathcal{N}\left({z_i^{sc}}^\top M^G_{mp}, \sigma^2\textbf{I}\right),
    \end{equation}
    where $M^G_{mp}$ is also a projected function to map $z_i^{sc}$ into meta-path space, and $\sigma^2\textbf{I}$ is covariance. We then apply one-layer MLP to enhance the expression of the fake samples:
    \begin{equation}
        \widetilde{z^{mp}_{i}}=G\left(z_i^{sc}\right)=\sigma\left(We^{mp}_{i}+b\right).
    \end{equation}
    Here, $\sigma$, $W$ and $b$ denote non-linear activation, weight matrix and bias vector, respectively. To fool the discriminator, generator is trained under network schema view by following loss:
    \begin{equation}
    \begin{aligned}
        \mathcal{L}_{i_G}^{sc}&=-\mathop{\mathbb{E}}\limits_{\widetilde{z^{mp}_{i}}\sim G\left(z_i^{sc}\right)}\log D(\widetilde{z^{mp}_{i}}|z_i^{sc}),\\
        \mathcal{L}_{G}&=\frac{1}{|B|}\sum\limits_{i\sim B}\frac{1}{2}\left(\mathcal{L}_{i_G}^{sc}+\mathcal{L}_{i_G}^{mp}\right).
    \end{aligned}
    \end{equation}
    Again, $\mathcal{L}_{i_G}^{mp}$ is attained like $\mathcal{L}_{i_G}^{sc}$.
\end{itemize}

\noindent These two steps are alternated for $I_{DG}$ times to fully train the D and G.

Once we get the well-trained G, high-quality negative samples $\widetilde{z^{mp}_i}$ and $\widetilde{z^{sc}_i}$ will be obtained, given $z^{sc}_i$ and $z^{mp}_i$, respectively. And they are combined with original negative samples from meta-path view or network schema view. Finally, the extended set of negative samples is fed into HeCo to boost the training for $K_{H}$ epochs.

The training processes of the proposed HeCo, discriminator D and generator G are employed iteratively until to the convergence.

\section{Model complexity analyses}
\label{time}
In this section, we give the detailed model complexity analyses for all variants, including HeCo\_sc, HeCo\_mp, HeCo and HeCo++. Specifically, all variants firstly need node feature transformation, whose complexity is $\mathcal{O}(|\mathcal{V}|d)$, where $|\mathcal{V}|$ is the number of all types of nodes and $d$ is dimension. Then, we analyze encoding network schema view, encoding meta-path view and collaboratively contrastive optimization:
\begin{itemize}
    \item \textbf{Encoding network schema view.} This part contains node-level and type-level attention. For the former, the complexity is $\mathcal{O}(|\mathcal{V}_t||\Phi|d^2)$, where $\mathcal{V}_t$ is the set of target type of nodes and $|\mathrm{\Phi}|$ means the average number of neighbors for each target node. For the latter, its complexity is $\mathcal{O}\left(\left|\mathcal{V}_t\right|Sd^2\right)$, where $S$ is the number of other types. Therefore, the overall complexity of this part is $\mathcal{O}\left(\left|\mathcal{V}_t\right||\mathrm{\Phi}|d^2+\left|\mathcal{V}_t\right|Sd^2\right)$.
    \item \textbf{Encoding meta-path view.} This part contains meta-path specific GCN and semantic attention. For the former, the complexity is $\mathcal{O}\left(M\left|\mathcal{P}\right|d\right)$, where $M$ is the number of meta-paths and $\left|\mathcal{P}\right|$ means the average number of edges in each meta-path based adjacency matrix. For the latter, its complexity is $\mathcal{O}\left(\left|\mathcal{V}_t\right|Md^2\right)$. Therefore, the overall complexity of this part is $\mathcal{O}\left(M\left|\mathcal{P}\right|d+\left|\mathcal{V}_t\right|Md^2\right)$.
    \item \textbf{Collaboratively contrastive optimization.} In this part, we first project embeddings into contrastive space, $\mathcal{O}\left(\left|\mathcal{V}_t\right|d\right)$. Then, in contrastive loss, we need to calculate the cosine similarity between target node and its positive set and negative set. Thus, this part consumes $\mathcal{O}\left(\left|\mathcal{V}_t\right|\left(\left|P_i\right|+\left|N_i\right|\right)d^2\right)$, where $P_i$ and $N_i$ are positive sample set and negative sample set, respectively. Therefore, the overall complexity of this part is $\mathcal{O}\left(\left|\mathcal{V}_t\right|d+\left|\mathcal{V}_t\right|\left(\left|P_i\right|+\left|N_i\right|\right)d^2\right)$.
\end{itemize}
According to above analyses, we can get the complexities of all variants:
\begin{itemize}
    \item \textbf{HeCo\_sc.} This variant consists of node feature transformation, encoding network schema view and one collaboratively contrastive optimization. So, its complexity is $\mathcal{O}(\left|\mathcal{V}\right|d+\left|\mathcal{V}_t\right||\mathrm{\Phi}|d^2+\left|\mathcal{V}_t\right|Sd^2+\left|\mathcal{V}_t\right|d+\left|\mathcal{V}_t\right|(\left|P_i\right|$+$\left|N_i\right|)d^2)$.
    \item \textbf{HeCo\_mp.} This variant consists of node feature transformation, encoding meta-path view and one collaboratively contrastive optimization. So, its complexity is $\mathcal{O}(\left|\mathcal{V}\right|d+M\left|\mathcal{P}\right|d+\left|\mathcal{V}_t\right|Md^2+\left|\mathcal{V}_t\right|d+\left|\mathcal{V}_t\right|(\left|P_i\right|$+$\left|N_i\right|)d^2)$.
    \item \textbf{HeCo.} This variant consists of node feature transformation, encoding network schema view, encoding meta-path view and two collaboratively contrastive optimizations. So, its complexity is $\mathcal{O}(\left|\mathcal{V}\right|d+\left|\mathcal{V}_t\right||\mathrm{\Phi}|d^2+\left|\mathcal{V}_t\right|Sd^2+M\left|\mathcal{P}\right|d+\left|\mathcal{V}_t\right|Md^2$+$2\left|\mathcal{V}_t\right|d+2\left|\mathcal{V}_t\right|\left(\left|P_i\right|+\left|N_i\right|\right)d^2)$.
    \item \textbf{HeCo++.} This variant consists of node feature transformation, encoding network schema view, encoding meta-path view and four collaboratively contrastive optimizations. So, its complexity is $\mathcal{O}(\left|\mathcal{V}\right|d+\left|\mathcal{V}_t\right||\mathrm{\Phi}|d^2+\left|\mathcal{V}_t\right|Sd^2+M\left|\mathcal{P}\right|d+\left|\mathcal{V}_t\right|Md^2$+$4\left|\mathcal{V}_t\right|d+4\left|\mathcal{V}_t\right|\left(\left|P_i\right|+\left|N_i\right|\right)d^2)$.
\end{itemize}

Compared with main baseline DMGI, our proposed HeCo and HeCo++ only additionally encode network schema view, whose complexity is linear to the number of target type of nodes. In the table~\ref{Runtime}, we give the real runtime per epoch of DMGI, HeCo and HeCo++.

\begin{table}[h]
  \caption{Runtime comparison between DMGI, HeCo and HeCo++}
  \label{Runtime}
  \resizebox{0.45\textwidth}{!}{
  \begin{tabular}{c|c|c|c|c}
    \bottomrule
        Runtime (s) $/$ epoch & ACM & DBLP & Freebase & AMiner\\
    \bottomrule
    DMGI & 0.351&	0.204&	0.272&	0.124\\
    HeCo & 0.332&	0.235&	0.375&	0.482 \\
    HeCo++ & 0.310&	0.239&	0.409&	0.501 \\
    \bottomrule
    HeCo w$/$o sample & 0.081&	0.067&	0.041&	0.075\\
    HeCo++ w$/$o sample & 0.054&	0.088&	0.050&	0.078\\
    \bottomrule
\end{tabular}}
\end{table}

From Table~\ref{Runtime}, we can see that for two small datasets, ACM and DBLP, our proposed models consume similar time compared with DMGI. As the size of dataset increases, our proposed models consume more time than DMGI. That’s because the number of nodes increases, so we have to take more time to encode network schema view as above time complexity analyses. In practical code review, we find that the most time-consuming part in encoding network schema view is sampling in each type of neighbors for every target node. If we use the same sampling results in each epoch, runtime will be sharply reduced, as shown in Table~\ref{Runtime} (HeCo w/o sample and HeCo++ w/o sample).

\section{Experiments}
\subsection{Datasets}
\label{Datasetss}
\begin{table}[t]
  \caption{The statistics of the datasets}
  \centering
  \label{statistics}
  \resizebox{0.4\textwidth}{!}{
  \begin{tabular}{|c|c|c|c|}
    \hline
        Dataset & Node & Relation & Meta-path\\
    \hline
    ACM & \makecell*[c]{paper (P):4019\\author (A):7167\\subject (S):60} & \makecell*[c]{P-A:13407\\P-S:4019} & \makecell*[c]{PAP\\PSP} \\
    \hline
    DBLP & \makecell*[c]{author (A):4057\\paper (P):14328\\conference (C):20\\term (T):7723} & \makecell*[c]{P-A:19645\\P-C:14328\\P-T:85810} & \makecell*[c]{APA\\APCPA\\APTPA} \\
    \hline
    Freebase & \makecell*[c]{movie (M):3492\\actor (A):33401\\direct (D):2502\\writer (W):4459} & \makecell*[c]{M-A:65341\\M-D:3762\\M-W:6414} & \makecell*[c]{MAM\\MDM\\MWM} \\
    \hline
    AMiner & \makecell*[c]{paper (P):6564\\author (A):13329\\reference (R):35890} & \makecell*[c]{P-A:18007\\P-R:58831} & \makecell*[c]{PAP\\PRP} \\
    \hline
\end{tabular}}
\end{table}

We employ the following four real HIN datasets, where the basic information are summarized in Table \ref{statistics}.
\begin{itemize}
\item \textbf{ACM}\footnote{https://github.com/Andy-Border/NSHE} \cite{nshe}. The target nodes are papers, which are divided into three classes. For each paper, there are 3.33 authors averagely, and one subject. 
\item \textbf{DBLP}\footnote{https://github.com/cynricfu/MAGNN} \cite{magnn}. The target nodes are authors, which are divided into four classes. For each author, there are 4.84 papers averagely. 
\item \textbf{Freebase}\footnote{https://github.com/dingdanhao110/Conch} \cite{freebase}. The target nodes are movies, which are divided into three classes. For each movie, there are 18.7 actors, 1.07 directors and 1.83 writers averagely. 
\item \textbf{AMiner}\footnote{https://github.com/librahu/HIN-Datasets-for-Recommendation-and-Network-Embedding} \cite{hegan}. The target nodes are papers. We extract a subset of original dataset, where papers are divided into four classes. For each paper, there are 2.74 authors and 8.96 references averagely. 
\end{itemize}

\subsection{Baselines}
\label{Baselines}
The publicly available implementations of baselines can be found at the following URLs:
\begin{itemize}
    \item GraphSAGE: https://github.com/williamleif/Graph-\\SAGE
    \item GAE:  https://github.com/tkipf/gae
    \item DGI: https://github.com/PetarV-/DGI
    \item Mp2vec: https://ericdongyx.github.io/metapath2vec/\\m2v.html
    \item HERec: https://github.com/librahu/HERec
    \item HetGNN: https://github.com/chuxuzhang/KDD2019\_\\HetGNN
    \item DMGI: https://github.com/pcy1302/DMGI
    \item HAN: https://github.com/Jhy1993/HAN
\end{itemize}

\begin{figure*}[t]
\centering
\hspace{-8mm}
\subfigure[Mp2vec]{
\label{Mp2vec_acm}
\includegraphics[scale=0.25]{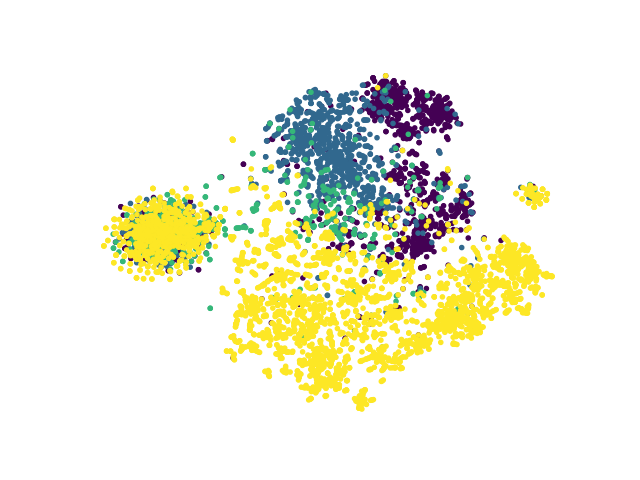}
}\hspace{-8mm}
\subfigure[DGI]{
\label{DGI_acm}
\includegraphics[scale=0.25]{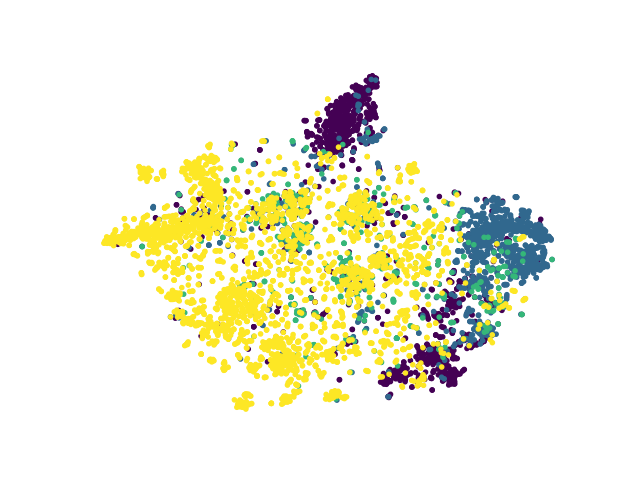}}\hspace{-8mm}
\subfigure[DMGI]{
\label{DMGI_v_acm}
\includegraphics[scale=0.25]{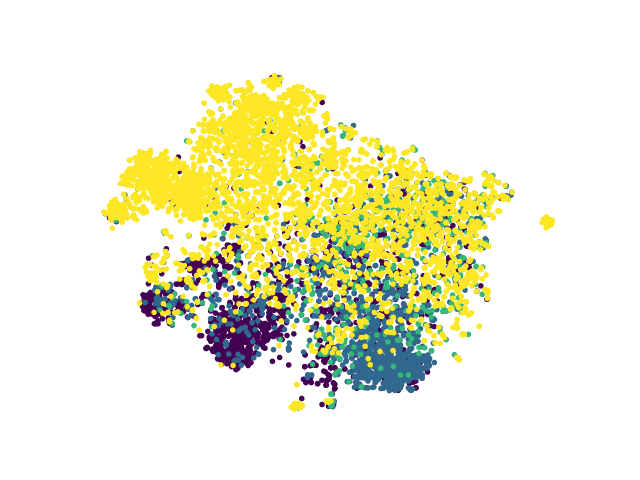}}\hspace{-8mm}
\subfigure[HeCo]{
\label{HeCo_v_acm}
\includegraphics[scale=0.25]{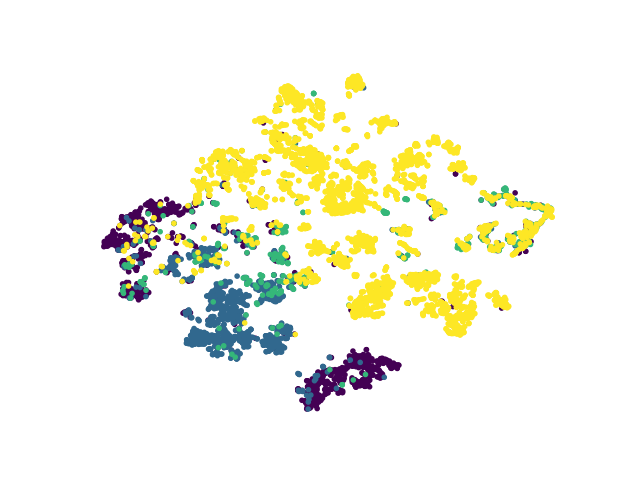}}\hspace{-8mm}
\subfigure[HeCo++]{
\label{HeCo++_v_acm}
\includegraphics[scale=0.25]{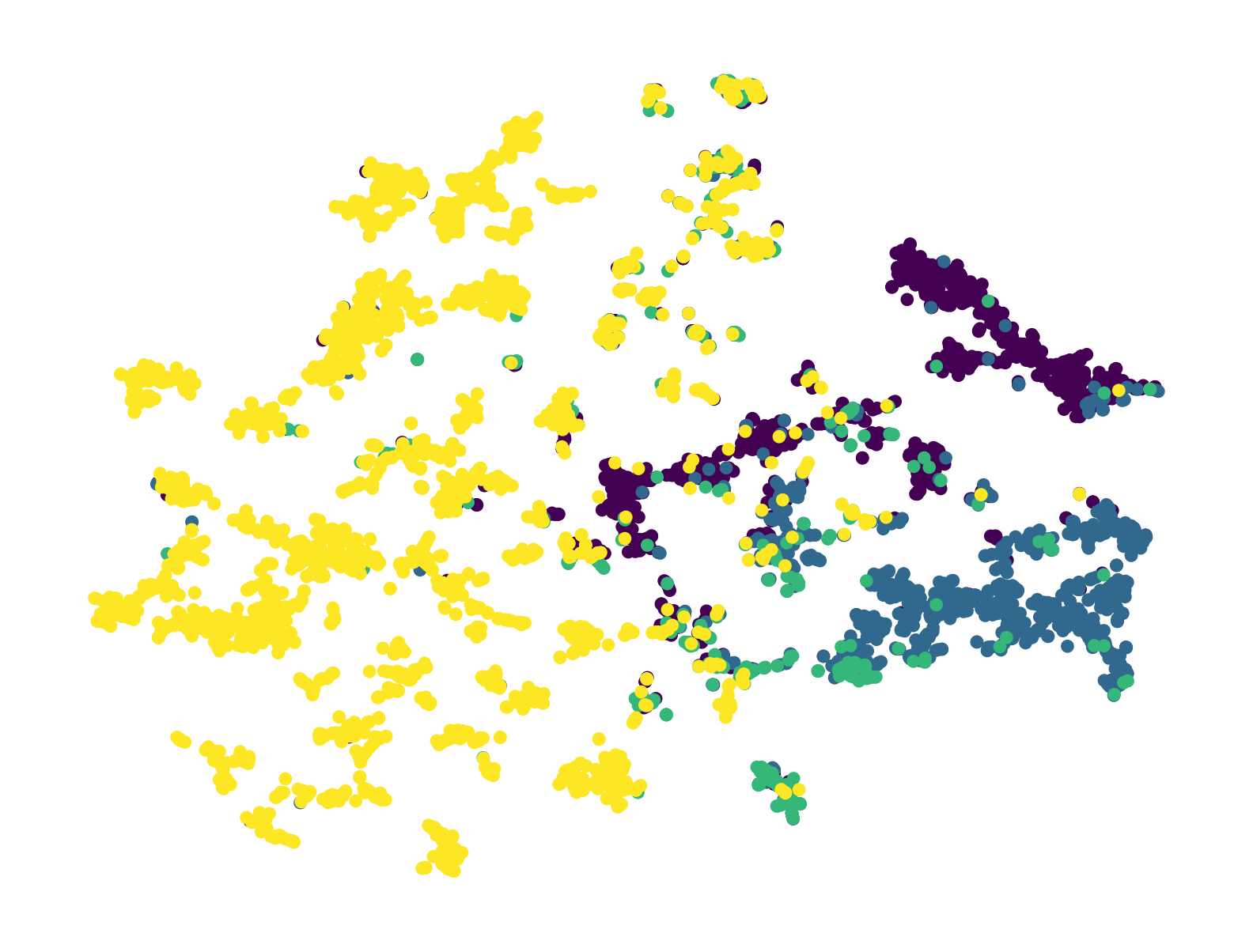}
}
\caption{Visualization of the learned node embedding on AMiner. The Silhouette scores for (a) (b) (c) (d) (e) are 0.0362, 0.1482, 0.1512, \underline{0.1699} and \textbf{0.1917}, respectively.}
\label{visiulization_aminer}
\end{figure*}

\subsection{Additional results on H\lowercase{e}C\lowercase{o}}
\subsubsection{Visualization on AMiner}
\label{Visualization on AMiner}
In this subsection, we visualize the node embeddings on AMiner dataset from Mp2vec, DGI, DMGI, HeCo and HeCo++. The results are shown in Figure~\ref{visiulization_aminer}. As can be seen in the figure, HeCo++ and HeCo again divide different classes more clearly than the other baselines, which is proved by their higher silhouette scores.

\subsection{Additional results on H\lowercase{e}C\lowercase{o}++}
\label{Additional results on}
\subsubsection{Analysis of balance parameters}
\begin{figure}[h]
\centering
\subfigure[ACM: Ma-F1]{
\label{acm_lam}
        \centering
\includegraphics[scale=0.3]{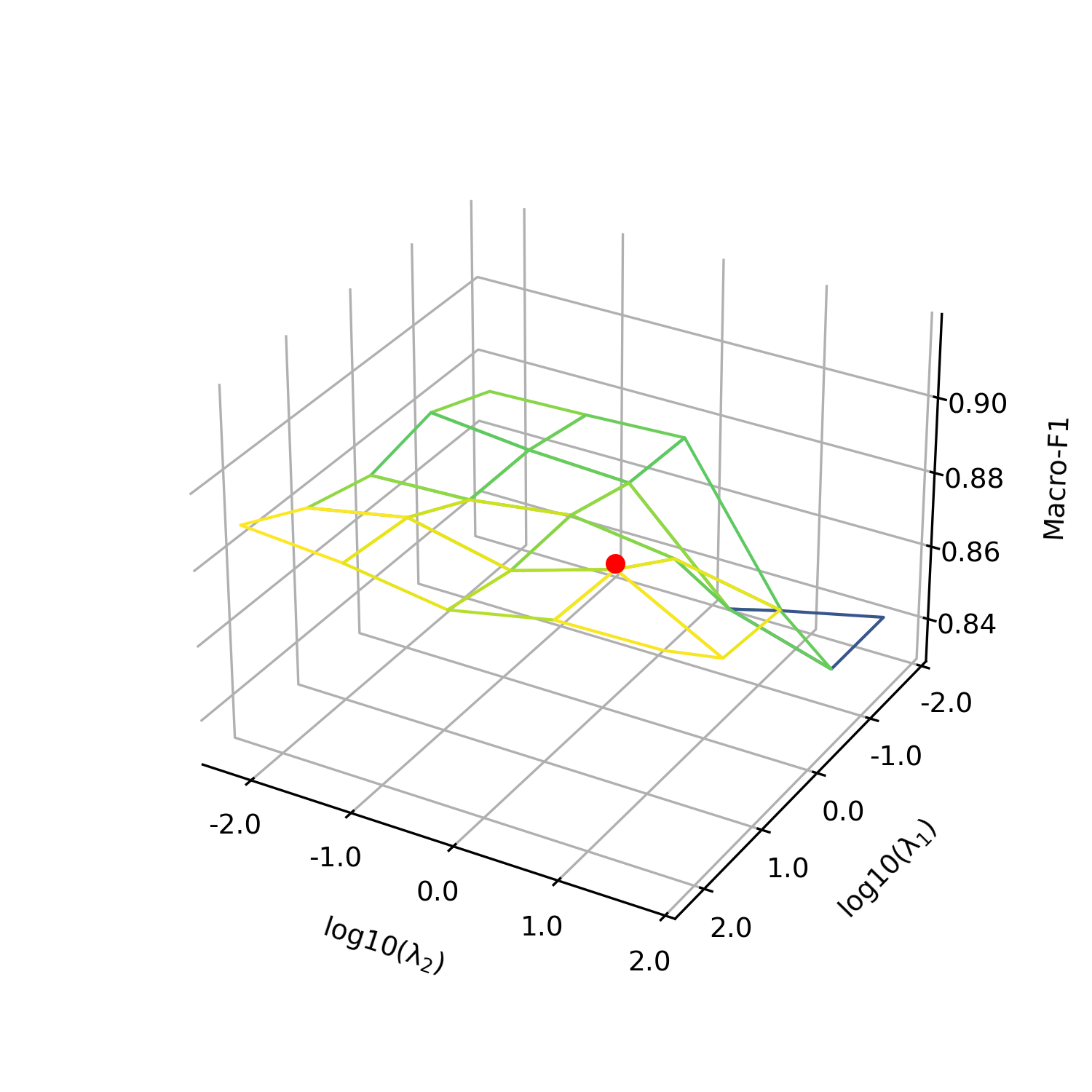}}
\subfigure[DBLP: Ma-F1]{
\label{dblp_lam}
        \centering
\includegraphics[scale=0.3]{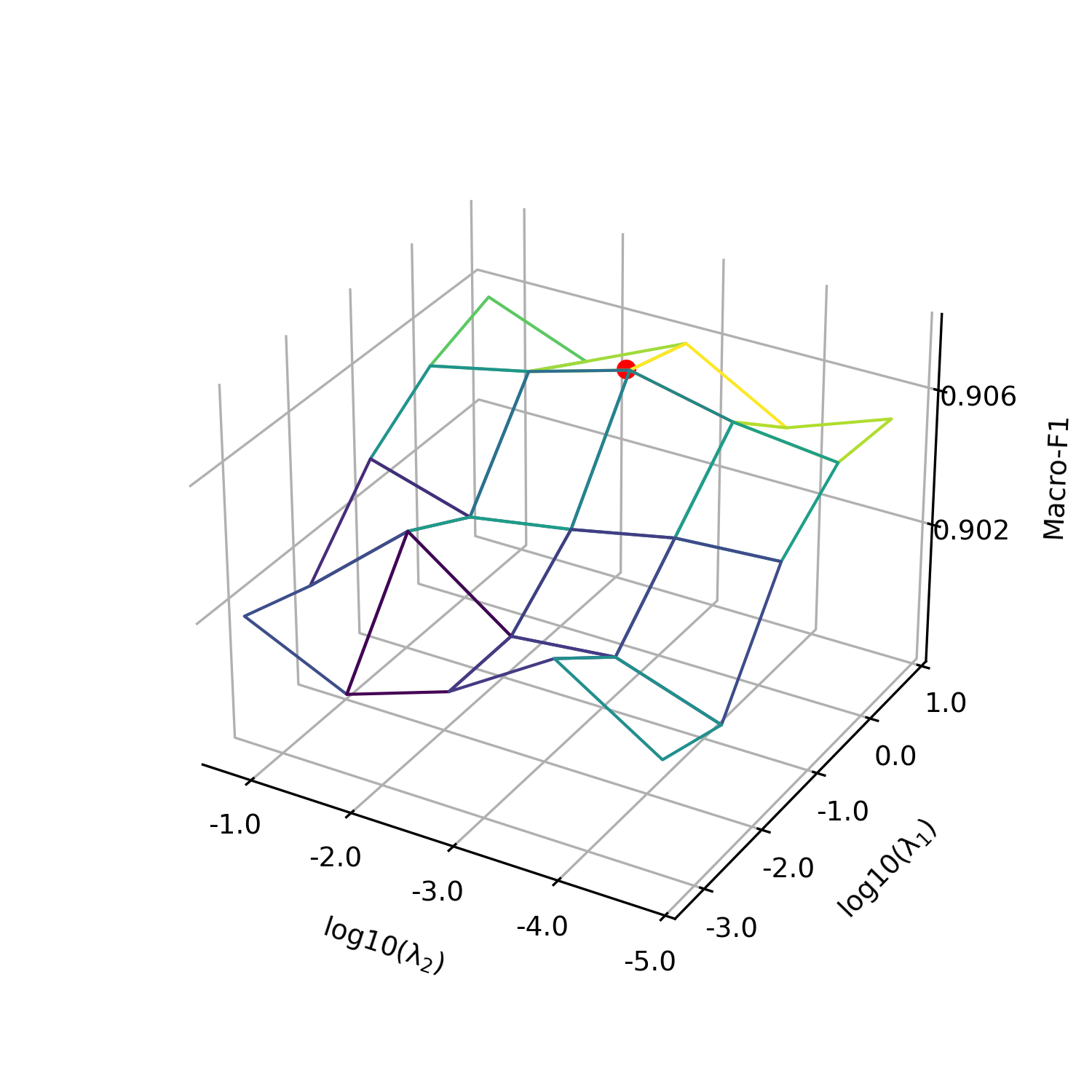}}
\caption{Analysis of the combinations of $\lambda_1$ and $\lambda_2$.}
\label{lam}
\end{figure}

To adjust the influence of two intra-view contrast, we introduce two coefficients $\lambda_1$ and $\lambda_2$ to balance the corresponding losses. We try different combinations of two parameters on ACM and DBLP, and plot all the results on Macro-F1 in Figure \ref{lam}, where the red node in each sub-figure means the best combination. From the results, we can see that the training of ACM is more stable with the change of $\lambda_1$ and $\lambda_2$, while DBLP is more sensitive. Besides, we realize that the final results are more susceptible to the change of $\lambda_1$, which represents intra-view contrast under meta-path view. This observation tallies with the variants analysis that meta-path is a powerful tool to deal with heterogeneous data.

\subsubsection{Analysis of embedding dimension}
\begin{figure}[h]
\centering
\subfigure[ACM]{
\label{sam_acm}
        \centering
\includegraphics[scale=0.17]{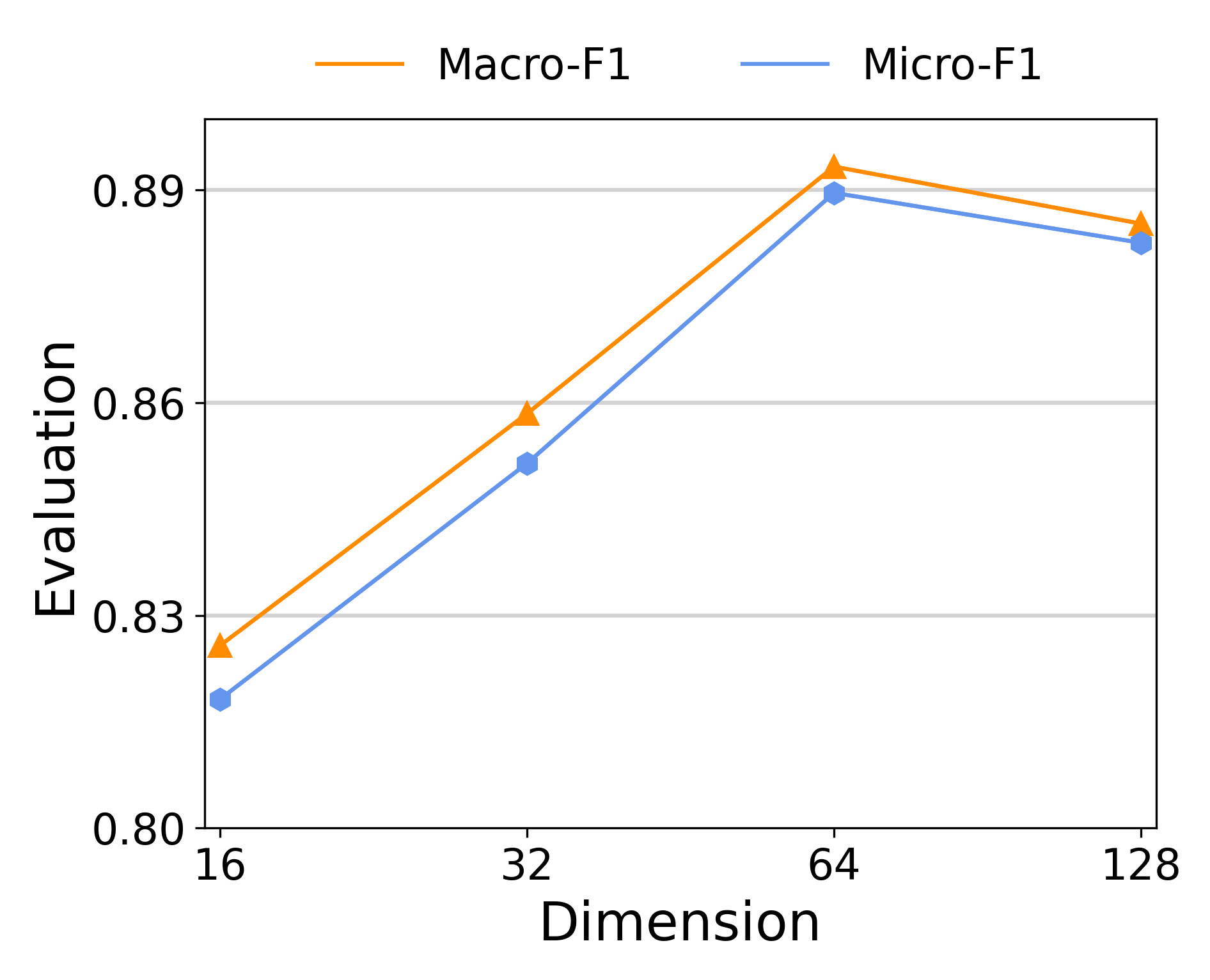}}
\subfigure[DBLP]{
\label{sam_a_aminer}
        \centering
\includegraphics[scale=0.17]{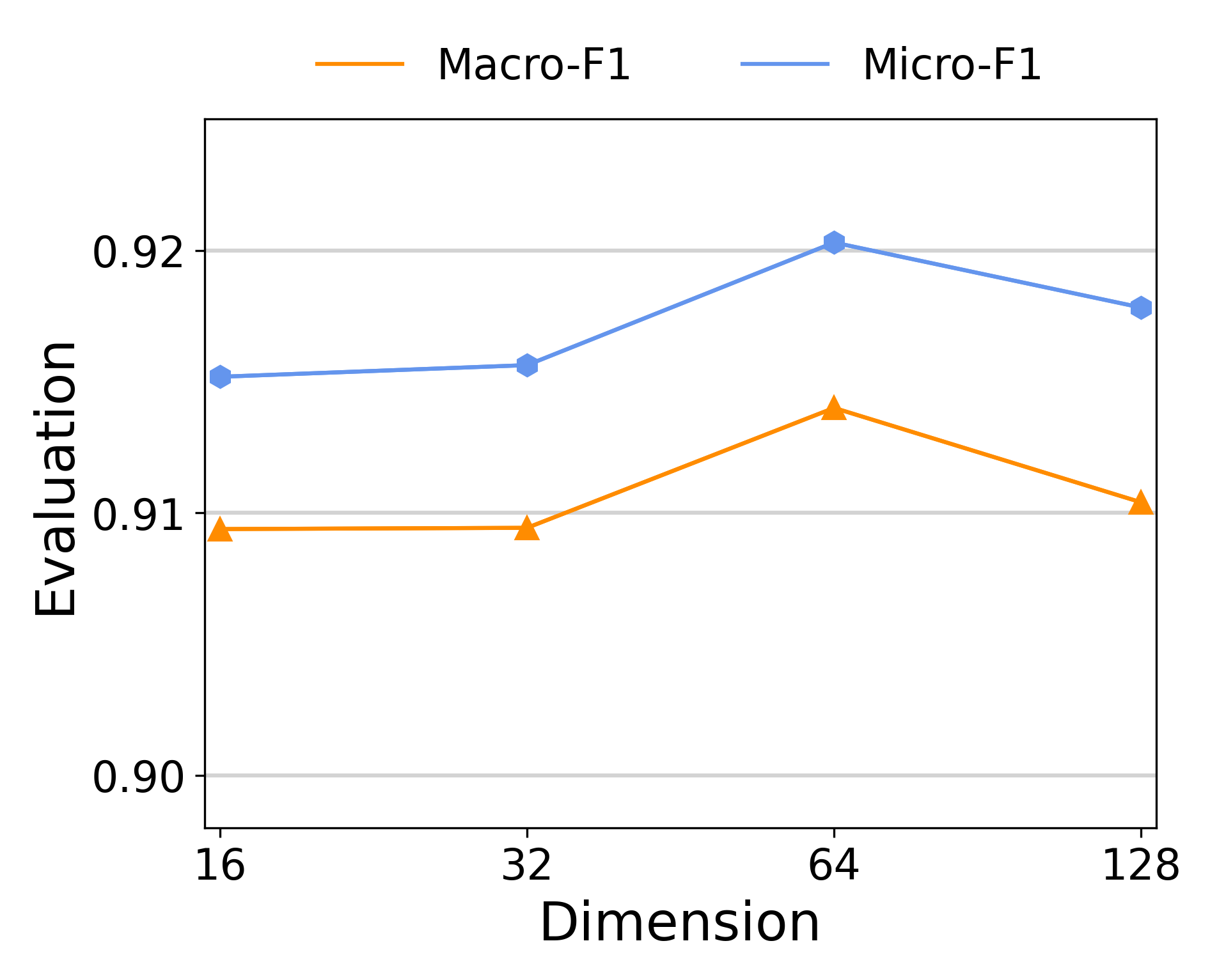}}
\subfigure[Freebase]{
\label{sam_r_aminer}
        \centering
\includegraphics[scale=0.17]{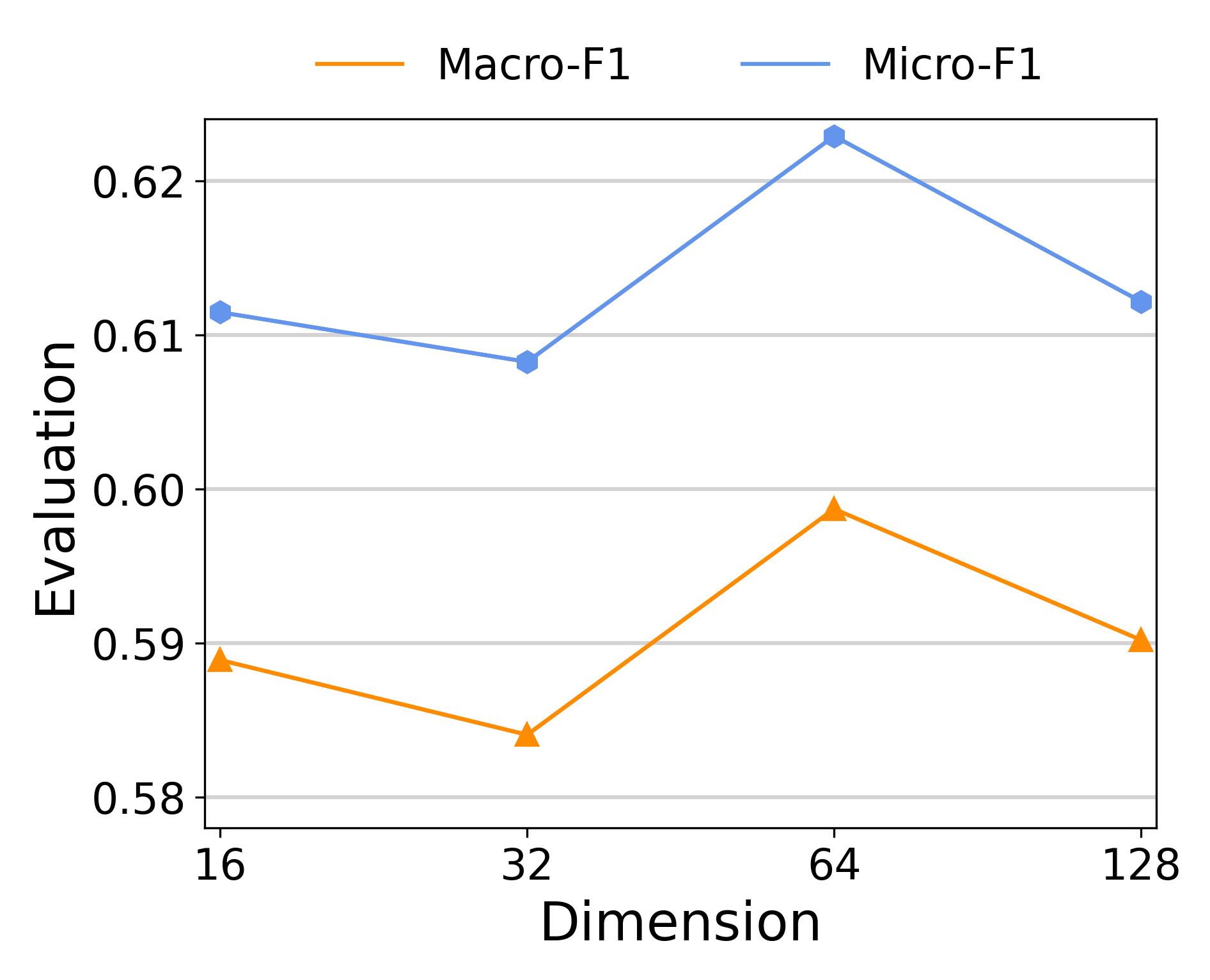}}
\caption{Analysis of embedding dimension.}
\label{dim}
\end{figure}

The embedding dimension is a key parameter to control the complexity and capacity of model. Therefore, we evaluate how it affects the classification performance. As shown in Figure \ref{dim}, as we gradually increase the embedding dimension, the performance generally grows since a larger dimension could enhance the representation capability. Nevertheless, when dimension is larger than the optimal value, increasing dimension will hurt the performance probably due to overfitting. Therefore, we employ the applicable embedding dimension 64 to balance the trade-off between performance and complexity.
\end{document}